\documentclass[12pt]{article}

\pdfoutput=1
\usepackage{adjustbox}
\usepackage{algorithm}
\usepackage{algpseudocode}
\usepackage{amsfonts}
\usepackage{amsmath}
\usepackage{amssymb}
\usepackage{amsthm}
\usepackage{bm}
\usepackage{boldline}
\usepackage{booktabs}
\usepackage{cellspace}
\usepackage{caption}
\usepackage{ctable}
\usepackage{color}
\usepackage[T1]{fontenc}
\usepackage[margin=1.375in]{geometry}
\usepackage{graphicx}
\usepackage{hhline}
\usepackage[hidelinks]{hyperref}
\usepackage[utf8]{inputenc}
\usepackage{multirow}
\usepackage[authoryear]{natbib}
\usepackage{numprint}
\usepackage{subcaption}
\usepackage{scalerel}
\usepackage{url}

\newcommand{\hypercorners}{\{-1, 1\}^N}
\newcommand{\codebooksets}{\mathbb{X}_1, \, \mathbb{X}_2, \, \ldots, \mathbb{X}_F}

\newcommand{\neginnerprod}{\mathcal{L} : \mathbf{x}, \mathbf{y}  \mapsto - \langle \mathbf{x} \, , \mathbf{y} \rangle}
\newcommand{\squarederror}{\mathcal{L} : \mathbf{x}, \mathbf{y} \mapsto \tfrac{1}{2}||\mathbf{x} - \mathbf{y}||^2_2}
\newcommand{\factorgradient}{\nabla_{\mspace{-4mu}\mathbf{a}_f} \mathcal{L}}
\newcommand{\weightgradient}{\nabla_{\mspace{-4mu}\mathbf{w}_f} \tilde{\mathcal{L}}}
\newcommand{\OPsynapses}{\mathbf{X}_f \mathbf{X}_f^\top}
\newcommand{\OLSsynapses}{\mathbf{X}_f{\big( \mathbf{X}_f^\top \mathbf{X}_f \big)}^{-1} \mathbf{X}_f^\top}
\newcommand{\OLSsynapsesdagger}{\mathbf{X}_f\mathbf{X}_f^\dagger}

\newcommand\numberthis{\addtocounter{equation}{1}\tag{\theequation}}

\DeclareMathOperator*{\argmax}{arg\,max}
\DeclareMathOperator*{\argmin}{arg\,min}

\title{\LARGE Resonator Networks outperform optimization methods at solving high-dimensional vector factorization}

\author{
\begin{tabular}[t]{c@{\extracolsep{4ex}}c}
Spencer J. Kent$^{1,2}$ & E. Paxon Frady$^{1,3,5}$ \\
\rule{0pt}{4.5ex}
Friedrich T. Sommer$^{1,3,5}$ \hspace{0.5ex} & Bruno A. Olshausen$^{1,3,4}$  \\
\end{tabular}
}
\date{\vspace{0.2cm}\small{
$^1$Redwood Center for Theoretical Neuroscience \\
$^2$Electrical Engineering and Computer Sciences \\
$^3$Helen Wills Neuroscience Institute,
$^4$School of Optometry \\
      University of California, Berkeley \\
      Berkeley, CA 94702 \\
      $^5$Intel Laboratories, Neuromorphic Computing Lab\\
      San Francisco, CA 94111 \\
    }}

\usepackage{fancyhdr}
\fancypagestyle{firstpage}{
\chead{To appear in Neural Computation, 2020}
}

\begin{document}

\maketitle
\thispagestyle{firstpage}

\vspace{-4ex}
\begin{abstract}
\end{abstract}
\vspace{-4ex}
We develop theoretical foundations of Resonator Networks, a new type of recurrent neural network introduced in \citet{frady2020resonator} to solve a high-dimensional vector factorization problem arising in Vector Symbolic Architectures.
Given a composite vector formed by the Hadamard product between a discrete set of high-dimensional vectors, a Resonator Network can efficiently decompose the composite into these factors.
We compare the performance of Resonator Networks against optimization-based methods, including Alternating Least Squares and several gradient-based algorithms, showing that Resonator Networks are
superior in several important ways. This advantage is achieved by leveraging a combination of nonlinear dynamics and ``searching in superposition,'' by which estimates of the correct solution are
formed from a weighted superposition of all possible solutions. While the alternative methods also search in superposition, the dynamics of Resonator Networks allow them to strike a more effective balance between exploring the solution space and exploiting local information to drive the network toward probable solutions.
Resonator Networks are not guaranteed to converge, but within a particular regime they almost always do. In exchange for relaxing the guarantee of global convergence,
Resonator Networks are dramatically more effective at finding factorizations than all alternative approaches considered.

\newpage
\section{Introduction}
\label{sec:intro}

This article is the second in a two-part series on Resonator Networks. Part one \citep{frady2020resonator} showed how distributed representations of data structures may be formed using the algebra of Vector Symbolic Architectures, and furthermore that decoding these representations often requires solving a vector factorization problem. Resonator Networks were introduced as a neural solution to this problem, and demonstrated on two examples. Here, our primary objective is to establish the theoretical foundations of Resonator Networks, and to perform a more comprehensive analysis of their convergence and capacity properties in comparison to optimization-based methods.

We limit our analysis to a particular definition of the factorization problem, which may seem
somewhat abstract, but in fact applies to practical usage of Vector Symbolic Architectures (VSAs).
We consider ``bipolar'' vectors, whose elements are $\pm1$, used in the popular ``Multiply, Add, Permute (MAP)'' VSA \citep{gayler1998multiplicative, gayler2004vector}.
These ideas extend to other VSAs, although we leave a detailed analysis to future work. Part one included commentary on the historical context and representational power of VSAs,
which we will not cover here. For the purposes of this paper, it is sufficient to stipulate that wherever VSAs are used to encode complex hierarchical data structures, a factorization problem must be solved.
By solving this problem, Resonator Networks make the VSA framework scalable to a larger range of problems.

The core challenge of factorization is that inferring the factors of a composite object amounts to searching through an enormous space of possible solutions. Resonator Networks do this,
in part, by ``searching in superposition,'' a notion that we make precise in Section \ref{sec:superposition_factorization}.
There are in fact many ways to search in superposition, and we introduce a number of them in Section \ref{sec:factorization_via_optimization} as a benchmark for our model and to understand what makes our approach different.
A Resonator Network is simply a nonlinear dynamical system designed to solve a particular factorization problem.
It is defined by equations (\ref{eq:resOP_vectors}) and (\ref{eq:resOLS_vectors}), each representing two separate variants of the network. The system is named
for the way in which correct factorizations seemingly `resonate' out of what is initially an uninformative network state.
The size of the factorization problem that can be reliably solved, as well as the speed with which solutions are found,
characterizes the performance of all the approaches we introduce  --  in these terms, Resonator Networks are by far the most effective.

The main results are as follows:
\begin{enumerate}
    \item We characterize stability at the correct solution, showing that one variant of Resonator Networks is always stable, while the other has stability properties related to classical Hopfield Networks.
          We show that Resonator Networks are less stable than Hopfield Networks because of a phenomenon we refer to as percolated noise (Section \ref{sec:results_percolated_noise}).
    \item We define ``operational capacity'' as a metric of factorization performance and use it to compare Resonator Networks against six benchmark algorithms. We find that Resonator Networks have dramatically
          higher operational capacity (Section \ref{sec:results_op_cap}).
    \item Through simulation, we determine that operational capacity scales as a quadratic function of vector dimensionality. This quantity is proportional to the number of idealized neurons in a
          Resonator Network (also Section \ref{sec:results_op_cap}).
    \item We propose a theory for \emph{why} Resonator Networks perform well on this problem (Section \ref{sec:explaining_operational_capacity}).
\end{enumerate}

\section{Statement of the problem}
\label{sec:problem_statement}
We formalize the factorization problem in the following way: $\codebooksets$ are sets of vectors called `codebooks'. The $f$th codebook contains $D_f$ `codevectors' $\mathbf{x}^{(f)}_1, \mathbf{x}^{(f)}_2, \, \ldots, \mathbf{x}^{(f)}_{D_f}$
\begin{equation*}
\mathbb{X}_f := \{\mathbf{x}^{(f)}_1, \mathbf{x}^{(f)}_2, \, \ldots, \mathbf{x}^{(f)}_{D_f}\} \quad \forall f = 1, 2, \, \ldots, F
\end{equation*}
and these vectors all live in $\hypercorners$.  A composite vector $\mathbf{c}$ is generated by computing the Hadamard product $\odot$ of $F$ vectors, one drawn from each of the codebooks $\codebooksets$.
\begin{gather*}
\mathbf{c} = \mathbf{x}_{\star}^{(1)} \odot \mathbf{x}_{\star}^{(2)} \odot \ldots \odot \mathbf{x}_{\star}^{(F)} \\[5pt]
\mathbf{x}_{\star}^{(1)} \in \mathbb{X}_1, \,\,  \mathbf{x}_{\star}^{(2)} \in \mathbb{X}_2, \,\, \ldots, \,\, \mathbf{x}_{\star}^{(F)} \in \mathbb{X}_F
\end{gather*}
The factorization problem we wish to study is
\begin{equation}
    \begin{tabular}{c c l}
        given & & $\mathbf{c}, \,\, \codebooksets $ \\
        find & & $\mathbf{x}_{\star}^{(1)} \in \mathbb{X}_1, \,\,  \mathbf{x}_{\star}^{(2)} \in \mathbb{X}_2, \,\, \ldots \,\, \mathbf{x}_{\star}^{(F)} \in \mathbb{X}_F$ \\
        such that & & $\mathbf{c} = \mathbf{x}_{\star}^{(1)} \odot \mathbf{x}_{\star}^{(2)} \odot \ldots \odot \mathbf{x}_{\star}^{(F)}$
    \end{tabular}
    \label{eq:general_factorization}
\end{equation}
Our assumption in this paper is that the factorization of $\mathbf{c}$ into $F$ codevectors, one from each codebook, is unique.
Then, the total number of composite vectors that can be generated by the codebooks is $M$:
\[M := \prod_{f=1}^F D_f\]
The problem involves searching among $M$ possible factorizations to find the one that generates $\mathbf{c}$. We will refer to $M$ as
the search space size, and at some level it captures the difficulty of the problem. The problem size is also influenced by $N$, the
dimensionality of each vector.

Suppose we were to solve (\ref{eq:general_factorization}) using a brute force strategy. We might form all possible composite vectors from the sets
$\codebooksets$, one at a time, until we generate the vector $\mathbf{c}$, which would indicate the appropriate factorization. Assuming no additional information
is available, the number of trials taken to find the correct factorization is a uniform random variable $K \sim \mathcal{U} \{1, M \}$ and thus $\mathbf{E}[K] = \frac{M+1}{2}$.
If instead we could easily store all of the composite vectors ahead of time, we could compare them to any new composite vector via a single matrix-vector
inner product, which, given our uniqueness assumption, will yield a value of $N$ for the correct factorization and values strictly less than $N$ for all
other factorizations. The matrix containing all possible composite vectors requires $MN$ bits to store. The core issue is that $M$ scales
\emph{very} poorly with the number of factors and number of possible codevectors to be entertained. If $F=4$ ($4$ factors) and $D_f = 100 \,\, \forall f$ ($100$ possible codevectors for each factor), then $M=100{,}000{,}000$. In the context of Vector Symbolic Architectures,
it is common to have $N=10{,}000$. Therefore, the matrix with all possible composite vectors would require $\approx 125 \, \text{GB}$ to store.
We aspire to solve problems of this size (and much larger), which are clearly out of reach for brute-force approaches. Fortunately, they are
solvable using Resonator Networks.

\section{Factoring by search in superposition}
\label{sec:superposition_factorization}
In our problem formulation (\ref{eq:general_factorization}) the factors interact multiplicatively to form $\mathbf{c}$, and this lies at the heart of what makes it hard to solve. One way to attempt a solution
is to produce an estimate for each factor in turn, alternating between updates to a single factor on its own, with the others held fixed. In addition, it may make sense to simulatenously entertain
all of the vectors in each $\mathbb{X}_f$, in some proportion that reflects our current confidence in each one being part of the correct solution. We call this \emph{searching in superposition} and it is the general approach
we take throughout the paper.
What we mean by `superposition' is that the estimate for the $f$th factor, $\hat{\mathbf{x}}^{(f)}$, is given by $\hat{\mathbf{x}}^{(f)} = g(\mathbf{X}_f \mathbf{a}_f)$ where
$\mathbf{X}_f$ is a matrix with each column a vector from $\mathbb{X}_f$. The vector $\mathbf{a}_f$ contains the coefficients that define a linear combination of the elements of
$\mathbb{X}_f$, and $g(\cdot)$ is a function from $\mathbb{R}^N$ to $\mathbb{R}^N$, which we will call the activation function. In this paper we consider the identity $g: \mathbf{x} \mapsto \mathbf{x}$, the sign function $g: \mathbf{x} \mapsto \text{sgn}(\mathbf{x})$, and nothing else. Other activation functions are appropriate for the other variants of Resonator Networks (for instance where the vectors are complex-valued), but we leave a discussion of this to future work.
`Search' refers to the method by which we adapt $\mathbf{a}_f$ over time.
The estimate for each factor leads to an estimate for $\mathbf{c}$ denoted
by $\hat{\mathbf{c}}$:
\begin{equation}
    \hat{\mathbf{c}} := \hat{\mathbf{x}}^{(1)} \odot \hat{\mathbf{x}}^{(2)} \odot \ldots \odot \hat{\mathbf{x}}^{(F)} = g(\mathbf{X}_1 \mathbf{a}_1) \odot g(\mathbf{X}_2
    \mathbf{a}_2) \odot \ldots \odot g(\mathbf{X}_F \mathbf{a}_F)
    \label{eq:chat_nonlinear}
\end{equation}
Suppose $g(\cdot)$ is the identity. Then $\hat{\mathbf{c}}$ becomes a \emph{multilinear} function of the coefficients $\mathbf{a}_1, \mathbf{a}_2, \ldots \mathbf{a}_F$.
\begin{equation}
    \hat{\mathbf{c}} = \hat{\mathbf{x}}^{(1)} \odot \hat{\mathbf{x}}^{(2)} \odot \ldots \odot \hat{\mathbf{x}}^{(F)} = \mathbf{X}_1 \mathbf{a}_1 \odot \mathbf{X}_2 \mathbf{a}_2 \odot \ldots \odot \mathbf{X}_F \mathbf{a}_F
    \label{eq:chat_linear}
\end{equation}
While this is a `nice' relationship in the sense that it is linear in each of the coefficients $\mathbf{a}_f$ separately (with the others held fixed), it is unfortunately not convex with respect to the
coefficients taken all at once. We can rewrite it as a sum of $M$ different terms, one for each of the possible factorizations of $\mathbf{c}$:
\begin{equation}
    \hat{\mathbf{c}} = \sum_{d_1, d_2, \ldots, d_F} \Big( {(\mathbf{a}_1)}_{\mspace{-2mu}d_1} {(\mathbf{a}_2)}_{\mspace{-2mu}d_2}\ldots{(\mathbf{a}_F)}_{\mspace{-2mu}d_F} \Big) \,\, \mathbf{x}_{d_1}^{(1)} \odot \mathbf{x}_{d_2}^{(2)} \odot \ldots \odot \mathbf{x}_{d_F}^{(F)}
    \label{eq:chat_linear_rearranged}
\end{equation}
Where $d_1$ ranges from $1$ to $D_1$, $d_2$ ranges from $1$ to $D_2$, and so on. The term in parentheses is a scalar that weights each of the possible Hadamard products. Our estimate $\hat{\mathbf{c}}$ is, at any given
time, purely a superposition of \emph{all} the possible factorizations. Moreover, the superposition weights $\Big( {(\mathbf{a}_1)}_{\mspace{-2mu}d_1} {(\mathbf{a}_2)}_{\mspace{-2mu}d_2}\ldots{(\mathbf{a}_F)}_{\mspace{-2mu}d_F} \Big)$ can be approximately
recovered from $\hat{\mathbf{c}}$ alone by computing the cosine similarity between $\hat{\mathbf{c}}$ and the vector $\mathbf{x}_{d_1}^{(1)} \odot \mathbf{x}_{d_2}^{(2)} \odot \ldots \odot \mathbf{x}_{d_F}^{(F)}$. The source of `noise' in this approximation
is the fact that $\mathbf{x}_{m_1}^{(1)} \odot \mathbf{x}_{m_2}^{(2)} \odot \ldots \odot \mathbf{x}_{m_F}^{(F)}$ will have a nonzero inner product with the other vectors in the sum. When the codevectors are uncorrelated and high-dimensional,
this noise is quite small: $\hat{\mathbf{c}}$ transparently reflects the proportion with which it contains each of the possible factorizations.
When $g(\cdot)$ is the sign function, \emph{this property is retained}. The vector $\hat{\mathbf{c}}$ is
no longer an exact superposition, but the scalar $\Big( {(\mathbf{a}_1)}_{\mspace{-2mu}m_1} {(\mathbf{a}_2)}_{\mspace{-2mu}m_2}\ldots{(\mathbf{a}_F)}_{\mspace{-2mu}m_F} \Big)$ can still be decoded from $\hat{\mathbf{c}}$ in the same way --
the vector $\hat{\mathbf{c}}$ is still an approximate superposition of all the possible factorizations, with the weight for each of these determined by the coefficients $\mathbf{a}_f$. This property, that thresholded superpositions retain
relative similarity to each of their superimposed components, is heavily relied on throughout Kanerva's and Gayler's work on Vector Symbolic Architectures \citep{kanerva1996binary, gayler1998multiplicative}.

One last point of notation before introducing our solution to the factorization problem -- we define the vector $\hat{\mathbf{o}}^{(f)}$ to be the product of the estimates for the \emph{other} factors:
\begin{equation}
\label{eq:defn_of_o_no_time}
\hat{\mathbf{o}}^{(f)} := \hat{\mathbf{x}}^{(1)} \odot \ldots \odot \hat{\mathbf{x}}^{(f-1)}
      \odot \hat{\mathbf{x}}^{(f+1)} \odot \ldots \odot \hat{\mathbf{x}}^{(F)}
\end{equation}
This will come up in each of the algorithms under consideration and simplify our notation.
The notation will often include an explicit dependence on time $t$ like so: $\hat{\mathbf{x}}_f[t] = g(\mathbf{X}_f \mathbf{a}_f[t])$.
Each of the algorithms considered in this paper updates one factor at a time, with the others held fixed so, at a given time $t$,
we will update the factors in order $1$ to $F$, although this is a somewhat arbitrary choice. Including time dependence with
$\hat{\mathbf{o}}^{(f)}$, we have
\begin{equation}
\label{eq:defn_of_o}
\hat{\mathbf{o}}^{(f)}[t] := \hat{\mathbf{x}}^{(1)}[t+1] \odot \ldots \odot \hat{\mathbf{x}}^{(f-1)}[t+1] \odot \hat{\mathbf{x}}^{(f+1)}[t] \odot \ldots \odot \hat{\mathbf{x}}^{(F)}[t]
\end{equation}
which makes explicit that at the time of updating $\hat{\mathbf{x}}_f$, the factors $1$ to $(f-1)$ have already been updated for this `iteration' $t$ while the factors $(f+1)$ to $F$ have yet to be updated.

\section{Resonator Networks}
\label{sec:resonator_networks}
A Resonator Network is a nonlinear dynamical system designed to solve the factorization problem (\ref{eq:general_factorization}), and it can be interpreted as a neural network in which
idealized neurons are connected in a very particular way. We define two separate variants of this system, which differ in terms of this pattern of connectivity.
A Resonator Network with outer product (OP) weights is defined by
\begin{equation}
    \label{eq:resOP_vectors}
    \hat{\mathbf{x}}^{(f)}[t+1] = \text{sgn} \Big(\OPsynapses \big(\hat{\mathbf{o}}^{(f)}[t] \odot \mathbf{c}\big) \Big)
\end{equation}
Suppose $\hat{\mathbf{x}}^{(f)}[t+1]$ indicates the state of a population of neurons at time $t+1$. Each neuron receives an input $\hat{\mathbf{o}}^{(f)}[t] \, \odot \, \mathbf{c}$, modified by synapses modeled as a row of
a ``weight matrix'' $\OPsynapses$. This ``synaptic current'' is passed through the activation function $\text{sgn}(\cdot)$ in order to determine the output, which is either $+1$ or $-1$.
Most readers will be familiar with the weight matrix $\OPsynapses$ as the so-called ``outer product'' learning rule of classical Hopfield Networks \citep{hopfield1982neural}.
This has the nice interpretation of Hebbian learning \citep{hebb1949organization} in which the strength of synapses between any two neurons (represented by this weight matrix) depends solely on their
pairwise statistics over some dataset, in this case the codevectors.

Prior to thresholding in (\ref{eq:resOP_vectors}), the matrix-vector product $\mathbf{X}^\top \big(\hat{\mathbf{o}}^{(f)}[t] \odot \mathbf{c}\big)$ produces coefficients $\mathbf{a}_f[t]$ which, when
premultiplied by $\mathbf{X}_f$, generate a vector in the linear subspace spanned by the codevectors (the columns of $\mathbf{X}_f$).
This projection does not minimize the squared distance between $\big(\hat{\mathbf{o}}^{(f)}[t] \odot \mathbf{c}\big)$ and the resultant vector.
Instead, the matrix ${\big( \mathbf{X}_f^\top \mathbf{X}_f \big)}^{-1} \mathbf{X}_f^\top$ produces such a projection, the so-called \emph{Ordinary Least Squares} projection onto $\mathcal{R}(\mathbf{X}_f)$.
This motivates the second variant of our model, Resonator Networks with Ordinary Least Squares (OLS) weights:
\begin{align*}
    \label{eq:resOLS_vectors}
    \hat{\mathbf{x}}^{(f)}[t+1] =& \, \text{sgn} \Big(\OLSsynapses \big(\hat{\mathbf{o}}^{(f)}[t] \odot \mathbf{c}\big) \Big) \\
    :=& \, \text{sgn} \Big(\OLSsynapsesdagger \big(\hat{\mathbf{o}}^{(f)}[t] \odot \mathbf{c}\big) \Big) \numberthis
\end{align*}
where we have used the notation $\mathbf{X}_f^\dagger$ to indicate the Moore-Penrose psuedoinverse of the matrix $\mathbf{X}_f$.
Hopfield Networks with this type of synapse were first proposed by Personnaz, Guyon, and Dreyfus \citep{personnaz1986collective}, who called this the ``projection'' rule.

If, contrary to what we have defined in (\ref{eq:resOP_vectors}) and (\ref{eq:resOLS_vectors}), the
input to each sub-population of neurons was $\hat{\mathbf{x}}^{(f)}[t]$, its own previous state, then one would in fact have a (``Bipolar'') Hopfield Network.
In our case however, rather than being autoassociative, in which $\hat{\mathbf{x}}^{(f)}[t+1]$ is a direct function of $\hat{\mathbf{x}}^{(f)}[t]$, our dynamics are heteroassociative,
basing updates on the states of the \emph{other} factors. This change has a dramatic effect on the network's convergence properties and is also in some sense what makes Resonator Networks useful in solving the factorization problem,
a fact that we will elaborate on in the following sections. We imagine $F$ separate subpopulations of neurons which evolve together in time, each one responsible for estimating a different factor of $\mathbf{c}$. For now we have just
specified this as a discrete-time network in which updates are made one-at-a-time, but it can be extended as a continuous-valued, continuous-time dynamical system along the same lines as was done for
Hopfield Networks \citep{hopfield1984neurons}. In that case, we can think about these $F$ subpopulations of neurons evolving in a truly parallel way. In discrete-time, one has the choice of making
`asynchronous' or `synchronous' updates to the factors, in a sense analogous to Hopfield Networks. Our formulation of $\hat{\mathbf{o}}^{(f)}[t]$ in (\ref{eq:defn_of_o}) follows the asynchronous
convention, which we find to converge faster. The formulation given in part one of this series \citep{frady2020resonator} employed the synchronous convention for pedagogical reasons, but the distinction
between the two vanishes in continuous-time, where updates are instantaneous.

In practice, we will have to choose an initial state $\hat{\mathbf{x}}^{(f)}[0]$ using no knowledge of the correct codevector $\mathbf{x}^{(f)}_{\star}$ other than the fact it is one of the elements of the codebook $\mathbb{X}_f$.
Therefore, we set $\hat{\mathbf{x}}^{(f)}[0] = \text{sgn} \Big (\sum_j \mathbf{x}^{(f)}_j \Big)$, which, as we have said above, has approximately equal cosine similarity to each term in the sum.

\subsection{Difference between OP weights and OLS weights}
The difference between outer product weights and Ordinary Least Squares weights is via ${\big( \mathbf{X}_f^\top \mathbf{X}_f \big)}^{-1}$, the inverse of the so-called Gram matrix for $\mathbf{X}_f$, which contains inner products
between each codevector. If the codevectors are orthogonal, the Gram matrix is $N \, \mathbf{I}$, with $\mathbf{I}$ the identity matrix. When $N$ is large (roughly speaking $> 5{,}000$), and the codevectors are chosen
randomly i.i.d. from $\{-1, 1\}^N$, then they will be \emph{very nearly} orthogonal, making $N \, \mathbf{I}$ a close approximation.
Clearly, in this setting, the two variants of Resonator Networks produce nearly the same dynamics. In section \ref{sec:results_op_cap}, we define and measure a performance metric
called operational capacity in such a way that does not particularly highlight the difference between the dynamics, i.e. it is the setting where codevectors are nearly orthogonal. In general, however, the dynamics are clearly different.
In our experience, applications that contain correlations between codevectors may enjoy higher operational capacity under Ordinary Least Squares weights, but it is hard to say whether this applies in every setting.

One application-relevant consideration is that, because $\mathbf{X}_f$ consists of entries that are $+1$ and $-1$, the outer product variant of a Resonator Network has an integer-valued weight matrix and
can be implemented without any floating-point computation -- hardware with large binary and integer arithmetic circuits can simulate this model very quickly.
Coupled with noise tolerance properties we will establish in Section \ref{sec:noisy_factorization}, this makes Resonator Networks (and more generally, VSAs) a good fit for emerging device nanotechnologies \citep{rahimi2017high}.

\section{The optimization approach}
\label{sec:factorization_via_optimization}
An alternative strategy for solving the factorization problem is to define a loss function which compares
the current estimate $\hat{\mathbf{c}} := \hat{\mathbf{x}}^{(1)} \odot \hat{\mathbf{x}}^{(2)} \odot \ldots \odot \hat{\mathbf{x}}^{(F)}$
with the composite that is to be factored, $\mathbf{c}$, choosing the loss function and a corresponding constraint set so that the global minimizer
of this loss over the constraints yields the correct solution to (\ref{eq:general_factorization}). One can then design an algorithm that finds the solution by minimizing this loss. This is the
approach taken by \emph{optimization} theory.
Here we consider algorithms that search in superposition, setting $\hat{\mathbf{x}}^{(f)} = g(\mathbf{X}_f \mathbf{a}_f)$ just as Resonator Networks, but that instead take the optimization approach.

Let the loss function be
$\mathcal{L}(\mathbf{c}, \hat{\mathbf{c}})$ and the feasible set for each $\mathbf{a}_f$ be $C_f$. We write this as a fairly generic optimization
problem:
\begin{equation}
    \begin{aligned}
    & \underset{\mathbf{a}_1, \mathbf{a}_2, \ldots, \mathbf{a}_F}{\text{minimize}} & & \mathcal{L} \big( \mathbf{c}, g(\mathbf{X}_1 \mathbf{a}_1) \odot g(\mathbf{X}_2 \mathbf{a}_2) \odot \ldots \odot g(\mathbf{X}_F \mathbf{a}_F) \big) \\
    & \text{subject to} & & \,\, \mathbf{a}_1 \in C_1, \mathbf{a}_2 \in C_2, \ldots, \mathbf{a}_F \in C_F
    \end{aligned}
    \label{eq:superposition_factorization}
\end{equation}
What makes a particular instance of this problem remarkable depends on our choices for $\mathcal{L}(\cdot, \cdot)$, $g(\cdot)$, $C_1, C_2, \ldots, C_F$, and the structure of the vectors
in each codebook. Different algorithms may be appropriate for this problem, depending on these details, and we propose \emph{six} candidate algorithms in this paper, which we refer to as the ``benchmarks''.
It is in \emph{contrast} to the benchmark algorithms that we can more fully understand the performance of Resonator Networks -- our argument, which we will develop in the Results section, is that Resonator Networks strike a more natural
balance between exploring the high-dimensional state space and using local information to move towards the solution. The benchmark algorithms are briefly introduced in Section
\ref{sec:benchmark_algs}, but they are each discussed at some length in the Appendix, including Table \ref{tab:linear_unit_dynamics}, which compiles the dynamics specified by each. We provide implementations of each algorithm in the small software library
that accompanies this paper\footnote{\normalsize \href{https://github.com/spencerkent/resonator-networks}{https://github.com/spencerkent/resonator-networks}}.

\subsection{Benchmark algorithms}
\label{sec:benchmark_algs}
A common thread among the benchmark algorithms is that they take the activation function $g(\cdot)$ to be the identity $g: \mathbf{x} \mapsto \mathbf{x}$, making $\hat{\mathbf{c}}$ a multilinear function of the coefficients,
as we discussed in section \ref{sec:superposition_factorization}. We experimented with other activation functions, but found none for which the optimization approach performed better.
We consider two straightforward loss functions for comparing $\mathbf{c}$ and $\hat{\mathbf{c}}$. The first is one half the squared Euclidean norm of the error, $\squarederror$, which we will call the squared error for short,
and the second is the negative inner product $\neginnerprod$. The squared error is minimized by $\hat{\mathbf{c}} = \mathbf{c}$, which is also true for the negative inner product when $\hat{\mathbf{c}}$ is constrained to $[-1, 1]^N$.
Both of these loss functions are convex,
meaning that $\mathcal{L}(\mathbf{c}, \hat{\mathbf{c}})$ is a convex function of each $\mathbf{a}_f$ separately\footnote{\normalsize through the composition of an affine function with a convex function}.
\emph{Some} of the benchmark algorithms constrain $\mathbf{a}_f$ directly, and when that is the case, our focus is on
three different convex sets, namely the simplex
$\Delta_{D_f} := \{\mathbf{x} \in \mathbb{R}^{D_f} \mid \sum_i x_i = 1, x_i \geq 0 \,\, \forall i\}$, the unit $\ell_1$ ball
$\mathcal{B}_{||\cdot||_1}[1] := \{\mathbf{x} \in \mathbb{R}^{D_f} \mid ||\mathbf{x}||_1 \leq 1 \}$, and the closed zero-one hypercube $[0, 1]^{D_f}$.
Therefore, solving (\ref{eq:superposition_factorization}) with respect to each $\mathbf{a}_f$ \emph{separately} is a convex optimization problem. In the case of the negative inner product loss $\neginnerprod$
and simplex constraints $C_f = \Delta_{D_f}$, it is a bonafide linear program. The correct factorization is given by $\mathbf{a}_1^\star, \mathbf{a}_2^\star, \ldots, \mathbf{a}_F^\star$ such that
$\hat{\mathbf{x}}^{(f)} = \mathbf{X}_f \mathbf{a}_f^\star = \mathbf{x}_{\star}^{(f)} \,\,\, \forall f$, which we know to be vectors with a single entry $1$ and the rest $0$
 --  these are the standard basis vectors $\mathbf{e}_i$ (where ${(\mathbf{e}_i)}_j = 1$ if $j = i$ and $0$ otherwise). The initial states $\mathbf{a}_1[0], \mathbf{a}_2[0], \ldots \mathbf{a}_F[0]$ must be set with
no prior knowledge of the correct factorization so, similar to how we do for Resonator Networks, we set each element of $\mathbf{a}_f[0]$ to the same
value (which in general depends on the constraint set).

\subsubsection{Alternating Least Squares}
Alternating Least Squares (ALS) locally minimizes the squared error loss in a fairly straightforward way:
for each factor, one at a time, it solves a least squares problem for $\mathbf{a}_f$ and updates the current state of
the estimate $\hat{\mathbf{c}}$ to reflect this new value, then moves onto the next factor and repeats. Formally, the updates given by Alternating Least Squares are:
\begin{align*}
    \label{eq:ALS_dynamics}
    \mathbf{a}_f[t+1] &= \argmin_{\mathbf{a}_f}  \,\, \tfrac{1}{2} {\big|\big| \, \mathbf{c} - \hat{\mathbf{o}}^{(f)}[t] \odot \mathbf{X}_f \mathbf{a}_f[t] \, \big|\big|}_2^2 \\
                      &= {\big( \bm{\xi}^\top \bm{\xi} \big)}^{-1} \bm{\xi}^\top \mathbf{c}  \quad | \quad \bm{\xi} := \text{diag}\big(\hat{\mathbf{o}}^{(f)}[t]\big) \, \mathbf{X}_f \numberthis
\end{align*}
Alternating Least Squares is an algorithm that features prominently in the tensor decomposition literature \citep{kolda2009tensor}, but
while ALS has been successful for a particular type of tensor decomposition, there are a few details which make our problem
different from what is normally studied (see Appendix \ref{appendix:Tensor_decomp}). The updates in ALS are quite greedy -- they exactly solve each least squares subproblem. It may make sense to more gradually modify the coefficients, a strategy that we turn to next.

\subsubsection{Gradient-following algorithms}
Another natural strategy for solving (\ref{eq:superposition_factorization}) is to make updates that incorporate the gradient of $\mathcal{L}$ with respect to the coefficients -- each of the next $5$ algorithms does this in a particular way
(we write out the gradients for both loss functions in Appendix \ref{appendix:general_grad_stuff}).
The squared error loss is globally minimized by $\hat{\mathbf{c}} = \mathbf{c}$, so one might be tempted to start from some initial values for the coefficients and make gradient updates $\mathbf{a}_f[t+1] = \mathbf{a}_f[t] - \eta \, \factorgradient$.
In Appendix \ref{appendix:fixed_stepsize_doesnt_work} we discuss why this does not work well --
the difficulty is in being able to guarantee that the loss function is {smooth} enough that gradient descent
iterates with a fixed stepsize will converge. Instead, the algorithms we apply to the squared error loss utilize a dynamic stepsize.
\begin{description}
    \item[Iterative Soft Thresholding:] The global minimizers of (\ref{eq:superposition_factorization}) are maximally sparse, $||\mathbf{a}^{\star}_f||_0 = 1$.
    If one aims to minimize the squared error loss while \emph{loosely} constrained to sparse solutions, it may make sense to solve the problem with
    Iterative Soft Thesholding (ISTA). The dynamics for ISTA are given by equation (\ref{eq:ISTA_dynamics}) in Table \ref{tab:linear_unit_dynamics}.

    \item[Fast Iterative Soft Thresholding:] We also considered Fast Iterative Soft Thesholding (FISTA), an enhancement due to \citet{beck2009fast}, which utilizes
    Nesterov's momentum for accelerating first-order methods in order to alleviate the sometimes slow convergence of ISTA \citep{bredies2008linear}.
    Dynamics for FISTA are given in equation (\ref{eq:FISTA_dynamics}).

    \item[Projected Gradient Descent:] Another benchmark algorithm we considered was Projected Gradient Descent on the negative inner product loss,
    where updates were projected onto either the simplex or unit $\ell_1$ ball (\ref{eq:PGD_dynamics}). A detailed discussion of this approach can be found
    in Appendix \ref{appendix:PGD}.

    \item[Multiplicative Weights:] This is an algorithm that can be applied to either loss function, although we found it worked best on the negative inner product.
    It very elegantly enforces a simplex constraint on $\mathbf{a}_f$ by maintaining a set of auxilliary variables, the `weights', which are used to set $\mathbf{a}_f$ at each iteration.
    See equation (\ref{eq:MW_dynamics}) for the dynamics of Multiplicative Weights, as well as Appendix \ref{appendix:MW}.

    \item[Map Seeking Circuits:] The final algorithm that we considered is called Map Seeking Circuits. Map Seeking Circuits are neural networks designed to solve invariant
    pattern recognition problems using the principle of superposition. Their dynamics are based on the gradient, but are different from what we have introduced so
    far -- see equation (\ref{eq:MSC_dynamics}) and Appendix \ref{appendix:MSC}.
\end{description}

\subsection{Contrasting Resonator Networks with the benchmarks}
\subsubsection{Convergence of the benchmarks}
A remarkable fact about the benchmark algorithms is that \emph{each one converges for all initial conditions}, which we directly prove, or refer to results proving, in the Appendix.
That is, given any starting coefficients $\mathbf{a}_f[0]$, their dynamics reach
fixed points which are local minimizers of the loss function. In some sense, this property is an immediate consequence of treating factorization as an optimization problem -- the
algorithms we chose as the benchmarks were \emph{designed} this way.
Convergence to a local minimizer is a desirable property, but unfortunately the fundamental non-convexity of the optimization problem implies that this may not guarantee good local minima in practice.
In Section \ref{sec:results}
we establish a standardized setting where we measure how likely it is that these local minima are actually \emph{global} minima. We find that
as long as $M$ -- the size of the search space -- is small enough, each of these algorithms can find the global minimizers reliably. The
point at which the problem becomes too large to reliably solve is what we call the operational capacity of the algorithm, and is a main point of comparison with Resonator Networks.

\subsubsection{An algorithmic interpretation of Resonator Networks}
\label{sec:alg_interp_of_res_networks}
The benchmark algorithms generate estimates for the factors, $\hat{\mathbf{x}}^{(f)}[t]$, that move through the interior of the $[-1, 1]$ hypercube. Resonator Networks, on the other hand, do not.
The $\text{sgn}(\cdot)$ function `bipolarizes' inputs to the nearest vertex of the hypercube, and this highly nonlinear function, which not only changes the length but also the \emph{angle} of an input vector, is key.
We know the solutions $\mathbf{x}_\star^{(f)}$ exist at vertices of the hypercube and
these points are very special geometrically in the sense that in high dimensions, most of the mass of $[-1, 1]^N$ is concentrated relatively \emph{far} from the vertices -- a fact we will not prove here but
that is based on standard results from the study of concentration inequalities \citep{boucheron2013concentration}. Our motivation for using the $\text{sgn}(\cdot)$ activation function is that moving through the interior of
the hypercube while searching for a factorization is unwise, a conjecture for which we will provide some empirical support in the Results section.

One useful interpretation of OLS Resonator Network dynamics is that the network is computing a bipolarized version of Alternating Least Squares.
Suppose we were to take the dynamics specified in ($\ref{eq:ALS_dynamics}$) for making ALS updates to $\mathbf{a}_f[t+1]$, but we also bipolarize the vector $\hat{\mathbf{x}}^{(f)}[t+1]$ at the end of each step.
When each $\hat{\mathbf{x}}^{(f)}[t+1]$ is bipolar, the vector $\hat{\mathbf{o}}^{(f)}[t]$ is bipolar and we can simplify ${\big( \bm{\xi}^\top \bm{\xi} \big)}^{-1} \bm{\xi}^\top$:
\begin{align*}
    \hat{\mathbf{o}}^{(f)}[t] \in {\{ -1, 1\}}^N \iff {\big( \bm{\xi}^\top \bm{\xi} \big)}^{-1} \bm{\xi}^\top &= {\big( \mathbf{X}_f^\top \text{diag}\big(\hat{\mathbf{o}}^{(f)}[t]\big)^2 \mathbf{X}_f \big)}^{-1} \mathbf{X}_f^\top \text{diag}\big(\hat{\mathbf{o}}^{(f)}[t]\big) \\
    &= {\big( \mathbf{X}_f^\top \mathbf{X}_f \big)}^{-1} \mathbf{X}_f^\top \, \text{diag}\big(\hat{\mathbf{o}}^{(f)}[t]\big) \\
    &= \mathbf{X}_f^\dagger \, \text{diag}\big(\hat{\mathbf{o}}^{(f)}[t]\big) \numberthis
\end{align*}
Now $\mathbf{a}_f[t+1] = \mathbf{X}_f^\dagger \big(\hat{\mathbf{o}}^{(f)}[t] \odot \mathbf{c}\big)$, which one can see from equation (\ref{eq:resOLS_vectors}) is precisely the update used by Resonator Networks with OLS weights.
An important word of caution on this observation: it is somewhat of a misnomer to call this algorithm Bipolarized Alternating Least Squares, because at each iteration it is $\emph{not}$ solving a least squares problem, and this
conceals a profound difference.
To set $\mathbf{a}_f[t+1] = \mathbf{X}_f^\dagger \big(\hat{\mathbf{o}}^{(f)}[t] \odot \mathbf{c}\big)$
is to take the term $g(\mathbf{X}_f\mathbf{a}_f[t])$ present in the loss function and treat the activation function $g(\cdot)$ \emph{as if} it were linear, which it clearly is not. These updates are not computing a
Least Squares solution at each step. We actually lose the guarantee of global convergence that comes with Alternating Least Squares, but \emph{this is an exchange well worth making}, as we will show in the Results section.

Unlike Hopfield Networks, which have a Lyapunov function certifying their global asymptotic stability, no such function (that we know of)
exists for a Resonator Network. While $\hat{\mathbf{c}} = \mathbf{c}$ is always a fixed point of the OLS dynamics, a network initialized to a random state is not guaranteed to converge.
We have observed trajectories that collapse to limit cycles and seemingly-chaotic trajectories that do not converge in any reasonable time. One \emph{a priori} indication that this is the case
comes from a simple rewriting of two-factor Resonator Network dynamics that concatenates the states for each factor into a single statespace. To make the transformation exact, we appeal to the continuous-time version of Resonator
Networks, which, just like Hopfield networks, define dynamics in terms of time derivatives of the pre-activation state $\dot{\mathbf{u}}^{(f)}(t) = \mathbf{X}_f \mathbf{X}_f^\dagger \big(\hat{\mathbf{o}}^{(f)}[t] \odot \mathbf{c}\big)$,
with $\hat{\mathbf{x}}^{(f)}(t) = g(\mathbf{u}^{(f)}(t))$. We write down the continuous-time dynamics \`a la autoassociative Hopfield Networks:
\begin{equation*}
\begin{pmatrix}
    \vspace{4pt}
    \dot{\mathbf{u}}^{(1)}(t) \\
    \dot{\mathbf{u}}^{(2)}(t)
    \end{pmatrix} =  \renewcommand\arraystretch{1.2}
                     \begin{pmatrix}
                       \begin{array}{c|c}
                         \mathbf{0} &  \mathbf{X}_1 \mathbf{X}_1^\dagger \, \text{diag}(\mathbf{c}) \\
                         \hline
                         \mathbf{X}_2 \mathbf{X}_2^\dagger \, \text{diag}(\mathbf{c}) & \mathbf{0}
                       \end{array}
                     \end{pmatrix} \begin{pmatrix}
                                      \vspace{4pt}
                                      \hat{\mathbf{x}}^{(1)}(t) \\
                                      \hat{\mathbf{x}}^{(2)}(t)
                                  \end{pmatrix}
\end{equation*}
One can see that the weight matrix is non-symmetric, which has a simple but important consequence: autoassociative networks with non-symmetric weights cannot be guaranteed to converge \emph{in general}. This result, first established by
Cohen and Grossberg \citep{cohen1983absolute} and then studied throughout the Hopfield Network literature, is not quite as strong as it may sound, in the sense that symmetry is a sufficient, but not necessary,
condition for convergence. One can design a globally-convergent autoassociative network with asymmetric
weights \citep{xu1996asymmetric}, and moreover, adding a degree of asymmetry has been advocated as a technique to
reduce the influence of spurious fixed points \citep{hertz1986memory, singh1995fixed, chengxiang2000retrieval}.

Resonator Networks have a large and practical regime of operation, where $M$ (the problem size) is small enough, in which non-converging trajectories are extremely rare.
It is simple to deal with these events, making the model still useful in practice despite the lack of a convergence guarantee.
It has also been argued in several places
(see \citet{van1996chaos}, for example) that cyclic or chaotic trajectories may be useful to a neural system, including in cases where there are multiple plausible states to entertain. This is just to say that we feel the lack of a convergence guarantee is
not a critical weakness of our model, but rather an interesting and potentially useful characteristic. We attempted many different modifications to the model's dynamics which would provably cause it to converge, but these changes always hindered its
ability to solve the factorization problem. We emphasize that unlike all of the models in Section \ref{sec:benchmark_algs}, a Resonator Network is \emph{not} descending a loss function. Rather, it makes use of the fact that:
\begin{itemize}
    \item Each iteration is a bipolarized ALS update -- it \emph{approximately} moves the state towards the Least Squares solution for each factor.
    \item The correct solution is a fixed point (guaranteed for OLS weights, highly likely for OP weights).
    \item There may be a sizeable `basin of attraction' around this fixed point, which the iterates help us descend.
    \item The number of spurious fixed points (which do not give the correct factorization) is relatively small.
\end{itemize}
This last point is really what distinguishes Resonator Networks from the benchmarks, which we will establish in Section \ref{sec:explaining_operational_capacity}.

\newpage
\section{Results}
\label{sec:results}
We present a characterization of Resonator Networks along three main directions. The first direction is the stability of the solutions
$\mathbf{x}_\star^{(f)}$, which we relate to the stability of classical Hopfield networks. The second is a fundamental
measure of factorization capability we call the ``operational capacity''. The third is the speed with which factorizations are found.
We argue that the marked difference in factorization performance between our model and the benchmark algorithms lies in the relative
\emph{scarcity of spurious fixed points} enjoyed by Resonator Network dynamics. We summarize the main results
\textbf{in bold} throughout this section.

In each of the simulations we choose codevectors randomly i.i.d. from the discrete uniform distribution over the vertices
of the hypercube -- each element of each codevector is a Rademacher random variable (assuming the value $-1$ with probability $0.5$ and $+1$ with probability $0.5$). We generate $\mathbf{c}$ by choosing one vector at random from each of the $F$ codebooks and then computing the Hadamard product among these vectors. The reason we choose vectors randomly is because it makes the analysis of performance somewhat easier and more standardized, and it is the setting in which most of the well-known results on Hopfield Network capacity apply -- we will
make a few connections to these results. It is also the setting in which we typically use the Multiply, Add, Permute VSA architecture \citep{gayler2004vector} and therefore these results on random vectors are immediately applicable to a variety of existing works.

\subsection{Stable-solution capacity with outer product weights}
\label{sec:results_percolated_noise}
Suppose $\hat{\mathbf{x}}^{(f)}[0] = \mathbf{x}_\star^{(f)}$ for all $f$ (we initialize it to the correct factorization; this will also apply to any $t$ at which the algorithm comes upon $\mathbf{x}_\star^{(f)}$ on its own).
What is the probability that the state stays there -- i.e. that the correct factorization is a fixed point of the dynamics?
This is the basis of what researchers have called the ``capacity'' of Hopfield Networks, where $\mathbf{x}_\star^{(f)}$ are patterns that the network has been trained to store.
We choose to call it the ``stable-solution capacity'' in order to distinguish it from operational capacity, which we define
in Section \ref{sec:results_op_cap}.

We first note that this analysis is necessary only for Resonator Networks with outer product weights -- Ordinary Least Squares weights guarantee that the solutions are stable, and this is one of the variant's desirable properties.
If $\hat{\mathbf{x}}^{(f)}[0] = \mathbf{x}_\star^{(f)}$ for all $f$, then factor $1$ in a Resonator Network ``sees'' an input $\mathbf{x}_\star^{(1)}$ at time $t=1$.
For OLS weights, the vector $\mathbf{X}_1 \mathbf{X}_1^\dagger \mathbf{x}_\star^{(1)}$ is exactly $\mathbf{x}_\star^{(1)}$ by the definition of orthogonal projection. True for all subsequent factors, this means that for OLS weights, $\mathbf{x}_\star^{(f)}$ is always a fixed point.

For a Resonator Network with outer product weights, we must examine the vector
$\bm{\Gamma} := \OPsynapses \big(\hat{\mathbf{o}}^{(f)}[0] \odot \mathbf{c}\big)$ at each $f$, and changing from the psuedoinverse $\mathbf{X}_f^\dagger$ to the transpose $\mathbf{X}_f^\top$ makes the situation significantly more
complicated. At issue is the probability that $\Gamma_i$ has a sign different from ${\big(\mathbf{x}_\star^{(f)}\big)}_i\,$, i.e. that there is a bitflip in any particular component of the updated state.
In general one may not care whether the state is completely stable -- it may be tolerable that the dynamics flip some small fraction of the bits of $\mathbf{x}_\star^{(f)}$ as long as it does not move the state too far away from $\mathbf{x}_\star^{(f)}$. Amit, Gutfreund, and Sompolinsky \citep{amit1985storing, amit1987information} established that in Hopfield Networks, an avalanche phenomenon occurs where bitflips accumulate and the network becomes essentially useless
for values of $D_f > 0.138N$, at which point the approximate bitflip probability is $0.0036$.
While we don't attempt any of this complicated analysis on Resonator Networks, we do derive an expression for the
bitflip probability of any particular factor that accounts for bitflips which ``percolate''
from factor to factor through the vector $\hat{\mathbf{o}}^{(f)}[0] \odot \mathbf{c}$.

We start by noting that for factor $1$, this bitflip probability is the same as
a Hopfield network. Readers familiar with the literature on Hopfield Networks will know that with $N$ and $D_f$ reasonably large (approximately $N\geq1{,}000$ and $D_f \geq 50$)
$\Gamma_i$ can be well-approximated by a Gaussian with mean ${\big(\mathbf{x}_\star^{(f)}\big)}_i \, (N + D_f - 1)$
and variance $(N-1)(D_f-1)$; see appendix \ref{appendix:stable_memory_cap} for a simple derivation. This is summarized as the \emph{Hopfield bitflip probability} $h_f$:
\begin{align*}
   \label{eq:hopfield_bitflip}
   h_f :=& \,\, Pr \big[\, {\big(\hat{\mathbf{x}}^{(f)}[1]\big)}_i \neq {\big(\mathbf{x}_\star^{(f)}\big)}_i \, \big] \\
   =& \,\, \Phi\Big(\frac{-N-D_f+1}{\sqrt{(N-1)(D_f-1)}} \Big) \numberthis
\end{align*}
Where $\Phi$ is the cumulative density function of the Normal distribution. Hopfield Networks are often specified with the diagonal of $\OPsynapses$ set to all zeros (having ``no self-connections''), in which case the bitflip probability is
$\Phi\Big(\frac{-N}{\sqrt{(N-1)(D_f-1)}} \Big)$. For large $N$ and $D_f$ this is often simplified to $\Phi(-\sqrt{N/D_f})$, which may be the expression most familiar to readers. Keeping the diagonal of $\OPsynapses$ makes the codevectors more stable
(see appendix \ref{appendix:stable_memory_cap}) and while there are some arguments in favor of eliminating it, we have found Resonator Networks to exhibit better performance by keeping these terms.

In Appendix \ref{appendix:stable_memory_cap} we derive the bitflip probability for an arbitrary factor in a Resonator Network with outer product weights.
This probability
depends on whether a component of the state has already been flipped by the previous $f-1$ factors, which is what we call \emph{percolated noise} passed between the factors, and which increases the bitflip probability. The four relevant probabilities are:
\begin{equation}
    \label{eq:bitflip}
    r_f := Pr \big[\, \big(\hat{\mathbf{x}}^{(f)}[1]\big)_i \neq \big(\mathbf{x}_\star^{(f)}\big)_i \, \big]
\end{equation}
\begin{equation}
    \label{eq:net_bitflip}
    n_f := Pr \big[\, \big(\hat{\mathbf{o}}^{(f+1)}[0] \odot \mathbf{c}\big)_i \neq \big(\mathbf{x}_\star^{(f+1)}\big)_i \, \big]
\end{equation}
\begin{equation}
    \label{eq:conditional_bitflip_1}
    r_{f^\prime} := Pr \big[\, \big(\hat{\mathbf{x}}^{(f)}[1]\big)_i \neq \big(\mathbf{x}_\star^{(f)}\big)_i \,\, \big| \,\, \big(\hat{\mathbf{o}}^{(f)}[0] \odot \mathbf{c}\big)_i = \big(\mathbf{x}_\star^{(f)}\big)_i \, \big]
\end{equation}
\begin{equation}
    \label{eq:conditional_bitflip_2}
    r_{f^{\prime\prime}} := Pr \big[\, \big(\hat{\mathbf{x}}^{(f)}[1]\big)_i \neq \big(\mathbf{x}_\star^{(f)}\big)_i \,\, \big| \,\, \big(\hat{\mathbf{o}}^{(f)}[0] \odot \mathbf{c}\big)_i \neq \big(\mathbf{x}_\star^{(f)}\big)_i \, \big]
\end{equation}
Equation (\ref{eq:bitflip}) is the probability of a bitflip compared to the correct value, the \emph{Resonator bitflip probability}. Equation (\ref{eq:net_bitflip}) gives the probability that the \emph{next} factor will see a net bitflip, a bitflip which has percolated through the previous factors.
Equations (\ref{eq:conditional_bitflip_1}) and (\ref{eq:conditional_bitflip_2}) give the probability of a bitflip conditioned on whether or not this factor sees a net bitflip, and they are \emph{different}.
It should be obvious that
\begin{equation}
    \label{eq:bitflip_in_terms_of_conditionals}
    r_f = r_{f^\prime}(1 - n_{f-1}) + r_{f^{\prime\prime}} n_{f-1}
\end{equation}
and also that
\begin{equation}
    \label{eq:net_bitflip_in_terms_of_conditionals}
    n_f = r_{f^\prime}(1 - n_{f-1}) + (1 - r_{f^{\prime\prime}}) n_{f-1}
\end{equation}
We show via straightforward algebra in Appendix \ref{appendix:stable_memory_cap} that the conditional probabilities $r_{f^\prime}$ and $r_{f^{\prime\prime}}$ can be written recursively in terms of $n_f$:
\begin{equation}
    \label{eq:conditional_bitflip_1.2}
    r_{f^{\prime}} = \Phi\Big(\frac{-N(1-2n_{f-1})-(D_f-1)}{\sqrt{(N-1)(D_f-1)}} \Big)
\end{equation}
\begin{equation}
    \label{eq:conditional_bitflip_2.2}
    r_{f^{\prime\prime}} = \Phi\Big(\frac{-N(1-2n_{f-1})+(D_f-1)}{\sqrt{(N-1)(D_f-1)}} \Big)
\end{equation}
The Resonator bitflip probability $r_f$ has to be computed recursively using these expressions. The base case is $n_0 = 0$ and this is sufficient to compute all the other probabilities -- in particular, it implies that
$r_1 = h_1 = \Phi\big(\frac{-N-D_1+1}{\sqrt{(N-1)(D_1-1)}} \big)$, which we have previously indicated.
We can verify these equations in simulation, and the agreement is very good -- see Figure \ref{fig:appendix_percolated_noise_empirical_bitflip_prob} in the Appendix, which measures $r_f$.

\textbf{The main analytical result in this section is the sequence of equations (\ref{eq:bitflip_in_terms_of_conditionals}) - (\ref{eq:conditional_bitflip_2.2}), which allow one to compute the bitflip probabilities for
each factor in an outer product Resonator Network}.
The fact that $r_f$ in general must be split between the two conditional probabilities and that there is a dependence on $n_{f-1}$ is what makes it different, for all but the first factor, from the bitflip
probability for a Hopfield Network (compare eqs. (\ref{eq:conditional_bitflip_1.2}) and (\ref{eq:conditional_bitflip_2.2}) against eq. (\ref{eq:hopfield_bitflip})). But how much different? We are interested in the quantity $r_f - h_f$.

Here is a simple intuition for what this is capturing: suppose there are $F$ Hopfield Networks all evolving under their own dynamics -- they are running simultaneously but not interacting in any way. At time $t=0$, the bitflip probabilities
$h_1, h_2, \ldots, h_F$ for the networks are all the same; there is nothing special about any particular one. A Resonator Network, however, is like a set of $F$ Hopfield networks that have been wired up to receive input
$\hat{\mathbf{o}}^{(f)}[t] \odot \mathbf{c}$, which reflects the state of the \emph{other} factors. The networks are no longer independent. In particular, a bitflip in factor $f$ gets passed onto factors $f+1$, $f+2$, and so on.
This affects the bitflip probability of these other factors, and the magnitude of this effect, which we call percolated noise, is measured by $r_f - h_f$.

\begin{figure}
    \centering
    \includegraphics[width=\textwidth]{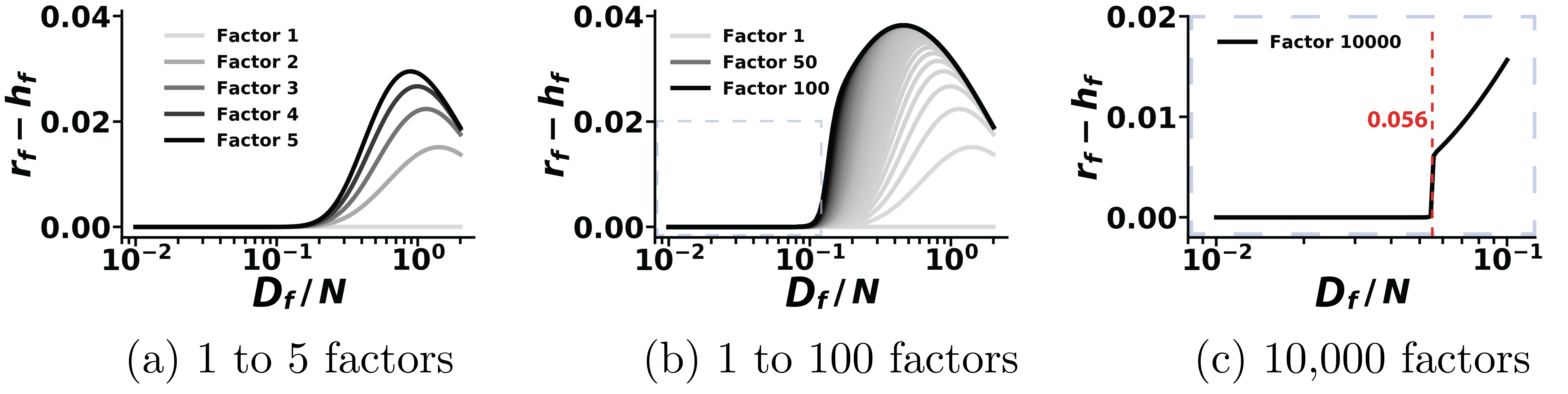}
    \caption{Extra bitflip probability $r_f - h_f$ due to percolated noise. In the limit of large $F$, there appears to be a phase change at $D_f / N = 0.056$. Below this value Resonator Networks are just as stable as Hopfield Networks, but above this value they are strictly less stable (by the amount $r_f - h_f$).}
    \label{fig:percolated_noise_diff_3part}
\end{figure}

Let us first note that for a Hopfield network \emph{with self connections} the maximum bitflip probability is $0.02275$, which occurs at $D_f = N$. The ratio $D_f / N$ is
what determines the bitflip probability. Please see Appendix \ref{appendix:stable_memory_cap} for an explanation. Percolated noise is measured by the difference $r_f - h_f$, which we plot in
Figure \ref{fig:percolated_noise_diff_3part}. Part (a) shows just five factors, illustrating that $r_1 = h_1$, but that $r_f \geq h_f$ in general.
To see if there is some limiting behavior, we simulated $100$ and $10{,}000$ factors; the differences $r_f - h_f$ are also shown in Figure \ref{fig:percolated_noise_diff_3part}.
In the limit of large $F$ there appears to be a phase change in residual bitflip probability that occurs at
$D_f / N = 0.056$. In the Hopfield Network literature this is a very important number. It gives the point at which the codevectors transition away
from being global minimizers of the Hopfield Network energy function. When $D_f / N$ falls in between $0.056$ and $0.138$, the codevectors are only local minimizers, and there exist \emph{spin-glass} states that have lower energy.
We do not further explore this phase-change phenomenon, but leave the (in all likelihood, highly technical) analysis to future work.

In conclusion, the second major result of the section is that we have shown, via simulation, that \textbf{for $\mathbf{D_f / N \leq 0.056}$, the stability of a Resonator Network with outer product weights is the same as the stability
of a Hopfield Network. For $\mathbf{D_f / N > 0.056}$, percolated noise between the factors causes the Resonator Network to be strictly less stable than a Hopfield Network}.

\subsection{Operational capacity}
\label{sec:results_op_cap}
We now define a new notion of capacity that is more appropriate to the factorization problem. This performance measure, called the \emph{operational capacity},
gives an expression for the maximum size of factorization problem that can be solved with high probability. This maximum problem size,
which we denote by $M_{max}$, varies as a function of the number of elements in each vector $N$ and the number of factors $F$.
It gives a very practical characterization of performance, and will form the basis of our comparison between Resonator Networks and
the benchmark algorithms we introduced in Section \ref{sec:benchmark_algs}.
When the problem size $M$ is below the operational capacity of the algorithm, one can be quite sure that the correct
factorization will be efficiently found.

\theoremstyle{definition}
\newtheorem{definition}{Definition}
\theoremstyle{definition}
\begin{definition}
\label{def:operational_capacity}
The $\{p, k \}$ operational capacity of a factorization algorithm that solves (\ref{eq:general_factorization}) is the largest
search space size $M_{max}$ such that the algorithm, when limited to a maximum number of iterations $k$, gives a total accuracy $\geq p$.
\end{definition}

We now define what we mean by total accuracy.
Each algorithm we have introduced attempts to solve the factorization problem (\ref{eq:general_factorization}) by
initializing the state $\hat{\mathbf{x}}^{(f)}[0]$ and letting the dynamics evolve until some termination criterion
is met. It is possible that the final state $\hat{\mathbf{x}}^{(f)}[\infty]$ may not equal the correct factors $\mathbf{x}_{\star}^{(f)}$
at each and every component, but we can `decode' each $\hat{\mathbf{x}}^{(f)}[\infty]$ by looking for its nearest neighbor
(with respect to Hamming distance or cosine similarity) among the vectors in its respective codebook $\mathbb{X}_f$.
This distance computation involves only $D_f$ vectors, rather than $M$, which was what we
encountered in one of the brute-force strategies of Section \ref{sec:problem_statement}. Compared to the other computations
involved in finding the correct factorization out of $M$ total possibilities, this last step of decoding has a very small cost, and
we always `clean up' the final state $\hat{\mathbf{x}}^{(f)}[\infty]$ using its nearest neighbor in the codebook.
We define the total accuracy to be the sum of accuracies for inferring each
factor, which is $1/F$ if the nearest neighbor to $\hat{\mathbf{x}}^{(f)}$ is $\mathbf{x}_{\star}^{(f)}$ and $0$ otherwise.
For instance, correctly inferring one of three total factors gives a total accuracy of $1/3$, two of three is $2/3$, and three
of three is $1$.

Analytically deriving the expected total accuracy appears to be quite challenging, especially for a Resonator Network, because it requires that we
essentially predict how the nonlinear dynamics will evolve over time. There may be a region around each $\mathbf{x}_{\star}^{(f)}$
such that states in this region rapidly converge to $\mathbf{x}_{\star}^{(f)}$, the so-called basin of attraction,
but our initial estimate $\hat{\mathbf{x}}_{(f)}[0]$ is likely not in the basin of attraction, and it is
hard to predict when, if ever, the dynamics will enter this region. Even for Hopfield Networks, which obey much simpler dynamics
than a Resonator Network, it is known that so-called ``frozen noise'' is built up in the network state, making the shapes of
the basins highly anisotropic and difficult to analyze \citep{amari1988statistical}.
Essentially all of the analytical results on Hopfield Networks consider only the stability of
$\mathbf{x}_{\star}^{(f)}$ as a (very poor) proxy for how the model behaves when it is initialized to other states.
This less useful notion of capacity, the stable-solution capacity, was what we examined in the previous section.

We can, however, estimate the total accuracy by simulating many factorization problems, recording the
fraction of factors that were correctly inferred over many, many trials. We remind the reader that our results in this paper
pertain to factorization of randomly-drawn vectors which bear no particular correlational structure, but that notions of total accuracy
and operational capacity would be relevant, and specific, to factorization of non-random vectors.
\begin{figure}[t]
    \centering
    \includegraphics[width=0.5\textwidth]{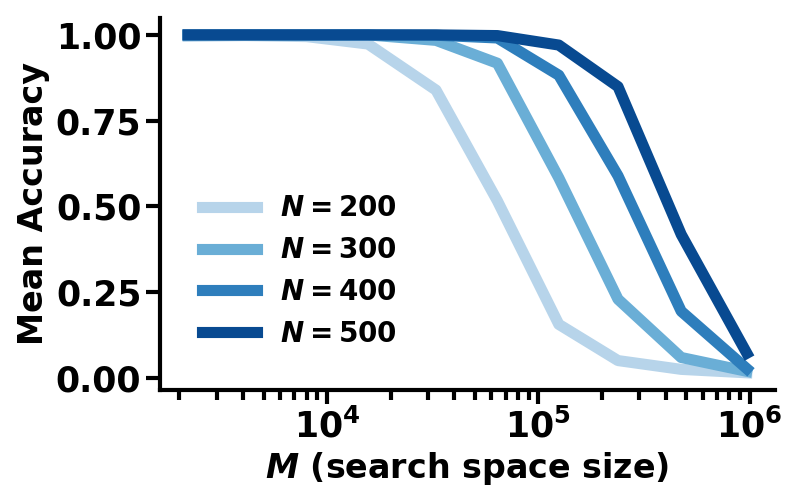}
    \caption{Accuracy as a function of $M$ for Resonator Network with outer product weights. Three factors ($F=3$), average over $5{,}000$ random trials.}
    \label{fig:total_accuracy_func_of_m}
\end{figure}
We first note that for fixed vector dimensionality $N$, the empirical mean of the total accuracy depends strongly on $M$,
the search space size. We can see this clearly
in Figure \ref{fig:total_accuracy_func_of_m}. We show this phenomenon for a Resonator Network with outer product weights, but
this general behavior is true for all of the algorithms under consideration -- one can always make the search
space large enough that expected total accuracy goes to zero.

Our notion of operational capacity is concerned with the $M$ that causes expected total accuracy to drop below some value $p$. We see here
that there are a range of values $M$ for which the expected total accuracy is $1.0$, beyond which this ceases to be the case. For all
values of $M$ within this range, the algorithm essentially always solves the factorization problem.

In this paper we estimate operational capacity when $p=0.99$ ($\geq 99\%$ of factors were inferred correctly) and $k=0.001M$
(the model can search over at most $1/1{,}000$ of the entire search space). These choices are largely practical: $\geq 99\%$ accuracy
makes the model very reliable in practice, and this operating point can be estimated from a reasonable number ($3{,}000$ to $5{,}000$)
of random trials. Setting $k=0.001M$ allows the number of iterations to scale with the size of the problem, but restricts the
algorithm to only consider a small fraction of the possible factorizations. While a Resonator Network has no guarantee of convergence,
it almost always converges in far less than $0.001M$ iterations, so long as we stay in this high-accuracy regime. Operational capacity is in general a function of
$N$ and $F$, which we will discuss shortly.

\subsubsection{Resonator Networks have superior operational capacity}
\label{sec:results_capacity_comparison}

\begin{figure}[t]
    \centering
    \includegraphics[width=0.9\textwidth]{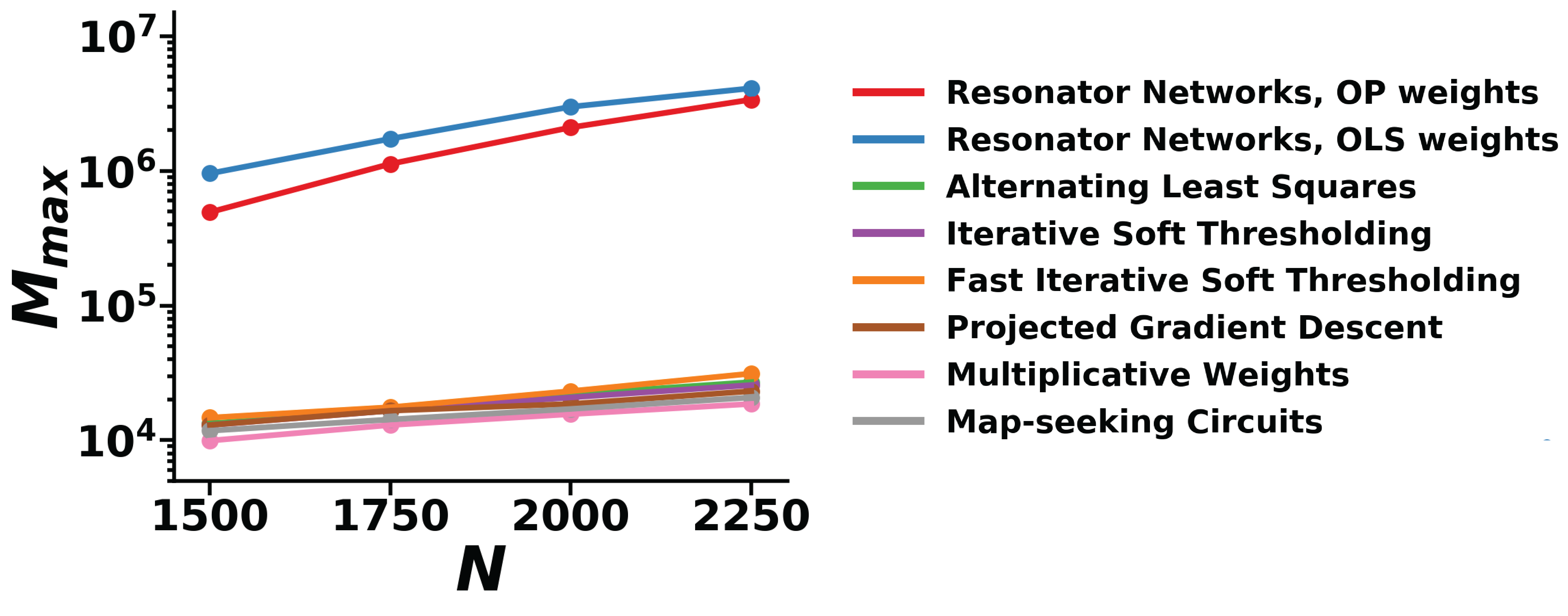}
    \caption{Operational capacity is dramatically higher for Resonator Networks (blue and red above) than for any of the benchmark algorithms.
             These points represent the size of factorization problem that can be solved reliably. Shown is operational capacity for $F=3$ factors.
             The gap is similarly large for other $F$, see plot for $F=4$ in the Appendix.}
    \label{fig:op_cap_comparison}
\end{figure}

We estimated the operational capacity of the benchmark algorithms in addition to the two variants Resonator Networks.
Figure \ref{fig:op_cap_comparison} shows the operational capacity estimated on several thousand random
trials, where we display $M_{max}$ as a function of $N$ for problems with three factors. One can see that
\textbf{the operational capacity of Resonator Networks is roughly two orders of magnitude greater than the operational
capacity of the other algorithms}. Each of the benchmark algorithms has a slightly different operational capacity
(due to the fact that they each have different dynamics and will, in general, find different solutions) but they are all similarly
poor compared to the two variants of Resonator Networks. See a similar plot for $F=4$ in Appendix \ref{appendix:op_cap}.

As $N$ increases, the performance difference between the two variants of Resonator Networks starts to
disappear, ostensibly due to the fact that ${\mathbf{X}_f\mathbf{X}_f^\dagger \approx \mathbf{X}_f \mathbf{X}_f^\top}$.
The two variants are different in general, but the simulations in this paper do not particularly highlight the difference between
them. Except for Alternating Least Squares, each of the benchmark algorithms has at least one hyperperparameter that must be chosen -- we simulated many thousand
random trials with a variety of hyperparameter settings for each algorithm and chose the hyperparameter
values that performed best on average. We list these values for each of the algorithms in the Appendix. All of the
benchmark algorithms converge on their own and the tunable stepsizes make a comparison of the number of iterations non-standardized, so we did not impose a maximum number of iterations on these algorithms -- the points
shown represent the best the benchmark algorithms can do, even when not restricted to a maximum number of iterations.

\subsubsection{Operational capacity scales quadratically in $N$}
\label{sec:results_op_cap_scaling_NF}
We carefully measured the operational capacity of Resonator Networks in search of a relationship between $M_{max}$ and $N$. We focused on Resonator Networks with outer product weights -- for $N \approx 5000$ and larger,
randomly-chosen codevectors are nearly orthogonal and capacity is approximately the same for OLS weights. We reiterate that operational capacity is specific to parameters
$p$ and $k$: $p$ is the threshold for total accuracy and $k$ is the maximum number of iterations the algorithm is allowed to take (refer to Definition \ref{def:operational_capacity}). Here we report operational capacity for $p=0.99$
and $k=0.001M$ on randomly-sampled codevectors. The operational capacity is specific to these choices, which are practical for Vector Symbolic Architectures.

\begin{figure}[t!]
  \centering
  \begin{subfigure}{\textwidth}
      \centering
      \includegraphics[width=\textwidth]{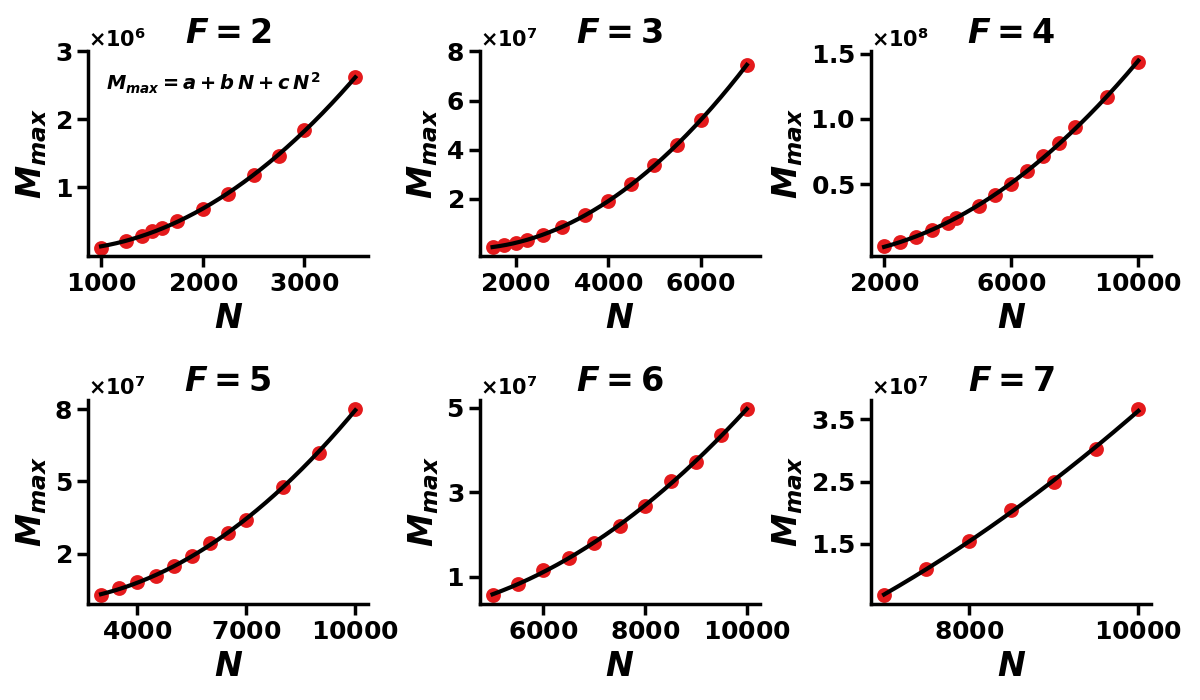}
      \caption{$M_{max}$ scales quadratically in $N$. Red points are measured from simulation; black curves are the least-squares quadratic fits.
               Parameters of fits included in Appendix \ref{appendix:op_cap}.}
      \label{fig:op_cap_quad}
  \end{subfigure}
  \par\bigskip
  \begin{subfigure}{\textwidth}
      \centering
      \includegraphics[width=0.5\textwidth]{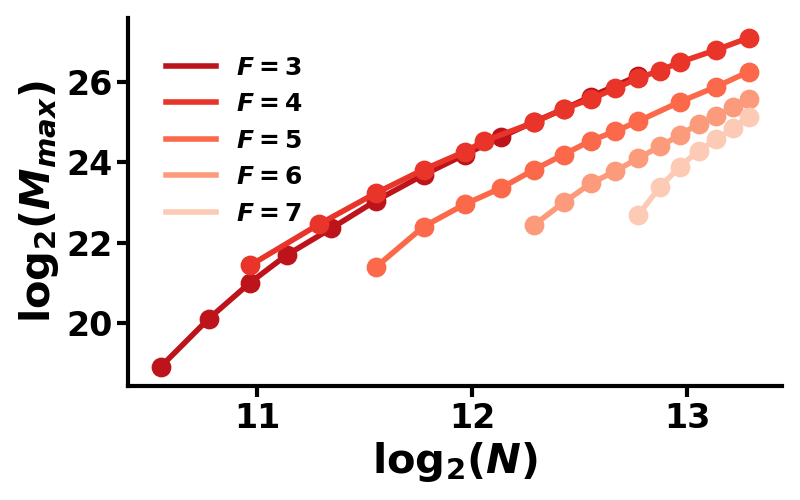}
      \caption{$M_{max}$ varies as a function of both $F$ and $N$. Over the measured range for $N$, capacity is highest for $F=3$ and $F=4$. Data for $F=2$ was omitted to better convey the trend for F=3 and higher, but see Appendix \ref{appendix:op_cap} for the full picture.}
      \label{fig:op_cap_loglog}
  \end{subfigure}
  \caption{Operational capacity of Resonator Networks with OP weights.}
  \label{fig:op_cap_mega}
\end{figure}

Our simulations revealed that, empirically, \textbf{Resonator Network operational capacity $\mathbf{M_{max}}$ scales as a quadratic function of $\mathbf{N}$}, which we illustrate in Figure \ref{fig:op_cap_mega}. The points in this figure are estimated from many thousands of random trials, over a range
of values for $F$ and $N$. In part (a) we show operational capacity separately for each $F$ from $2$ to $7$, with the drawn curves indicating the least-squares quadratic fit to the measured points. In part (b)
we put these points on the same plot, following a logarithmic transformation to each axis, in order to illustrate that capacity also varies as a function of $F$.
Appendix \ref{appendix:op_cap} provides some additional commentary on this topic, including some speculation on a scaling law that combines $F$ and $N$.
The parameters of this particular combined scaling are estimated from simulation and not derived analytically -- therefore they may deserve additional scrutiny and we do not
focus on them here. The main message of this section is that capacity scales quadratically in $N$, regardless of how many factors are used.

The curves in Figure \ref{fig:op_cap_mega} are constructive in the following sense: given a fixed $N$, they indicate the largest factorization problem that can be solved reliably.
Conversely, and this is often the case in VSAs, the problem
size $M$ is predetermined, while $N$ is variable -- in this case we know how large one must make $N$. We include in the official software implementation that accompanies this paper\footnote{\normalsize \href{https://github.com/spencerkent/resonator-networks}{https://github.com/spencerkent/resonator-networks}} a text file with all of the measured operational capacities.

Quadratic scaling means that one can aspire to solve very large factorization problems, so long as he or she can build a Resonator Network with big enough $N$.
We attempted to estimate capacity for even larger values of $N$ than we report in Figure \ref{fig:op_cap_mega}, but this was beyond the capability of our current
computational resources. A useful contribution of follow-on work would be to leverage high-performance computing to measure some of these values.
Applications of Vector Symbolic Architectures typically use $N \leq 10{,}000$, but there are other reasons one might attempt to push Resonator Networks further.
Early work on Hopfield Networks proposed a technique for storing solutions to the Travelling Salesman Problem as fixed points of the model's dynamics \citep{hopfield1985neural}, and this became part of a
larger approach using nonlinear dynamical systems to solve hard search problems. We do not claim that any particular search problem, other than the factorization we have defined (\ref{eq:general_factorization}),
can be solved by Resonator Networks. Supposing, however, that some other hard problem can be cast in the form of (\ref{eq:general_factorization}), the quadratic scaling of operational capacity makes this a potentially
power tool.

Capacity is highest when the codebooks $\mathbb{X}_f$ each have the same number of codevectors ($D_1 = D_2 = \ldots = D_F = \sqrt[F]{M}$),
and this was the case for the operational capacity results we have shown so far. We chose this in order to have a simple standard for
comparison among the different algorithms, but in general it is possible that the codebooks are unbalanced, so that
we have the same $M = \prod_f D_f$ but $D_1 \neq D_2 \neq \ldots \neq D_f$. In this case, capacity is lower than for balanced codebooks.
We found that the most meaningful way to measure the degree of balance between codebooks was by the ratio of the smallest codebook to the
largest codebook:
\begin{equation}
    \xi := \big(\underset{f}{\min}\, D_f\big) \, / \, \big(\underset{f}{\max} \, D_f\big)
\end{equation}
For $\xi \geq 0.2$ we found that the effect on $M_{max}$ was simply an additive factor which can be absorbed into a
(slightly smaller) y-intercept $a$ for the quadratic fit. For extreme values of $\xi$, where there is one codebook that is
for instance $10$ or $20$ times larger than another, then all three parameters $a$, $b$, and $c$ are affected, sometimes significantly.
Scaling is still quadratic, but the actual capacity values may be significantly reduced.

Our result -- measured operational capacity which indicates an approximately quadratic relationship between $M_{max}$ and $N$ -- is
an important characterization of Resonator Networks. It suggests that our framework scales to very large factorization problems and serves as
a guideline for implementation. Our attempts to analytically derive this result were stymied by the toolbox of nonlinear dynamical
systems theory. Operational capacity involves the probability that this system, when initialized to an effectively random state, converges to
a particular set of fixed points. No results from the study of nonlinear dynamical systems, that we are aware of, allow us to derive
such a strong statement about Resonator Networks. Still, the scaling of Figure \ref{fig:op_cap_mega} is fairly suggestive of some underlying
law, and we are hopeful that a theoretical explanation exists, waiting to be discovered.

\subsection{Search speed}
\label{sec:results_search_efficiency}
\begin{figure}[t]
    \centering
    \includegraphics[width=0.5\textwidth]{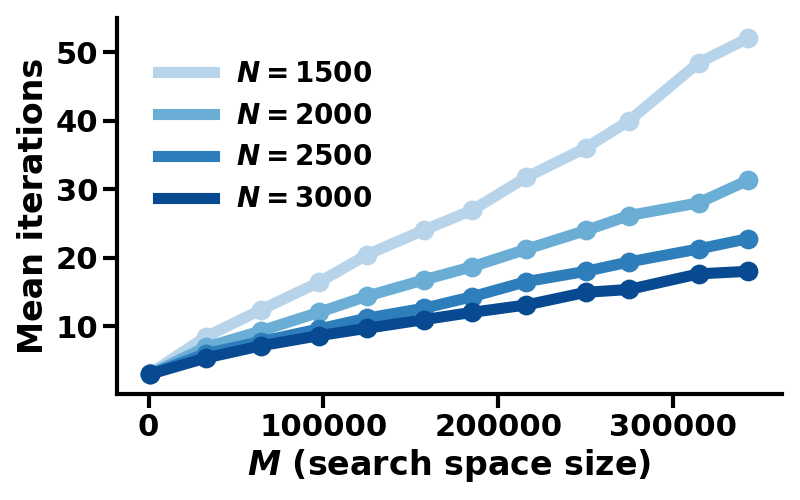}
    \caption{Iterations until convergence, Resonator Network with outer product weights and $F=3$. The number of iterations is a very small compared to the size of the search space}
    \label{fig:num_iters_resonator}
\end{figure}

If a Resonator Network is not consistently descending an energy function, is it just aimlessly wandering around the space, trying every possible factorization until it finds the correct one? Figure \ref{fig:num_iters_resonator} shows that it is not.
We plot the mean number of iterations over $5{,}000$ random trials, as a fraction of $M$, the search space size. This particular plot is based on a Resonator Network with outer product weights and $F=3$. In the high-performance regime where $M$ is below operation capacity,
the number of iterations is far less than the $0.001M$ cutoff we used in the simulations of Section \ref{sec:results_op_cap} -- the algorithm is
only ever considering a tiny fraction of the possible factorizations before it finds the solution.

\begin{figure}[t!]
    \centering
    \begin{subfigure}[]{\textwidth}
        \includegraphics[width=0.98\textwidth]{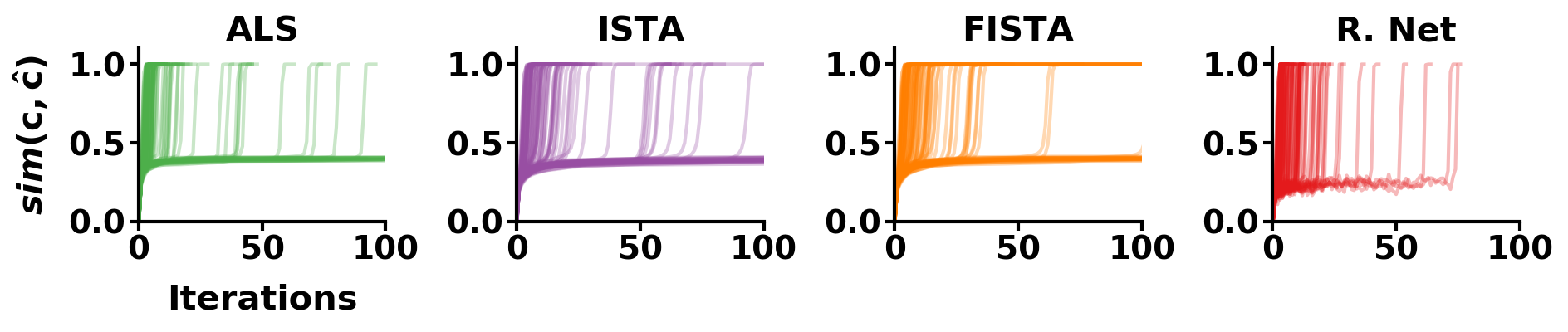}
        \caption{Convergence traces for 100 randomly-drawn factorization problems -- each line is the cosine similarity
                 between $\mathbf{c}$ and $\hat{\mathbf{c}}$ over iterations of the algorithm. Each of the four algorithms is run on \emph{the same} $100$ factorization problems. All of the instances are solved by the Resonator Network,
                 whereas a sizeable fraction (around $30\%$) of the instances are not solved by the benchmark algorithms,
                 at least within $100$ iterations.}
        \label{fig:num_iters_indv_trials}
    \end{subfigure}
    \begin{subfigure}[]{0.40\textwidth}
        \centering
        \vspace{2ex}
        \includegraphics[width=\textwidth]{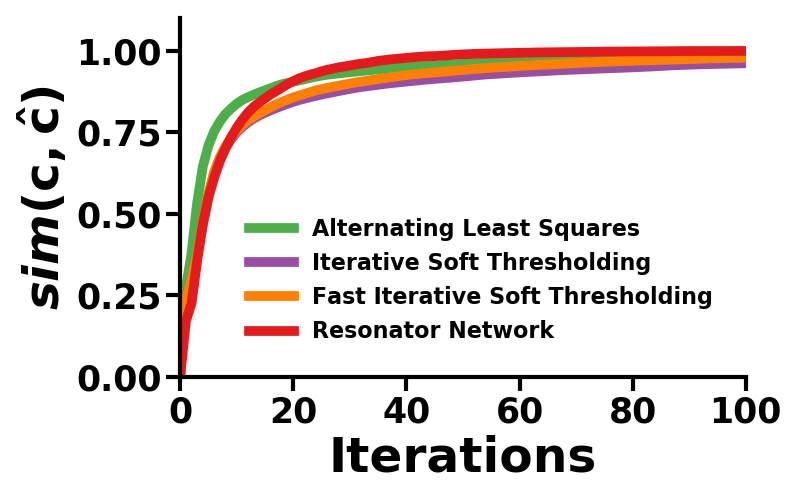}
        \caption{Cosine sim. vs. iteration \# (only trials with accuracy $1.0$)}
        \label{fig:avg_sim_vs_iters}
    \end{subfigure}
    \hspace{0.075\textwidth}
    \begin{subfigure}[]{0.40\textwidth}
        \centering
        \vspace{2ex}
        \includegraphics[width=\textwidth]{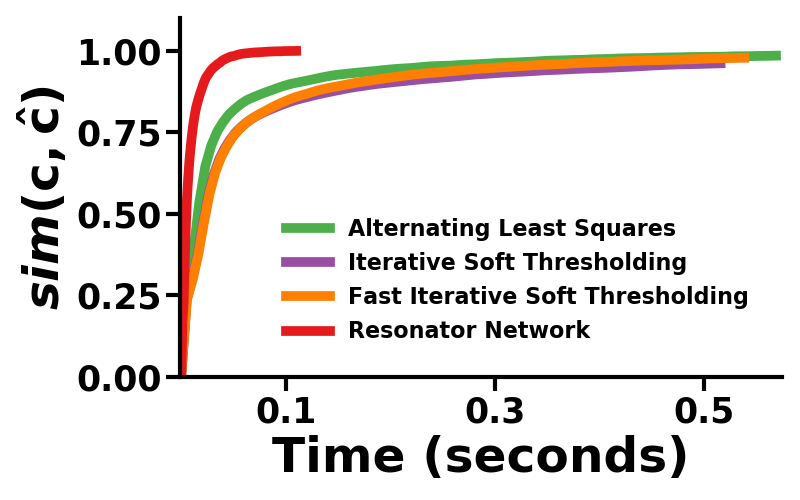}
        \caption{Cosine sim. vs. wall-clock time (only trials with accuracy $1.0$)}
        \label{fig:avg_sim_vs_time}
    \end{subfigure}
    \vspace{2ex}
    \caption{Our benchmark of factorization speed. Implementation in Python with NumPy. Run on machine with Intel Core i7-6850k processor and 32GB RAM. We generated $5,000$ random instantiations of the factorization problem with $N=1500$, $F=3$, and $D_f=40$, running each of the four algorithms in turn. Figure \ref{fig:num_iters_indv_trials} gives a snapshot of $100$ randomly selected trials.
    Figures \ref{fig:avg_sim_vs_iters} and \ref{fig:avg_sim_vs_time} show average performance \emph{conditioned on the algorithms finding the correct factorization}.}
    \label{fig:speed_compared_to_benchmarks}
\end{figure}

Section \ref{sec:results_capacity_comparison} compared the operational capacity of different algorithms and showed that compared to the benchmarks, Resonator Networks can solve much larger factorization problems. This is in the sense that the dynamics
eventually converge (with high probability) on the correct factorization while, the dynamics of the other algorithms converge on spurious
factorizations. This result, however, does not directly demonstrate the relative speed with which factorization are found in terms of either the
number of iterations or the amount of time to convergence. We set up a benchmark to determine the
relative speed of Resonator Networks and our main finding is depicted in Figure \ref{fig:speed_compared_to_benchmarks}.

\textbf{Measured in number of iterations, Resonator Networks are comparable to the benchmark algorithms}. We noted that Alternating Least Squares is the most greedy of the benchmarks, and one can
see from Figure \ref{fig:speed_compared_to_benchmarks} that it is the fastest in this sense. We are considering only trials that ultimately found the correct factorization, which in
this simulation was roughly $70\%$ for each of the benchmarks. In contrast, Resonator Networks always eventually found the correct factorization.
\textbf{Measured in terms of wall-clock time, Resonator Networks are significantly faster than the benchmarks.} This can be attributed to their nearly $5\times$
lower per-iteration cost. Resonator Networks with outer product weights utilize very simple
arithmetic operations and this explains the difference between Figures \ref{fig:avg_sim_vs_iters} and \ref{fig:avg_sim_vs_time}.

\newpage
\subsection{Dynamics that do not converge}
One must be prepared for the possibility that the dynamics of a Resonator Network will not converge. Fortunately, for $M$ below the $p=0.99$ operational capacity, these will be exceedingly rare. From simulation,
we identified three major regimes of different convergence behavior, which are depicted in Figure \ref{fig:non_conv_regime}:
\begin{figure}[t]
    \centering
    \includegraphics[width=\textwidth]{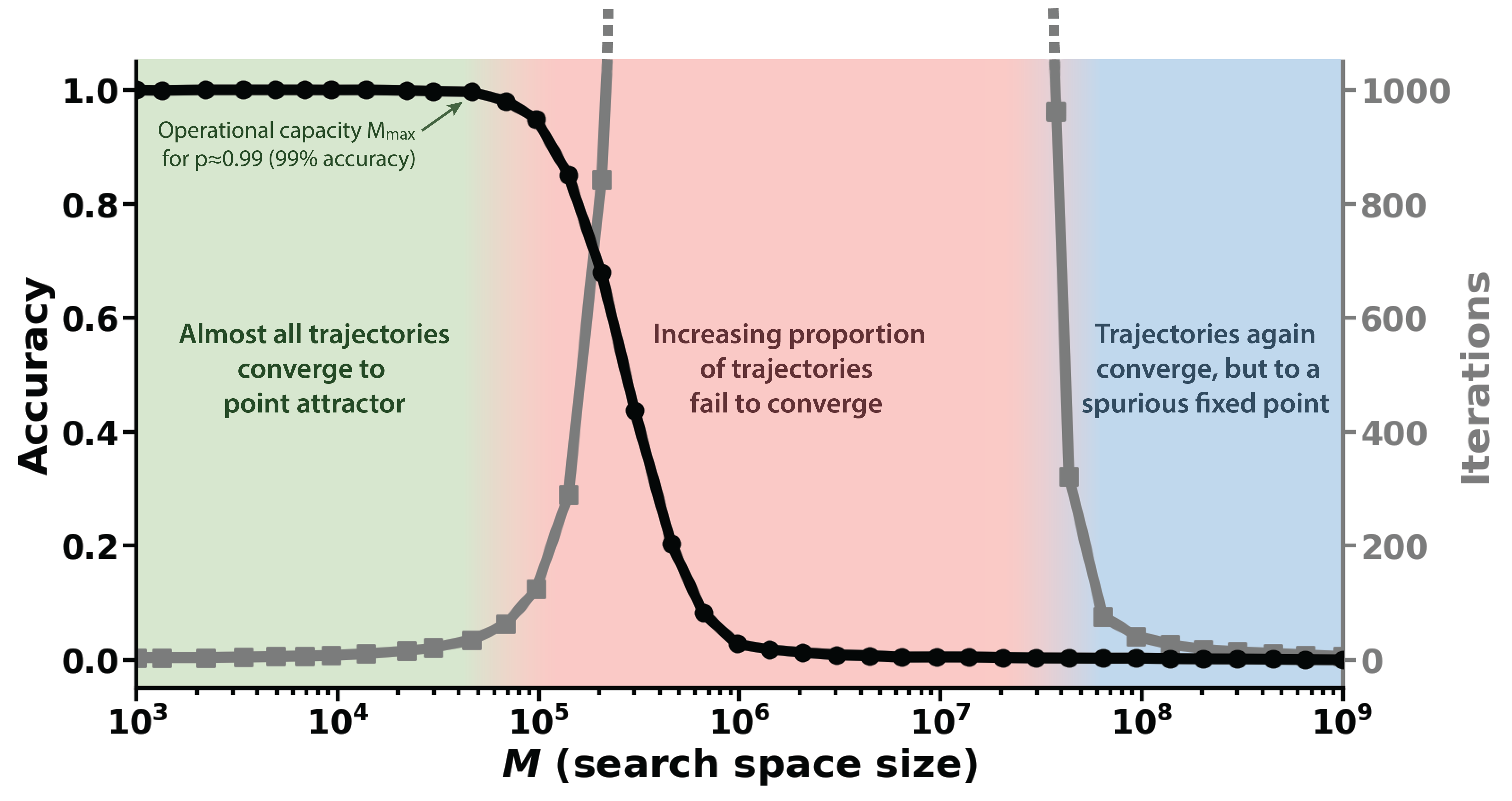}
    \caption{Regimes of different convergence behavior. Curves show measurement from simulation of an outer product Resonator Network with $3$ factors and $N=400$. This is also meant as a diagram of convergence behavior for Resonator Networks in general.
             Shown in black is the average decoding accuracy and shown in gray is the median number of iterations taken by the network.
             For low enough $M$, the network always finds a fixed point yielding $100\%$ accuracy. The network will not converge to spurious fixed points in this regime (green). As $M$ is increased, more trajectories wander, not converging in any reasonable time (red).
             Those that are forcibly terminated yield incorrect factorizations. For large enough $M$, the network is completely saturated and most states are fixed points, regardless of whether they yield the correct factorization (blue). Resonator Networks with
             OLS weights are always stable when $D_f=N$, but OP weights give a bitflip probability that is zero only asymptotically in $M$ (see Section \ref{sec:results_percolated_noise} and Appendix \ref{appendix:stable_memory_cap}).}
    \label{fig:non_conv_regime}
\end{figure}
\begin{itemize}
    \item For $M$ small enough, almost all trajectories converge, and moreover they converge to a state that yields the correct factorization. Limit cycles are possible but rare, and often still yield
    the correct factorization. There appear to be few if any spurious fixed points (those yielding an incorrect factorization) -- if the trajectory converges to a point attractor or limit cycle, one can be confident
    this state indicates the correct factorization.
    \item As $M$ increases, non-converging trajectories appear in greater proportion and yield incorrect factorizations. Any trajectories which converge on their own continue to yield the correct factorization, but these become less common.
    \item Beyond some saturation value $M_{sat}$ (roughly depicted as the transition from red to blue in the figure), both limit cycles and point attractors re-emerge, and they yield the incorrect factorization.
\end{itemize}
In theory, limit cycles of any length may appear, although in practice they tend to be skewed towards small cycle lengths. Networks with two factors are the most likely to find limit cycles, and this likelihood appears to decrease with increasing numbers of factors. Our intuition about what happens in the middle section of Figure \ref{fig:non_conv_regime} is that the basins of
attraction become very narrow and hard to find for the Resonator Network dynamics. The algorithm will wander, since it has so few spurious
fixed points (see Section \ref{sec:explaining_operational_capacity}), but not be able to find any basin of attraction.

\subsection{Factoring a `noisy' composite vector}
\label{sec:noisy_factorization}
\begin{figure}[t!]
    \centering
    \includegraphics[width=0.6\textwidth]{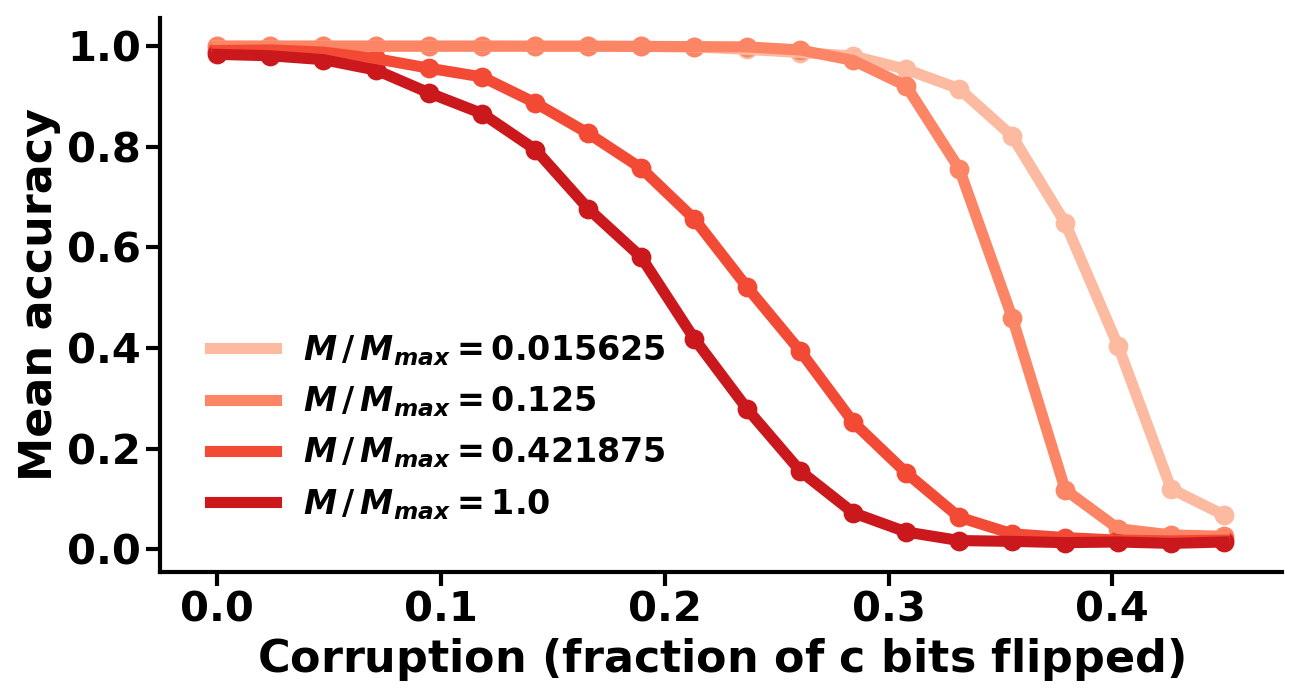}
    \caption{Factoring a corrupted $\mathbf{c}$. For $M$ well below capacity (lighter curves above) one can sustain heavy corruption to $\mathbf{c}$ and still find the correct factorization.}
    \label{fig:corrupted_c}
\end{figure}

Our assumption has been that one combination of codevectors from our codebooks $\mathbb{X}_f$ generates $\mathbf{c}$ exactly. What if this is not the case? Perhaps the vector we are given for factorization has had some proportion $\zeta$ of its components flipped, that is,
we are given $\tilde{\mathbf{c}}$ where $\tilde{\mathbf{c}}$ differs from $\mathbf{c}$ in exactly $\left \lfloor{\zeta N }\right \rfloor$ places. The vector $\mathbf{c}$ has a factorization based on our codebooks but $\tilde{\mathbf{c}}$ does not. We should hope that
a Resonator Network will return the factors of $\mathbf{c}$ so long as the corruption is not too severe. This is an especially important capability in the context of Vector Symbolic Architectures, where $\tilde{\mathbf{c}}$ will often be the result of some
algebraic manipulations that generate noise and corrupt the original $\mathbf{c}$ to some degree. We show in Figure \ref{fig:corrupted_c} that a Resonator Network can still produce the correct factorization even after a significant number of bits have been flipped.
This robustness is more pronounced when the number of factorizations is well below operational capacity, at which point the model can often still recover the correct factorization even when $30\%$ of the bits have been flipped.

\newpage
\subsection{A theory for differences in operational capacity}
\label{sec:explaining_operational_capacity}
The failure mode of each benchmark algorithm is getting stuck at a \emph{spurious fixed point} of the dynamics.
This section develops a simple comparison between the spurious fixed points of Resonator Networks and the benchmarks as an explanation for why Resonator Networks
enjoy relatively higher operational capacity. From among the benchmarks we focus on Projected Gradient Descent
(applied to the negative inner product with the simplex constraint) to illustrate this point. We will show that the correct factorization is always stable under Projected Gradient Descent
(as it is with the OLS variant of Resonator Networks), but that incorrect factorizations are much more likely to be fixed points under Projected Gradient Descent.
The definition of Projected Gradient Descent can be found in Table \ref{tab:linear_unit_dynamics}, with some comments in Appendix \ref{appendix:PGD}.

\subsubsection{Stability of the correct factorization}
The vector of coefficients $\mathbf{a}_f$ is a fixed point of Projected Gradient Descent dynamics when the gradient at this point is exactly
$\mathbf{0}$ or when it is in the nullspace of the projection operator. We write
\begin{equation}
    \label{eq:defn_projection_operator}
    \mathcal{N}\big(\mathcal{P}_{C_f}[\mathbf{x}] \big) := \{ \mathbf{z} \mid \mathcal{P}_{C_f} \big[ \mathbf{x} + \mathbf{z} \big] = \mathcal{P}_{C_f} \big[ \mathbf{x} \big] \}
\end{equation}
to denote this set of points. The nullspace of the projection operator is relatively small on the faces and edges of the simplex, but it becomes somewhat large at the vertices. We denote a vertex by
$\mathbf{e}_i$ (where ${(\mathbf{e}_i)}_j = 1$ if $j = i$ and $0$ otherwise).
The nullspace of the projection operator at a vertex of the simplex
is an intersection of halfspaces (each halfspace given
by an edge of the simplex). We can compactly represent it with the following expression:
\begin{equation}
    \label{eq:null_space_ver1}
    \mathcal{N}\big(\mathcal{P}_{\Delta_{D_f}}[\mathbf{e}_i] \big) = \big\{ \mathbf{z} \, \mid \, \bigcap_{j\neq i} {(\mathbf{e}_i - \mathbf{e}_j)}^\top \mathbf{z} \geq 1 \big\}
\end{equation}
An equivalent way to express the nullspace is
\begin{equation}
    \label{eq:null_space_ver2}
    \mathcal{N}\big(\mathcal{P}_{\Delta_{D_f}}[\mathbf{e}_i] \big) = \big\{ \mathbf{z} \, \mid \, z_j \leq z_i - 1 \,\,\, \forall j \neq i \big\}
\end{equation}
In other words, for a vector to be in the nullspace at $\mathbf{e}_i$, the $i$th element of the vector must be the largest by a margin of $1$ or more. This condition is met for the vector
$- \factorgradient$ at the correct factorization, since $- \factorgradient = \mathbf{X}_f^\top \big(\hat{\mathbf{o}}^{(f)}[0] \odot \mathbf{c}\big) = \mathbf{X}_f^\top \mathbf{x}_\star^{(f)}$.
This vector has a value $N$ for the component corresponding to $\mathbf{x}_\star^{(f)}$ and values that are $\leq N - 1$ for
all the other components. Thus, the correct factorization (the solution to (\ref{eq:general_factorization}) and global minimizer
of (\ref{eq:superposition_factorization})) is always a fixed point under the dynamics of Projected Gradient Descent (PGD).

This matches the stability of OLS Resonator Networks which are, by construction, always stable at the correct factorization. We showed in Section
\ref{sec:results_percolated_noise} that OP weights induce instability and that percolated noise makes the model marginally less stable than Hopfield Networks, but there is still a
large range of factorization problem sizes where the network is stable with overwhelming probability. What distinguishes the benchmarks from Resonator Networks is what we cover next,
the stability of \emph{incorrect} factorizations.

\subsubsection{Stability of incorrect factorizations}
Suppose initialization is done with a random combination of codevectors that do not produce $\mathbf{c}$. The vector $\hat{\mathbf{o}}^{(f)}[0] \odot \mathbf{c}$ will be a
\emph{completely random bipolar vector}. So long as $D_f$ is significantly smaller than $N$, which it always is in our applications,
$\hat{\mathbf{o}}^{(f)}[0] \odot \mathbf{c}$ will be nearly orthogonal to every vector in $\mathbb{X}_f$ and its projection onto $\mathcal{R}(\mathbf{X}_f)$ will be small,
with each component equally likely to be positive or negative.
Therefore, under the dynamics of a Resonator Network with OLS weights, each component will flip its sign compared to the initial state with probability $1/2$, and
the state for this factor will remain unchanged with the minuscule probability $1/2^N$. The total probability of this incorrect factorization being stable, accounting for each factor,
is therefore ${(1/2^N)}^F$. Suboptimal factorizations are \emph{very} unlikely to be a fixed points.
The same is true for a Resonator Network with OP weights because each element of the vector $\OPsynapses \big(\hat{\mathbf{o}}^{(f)}[0] \odot \mathbf{c}\big)$ is approximately Gaussian with mean zero
(see Section \ref{sec:results_percolated_noise} and Appendix \ref{appendix:stable_memory_cap}).

Contrast this against Projected Gradient Descent. We recall from (\ref{eq:null_space_ver2}) that the requirement for $\mathbf{e}_i$ to be a fixed point is that the $i$th component of the gradient at this point be largest by a margin of $1$ or more. \emph{This is a much
looser stability condition than we had for Resonator Networks} -- such a scenario will actually occur with probability $1/D_f$ for each factor, and the total probability is $1/M$. While still a relatively
small probability, in typical VSA settings $1/M$ is much larger than ${(1/2^N)}^F$, meaning that compared to Resonator Networks, Projected Gradient Descent is much more stable at incorrect factorizations.
Empirically, the failure mode of Projected Gradient Descent involves it settling on one of these spurious fixed points.

\subsubsection{Stability in general}
The cases of correct and incorrect factorizations drawn from the codebooks are two extremes along a continuum of possible states the algorithm can be in. For Projected Gradient Descent any state will
be stable with probability in the interval $[\tfrac{1}{M}, 1]$, while for Resonator Networks (with OLS weights) the interval is $[\tfrac{1}{2^{FN}}, 1]$.
In practical settings for VSAs, the interval $[\tfrac{1}{2^{FN}}, 1]$ is, in a relative sense, much larger than $[\tfrac{1}{M}, 1]$.
Vectors drawn uniformly from either $\{-1, -1\}^N$ or $[-1, -1]^N$ concentrate near the lower end of these intervals, suggesting that on average, \textbf{Projected Gradient Descent has many more spurious fixed points}.

This statement is not fully complete in the sense that dynamics steer the state along specific trajectories, visiting states in a potentially non-uniform way, but it does suggest that
Projected Gradient Descent is much more susceptible to spurious fixed points. The next section shows that these trajectories do in fact converge on spurious fixed points as the factorization problem
size grows.

\subsubsection{Basins of attraction for benchmark algorithms}
It may be that while there are sizable basins of attraction around the correct factorization, moving through the interior of the hypercube causes state trajectories to fall into the basin corresponding to a spurious fixed point.
In a normal setting for several of the optimization-based approaches,
we initialize $\mathbf{a}_f$ to be at the center of the simplex, indicating that each of the factorizations is equally likely.
Suppose we were to initialize $\mathbf{a}_f$ so that it is just slightly nudged toward one of the simplex vertices. We might nudge it toward the correct vertex (the one given by $\mathbf{a}_f^\star$) or we might
nudge it toward any of the other vertices, away from $\mathbf{a}_f^\star$. We can parameterize this with a single scalar $\theta$ and $\mathbf{e}_i$ chosen uniformly among the possible vertices:
\begin{equation}
    \mathbf{a}_f[0] = \theta \mathbf{e}_i + (1 - \theta)\frac{1}{D_f} \mathbf{1} \quad \mid \quad \theta \in [0, 1] \, , \,\, i \sim \mathcal{U}\{1, D_f\}
    \label{eq:simplex_interp_setup}
\end{equation}

\begin{figure}[t]
    \centering
    \includegraphics[width=0.5\textwidth]{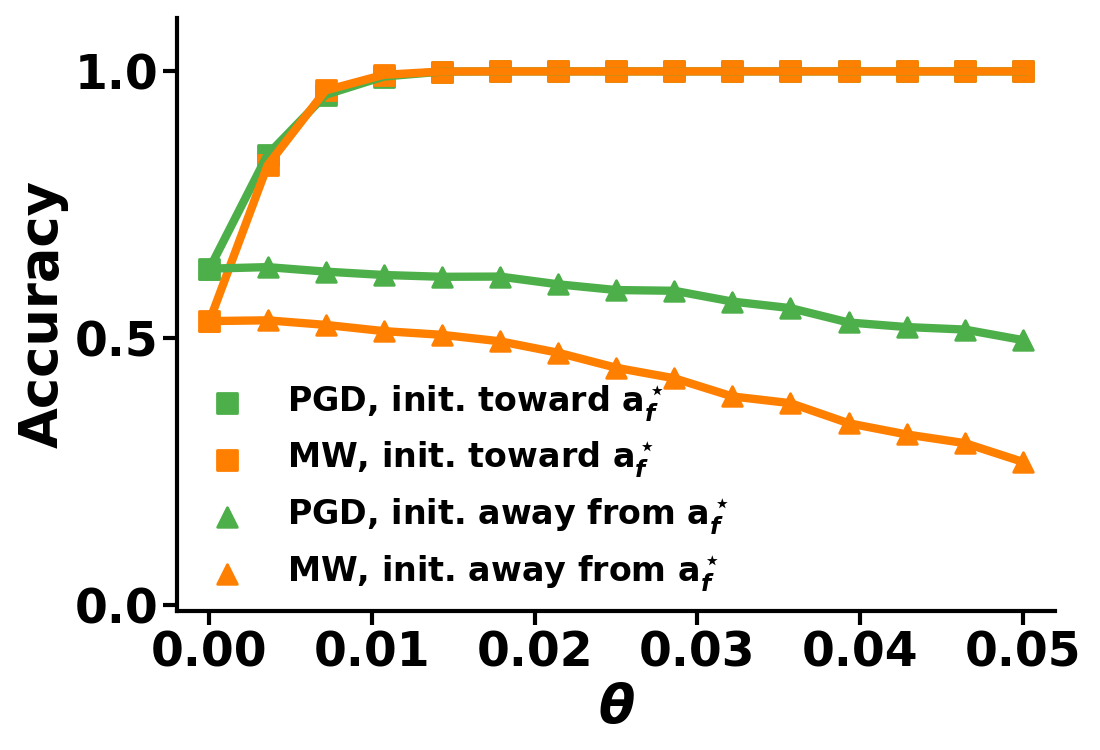}
    \caption{States in hypercube interior get pulled into spurious basins of attraction. Projected Gradient Descent is in green and Multiplicative Weights is in orange. Network is initialized at a distance $\theta$ from the center of the simplex (see equation (\ref{eq:simplex_interp_setup})), and allowed to converge. The y-axis is the accuracy of the factorization implied by the converged state. Triangles
    indicate initialization slightly away from $\mathbf{a}_f^\star$ toward any of the other simplex vertices, which is most directions in the space. These initial states get quickly pulled into a spurious
    basin of attraction.}
    \label{fig:simplex_basins}
\end{figure}

We ran a simulation with $N=1500$ and $D_1 = D_2 = D_3 = 50$, at which Projected Gradient Descent and Multiplicative Weights have a total accuracy of $0.625$ and $0.525$, respectively.
We created $5{,}000$ random factorization problems, initializing the state according to (\ref{eq:simplex_interp_setup}) and allowing the dynamics to run until convergence. We did this first with a
nudge toward the correct factorization $\mathbf{a}_f^\star$ (squares in Figure \ref{fig:simplex_basins}) and then with a nudge away from $\mathbf{a}_f^\star$, toward a randomly-chosen
spurious factorization (triangles in Figure \ref{fig:simplex_basins}).

What Figure \ref{fig:simplex_basins} shows is that by moving just a small distance toward the correct vertex, we very quickly fall into its basin of attraction. However, moving toward any of the other vertices is actually somewhat likely to take us into a spurious basin of attraction (where
the converged state is decoded into an incorrect factorization). The space is \emph{full} of these bad directions. It would be very lucky indeed to start from the center of the simplex and move immediately toward the solution -- it is far more likely that initial updates take us somewhere else
in the space, toward one of the other vertices, and this plot shows that these trajectories often get pulled towards a spurious fixed point. What we are demonstrating here is that empirically, the interior of the hypercube is somewhat treacherous from an optimization perspective, and this lies at the heart of why the benchmark algorithms fail.

From among the benchmarks, we restricted our analysis of spurious fixed points to Projected Gradient Descent and, in Figure \ref{fig:simplex_basins}, Multiplicative Weights. This choice was made for
clarity, and similar arguments apply for all of the benchmarks. While the details may differ slightly (for instance, spurious fixed points of ALS appear near the simplex center, not at a vertex), the
failure mode of the benchmarks is strikingly consistent. They all become overwhelmed by spurious fixed points, long before this affect is felt by Resonator Networks.
We have shown that \textbf{in expectation, Projected Gradient Descent has many more spurious fixed points than Resonator Networks}. We have also show that
\textbf{trajectories moving through the interior of the hypercube are easily pulled into these spurious basins of attraction.}

\section{Discussion}
\label{sec:discussion}

We studied a vector factorization problem which arises in the use of Vector Symbolic Architectures (as introduced in part
one of this series \citep{frady2020resonator}, showing that Resonator Networks solve this problem remarkably well. Their performance comes from a particular form of nonlinear
dynamics, coupled with the idea of searching in superposition. Solutions to the factorization problem lie in a small sliver of $\mathbb{R}^N$ -- i.e., the corners of the bipolar hypercube $\{-1, 1\}^N$ --
and the highly nonlinear activation function of Resonator Networks serves to constrain the search to this subspace. We drew connections between Resonator Networks and a number of benchmark algorithms which cast factorization as a problem of \emph{optimization}. This intuitively-satisfying formulation appears to come at a steep cost. None of the benchmarks were competitive with Resonator Networks in terms of key metrics that characterize factorization performance. One explanation for this is that the benchmarks have comparatively many more spurious fixed points of their dynamics, and that the loss function landscape in the interior of the hypercube induces trajectories that approach these spurious fixed points.

Unlike the benchmarks, Resonator Networks do not have a global convergence guarantee, and in some respects we see this as beneficial characteristic of the model. Requiring global convergence appears to
unnecessarily constrain the search for factorizations, leading to lower capacity. Besides, operational capacity (defined in this paper) specifies a regime where the lack of a convergence guarantee can be
practically ignored. Resonator Networks almost always converge in this setting, and the fixed points yield the correct solution. The benchmarks are, by steadfastly descending a loss function, in some sense more greedy than Resonator Networks. It appears that Resonator Networks strike a more natural balance between 1) making updates based on the best-available local information and 2) still exploring the
solution space while not getting stuck. Our approach follows a kind of ``Goldilocks principle'' on this trade off -- not too much, not too little, but just right.

We are not the first to consider eschewing convergence guarantees to better
solve hard search problems. For instance, randomized search algorithms utilize some explicit form of randomness to find better solutions, typically only converging if this randomness
is reduced over time \citep{spall2005introduction}. In contrast, our model is completely deterministic, and the searching behavior comes from nonlinear heteroassociative dynamics.
Another example is the proposal to add small amounts of random asymmetry to the (symmetric) weight matrix of Hopfield Networks \citep{hertz1986memory}. This modification removes the guaranteed absence of cyclic and chaotic trajectories that holds for the traditional Hopfield model. But, at the same time, and without significantly harming the attraction of memory states, adding asymmetry to the weights can improve associative memory recall by shrinking the basins of attraction associated with spurious fixed points \citep{singh1995fixed,chengxiang2000retrieval}.

We emphasize that, while Resonator Networks appear to be better than alternatives for the particular vector factorization problem (\ref{eq:general_factorization}), this is not a claim they are appropriate for
other hard search problems. Rather, Resonator Networks are specifically designed for the vector factorization problem at hand.
There exist several prior works involving some aspect of factorization that we would like to mention here, but we emphasize that each one of them deals with a problem or approach that is distinct from what we have
introduced in this paper.

Tensor decomposition is a family of problems that bear some resemblance to the factorization problem we have introduced (\ref{eq:general_factorization}). Key differences include the object to be factored,
which is a higher-order tensor, not a vector, and constraints on the allowable factors.
We explain in Appendix \ref{appendix:Tensor_decomp} how our factorization problem is
different from traditional tensor decompositions. Our benchmarks actually included the standard tensor decomposition algorithm, Alternating Least Squares,
re-expressed for (\ref{eq:general_factorization}), and we found that it is not well-matched for this factorization problem.
Bidirectional Associative
Memory, proposed by \citet{kosko1988bidirectional}, is an extension of Hopfield Networks that stores \emph{pairs} of factors in a matrix using the outer product learning rule. The composite object is a matrix, rather than a vector, and is much closer to
a particular type of tensor decomposition called the `CP decomposition', which we elaborate on in Appendix \ref{appendix:Tensor_decomp}. Besides the fact that this model applies only to \emph{two} factor problems, its dynamics are different from ours and its capacity is relatively low
\citep{kobayashi2002multidirectional}. Subsequent efforts to extend this model to factorizations with $3$ or more factors \citep{huang1999new, kobayashi2002multidirectional} have had very limited success and still rely on matrices that connect pairs of factors
rather than a single multilinear product, which we have in our model.
Bilinear Models of Style and Content \citep{tenenbaum2000separating} was an inspiration for us in deciding to work on factorization problems. This paper applies a different type of
tensor decomposition, a `Tucker decomposition' (again see Appendix \ref{appendix:Tensor_decomp}), to a variety of different real-valued datasets using what appears to be in one case a closed-form solution based on the
Singular Value Decomposition, and in the other case a variant of Alternating Least Squares. In that sense, their method is different from ours, the factorization problem is itself different,
and they consider only pairs of factors. \citet{memisevic2010learning} revisits the Tucker decomposition problem, but factors the `core' tensor representing interactions between factors
in order to make estimation more tractable. They propose a Boltzmann Machine that computes the factorization and show some results on modeling image transformations.
Finally, there is a large body of work on matrix factorization of the form
$\mathbf{V} \approx \mathbf{W}\mathbf{H}$, the most well-known of which is probably Non-negative Matrix Factorization \citep{lee2001algorithms}. The matrix $\mathbf{V}$ can be thought of a sum of outer products, so this is really a type of
CP decomposition with an additional constraint on the sign of the factors. Different still is the fact that $\mathbf{W}$ is often interpreted as a basis for the columns of $\mathbf{V}$, with $\mathbf{H}$ containing the
coefficients of each column with respect to this basis. In this sense, vectors are being \emph{added} to explain $\mathbf{V}$, rather than combined multiplicatively -- Non-negative Matrix Factorization is much closer to Sparse Coding \citep{hoyer2004non}.

The first paper in this series \citep{frady2020resonator} illustrated how distributed representations of data structures can be built with the algebra of Vector Symbolic Architectures,
as well as how Resonator Networks can decompose these datastructures.
VSAs are a powerful way to think about structured connectionist representations, and Resonator Networks make the framework much more scalable.
Extending the examples found in that paper to more realistic data (for example, complex 3-dimensional visual scenes) could be a useful application of Resonator Networks. This
will likely require learning a transform from pixels into the space of high-dimensional symbolic vectors, and this learning should ideally occur in the context of the factorization
dynamics -- we feel that this is an exciting avenue for future study. Here we have not shown Resonator Circuits for anything other than bipolar vectors. However, a version of the model wherein
vector elements are unit-magnitude complex phasors is a natural next extension, and relevant to Holographic Reduced Representations, a VSA developed by Tony Plate \citep{plate2003holographic}.
A recent theory of sparse phasor associative memories \citep{frady2019robust} may allow one to perform this factorization with a network of spiking neurons.

Resonator networks are an abstract neural model of factorization, introduced for the first time in this two-part series. We believe that as the theory and applications of Resonator Networks
are further developed, they may help us understand factorization in the brain, which still remains an important mystery.

\section{Acknowledgements}
We would like to thank members of the Redwood Center for Theoretical Neuroscience for helpful discussions, in particular Pentti Kanerva, whose work on Vector Symbolic Architectures originally motivated this project. This work was generously supported by the National Science Foundation, under graduate research fellowship DGE1752814 and research grant IIS1718991, by the National Institute of Health, grant 1R01EB026955-01, the Seminconductor Research Corporation under E2CDA-NRI, and DARPA's Virtual Intelligence Processing (VIP) program.

\pagebreak
\bibliographystyle{apalike}
\bibliography{main}

\pagebreak
\appendix
\addcontentsline{toc}{section}{Appendices}

\section*{\huge{Appendices}}
\section{Implementation details}
\label{appendix:pseudocode}
This section includes a few comments relevant to the implementation of Resonator Networks. Algorithm \ref{alg:pseudocode} gives psuedocode for Ordinary Least Squares weights -- the only change
for outer product weights is to use $\mathbf{X}^\top$ instead of $\mathbf{X}^\dagger$. So long as $D_f < N/2$, computing $\mathbf{X}_f\mathbf{X}_f^\dagger \big(\hat{\mathbf{o}} \odot \mathbf{c}\big)$ has lower computational
complexity than actually forming a single ``synaptic matrix'' $\mathbf{T}_f := \mathbf{X}_f\mathbf{X}_f^\dagger$ and then computing $\mathbf{T}_f \big(\hat{\mathbf{o}} \odot \mathbf{c}\big)$ in each
iteration -- it is faster to keep the matrices $\mathbf{X}_f$ and $\mathbf{X}_f^\dagger$ separate. This, of course, assumes that implementation is on a conventional computer.
If one can use specialized analog computation, such as large mesh circuits that directly implement matrix-vector multiplication in linear time \citep{cannon1969cellular}, then it would be preferable to store the
synaptic matrix directly.

Lines \ref{alg:_line_cleanup_for_start} - \ref{alg:_line_cleanup_for_end} in Algorithm
\ref{alg:pseudocode} ``clean up'' $\hat{\mathbf{x}}^{(f)}$ using the nearest neighbor in the codebook, and also resolve a sign ambiguity inherent to the factorization problem. The sign ambiguity is simply this:
while $\mathbf{c} = \mathbf{x}_{\star}^{(1)} \odot \mathbf{x}_{\star}^{(2)} \odot \ldots \odot \mathbf{x}_{\star}^{(F)}$ is the factorization we are searching for, we also have
$\mathbf{c} = -\mathbf{x}_{\star}^{(1)} \odot -\mathbf{x}_{\star}^{(2)} \odot \ldots \odot \mathbf{x}_{\star}^{(F)}$, and, more generally, any \emph{even} number of factors can have their signs flipped but still
generate the correct $\mathbf{c}$. Resonator Networks will sometimes find these solutions. We clean up using the codevector with the largest \emph{unsigned} similarity to the converged
$\hat{\mathbf{x}}^{(f)}$, which remedies this issue.
One will notice that we have written Algorithm \ref{alg:pseudocode} to update factors in order from $1$ to $F$. This is completely arbitrary, and any ordering is fine. We have experimented with
choosing a random update order during each iteration, but this did not seem to significantly affect performance.

Computing $\hat{\mathbf{o}}$ with the most-recently updated values for factors
$1$ to $f-1$ (see equation (\ref{eq:defn_of_o}))
is a convention we call `asynchronous' updates, in rough analogy to the same term used in the context of Hopfield Networks. An alternative convention is to, when computing $\hat{\mathbf{o}}$,
not used freshly updated values for factors $1$ to $f-1$, but rather their values before the update. This treats each factor as if it is being updated simultaneously,
a convention we call `synchronous' updates. This
distinction is an idiosyncrasy of modeling Resonator Networks in discrete-time, and the difference between the two disappears in continuous-time, where things happens instantaneously.
Throughout this paper, our analysis and simulations
have been with `asynchronous' updates, which we find to converge significantly faster.

Not shown in Algorithm \ref{alg:pseudocode} is the fact that, in practice, we record a
buffer of past states, allowing us to detect when the dynamics fall into a limit cycle, and to terminate early.

\newpage
\begin{algorithm}[h]
\caption{Resonator Network with Ordinary Least Squares weights}
\label{alg:pseudocode}
\begin{algorithmic}[1]
    \Require{$\mathbf{c}$}\Comment{Composite vector to be factored}
    \Require{$\mathbf{X}_1, \mathbf{X}_2, \ldots, \mathbf{X}_F$}\Comment{Codebook matrices \big($\mathbf{x}_j^{(f)} = \mathbf{X}_f[\,\,:\, , \, j \,]$\big)}
    \Require{k}\Comment{Maximum allowed iterations}
    \State $\hat{\mathbf{x}}^{(f)} \gets \text{sgn} \big (\sum_j \mathbf{x}^{(f)}_j \big) \quad \forall f=1, \ldots, F$
    \State $\mathbf{X}_f^\dagger \gets \text{pinv}(\mathbf{X}_f) \quad \forall f=1, \ldots, F$
    \State $i \gets 0$
    \While{not converged \textbf{and} $i < k$}
        \For{$f=1$ \textbf{to} $F$}
            \State $\hat{\mathbf{o}} \gets \hat{\mathbf{x}}^{(1)} \odot \ldots \odot \hat{\mathbf{x}}^{(f-1)}
      \odot \hat{\mathbf{x}}^{(f+1)} \odot \ldots \odot \hat{\mathbf{x}}^{(F)}$
            \State $\hat{\mathbf{x}}^{(f)} \gets \text{sgn} \Big(\mathbf{X}_f\mathbf{X}_f^\dagger \big(\hat{\mathbf{o}} \odot \mathbf{c}\big) \Big)$
        \EndFor
        \State $i \gets i + 1$
    \EndWhile
\For{$f=1$ \textbf{to} $F$}\Comment{Nearest Neighbor decoding}\label{alg:_line_cleanup_for_start}
    \State $u \gets \argmax_j \, | \text{sim}(\hat{\mathbf{x}}^{(f)}, \,\mathbf{x}^{(f)}_j) |$ \Comment{\emph{Un-signed} NN w.r.t $\cos$-similarity}
    \State $\hat{\mathbf{x}}^{(f)} \gets \mathbf{x}^{(f)}_u$ \label{alg:_line_cleanup_for_end}
\EndFor
\State \Return $\hat{\mathbf{x}}^{(f)} \,\, \forall f=1, \ldots, F$
\end{algorithmic}
\end{algorithm}

\newpage
\newpage
\section{Operational Capacity}
\label{appendix:op_cap}
The main text introduced our definition of operational capacity and highlighted our two main results -- that Resonator Networks have superior operational capacity compared to the benchmark algorithms, and that Resonator Network
capacity scales as a quadratic function of $N$. This appendix provides some additional support and commentary on these findings.

\begin{figure}[h!]
    \centering
    \includegraphics[width=0.8\textwidth]{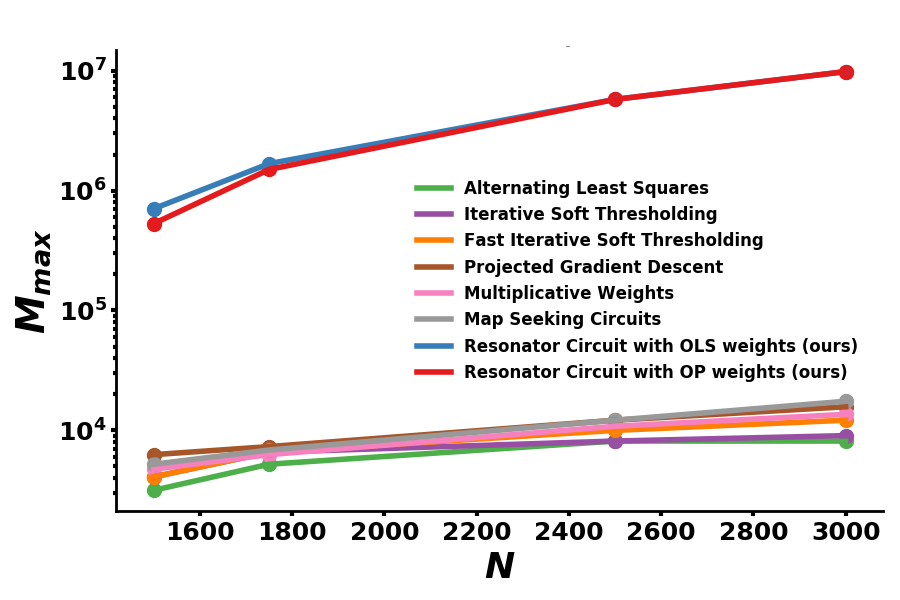}
    \caption{Comparing operational capacity against the benchmarks for $F=4$ ($4$ factors)}
    \label{fig:appendix_op_cap_comparison_4fac}
\end{figure}

Figure \ref{fig:appendix_op_cap_comparison_4fac} compares operational capacity among all of the considered algorithms when $F$, the number of factors, is $4$. We previously showed this type of plot for $F=3$, which was Figure
\ref{fig:op_cap_comparison} in the main text. Resonator Networks have an advantage of between two and three orders of magnitude compared to all of our benchmarks; the general size of this gap was consistent in
all of our simulations.

We concluded in Section \ref{sec:results_op_cap} that the operational capacity of Resonator Networks scales quadratically in $N$, which was shown in Figure \ref{fig:op_cap_mega}. In Table
\ref{table:op_cap_quadratic_fits} we provide parameters of the least-squares quadratic fits shown in that plot. One can see from Figure \ref{fig:op_cap_loglog} that capacity is different depending on the number of factors involved,
and in the limit of large $N$ this difference is determined by the parameter $c$. $c$ first rises from $2$ to $3$ factors, and then falls with increasing $F$.
This implies that factorization is easiest for Resonator Networks when the decomposition is into $3$ factors, an interesting phenomenon for which we do not have an
explanation at this time.

\begin{figure}[t]
    \centering
    \begin{subfigure}{0.32\textwidth}
        \centering
        \includegraphics[width=\textwidth]{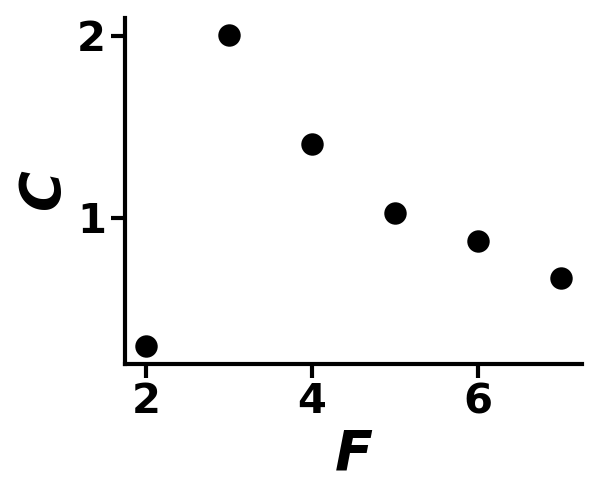}
    \end{subfigure}
    \hspace{8ex}
    \begin{subfigure}{0.32\textwidth}
        \centering
        \hspace{-7ex}
        \includegraphics[width=\textwidth]{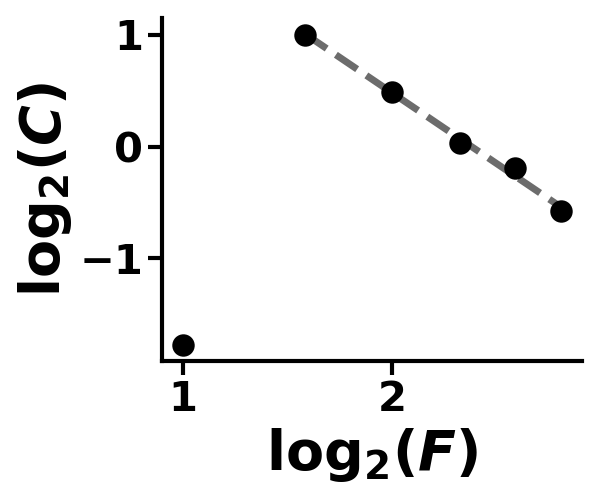}
    \end{subfigure}
    \caption{Parameter $c$ of the quadratic scaling depends on $F$. We find that it may follow an inverse power law for $F \geq 3$.}
    \label{fig:appendix_c_vs_F}
\end{figure}

Figure \ref{fig:appendix_c_vs_F} visualizes $c$ as a function of $F$. The data indicates that for $F \geq 3$, $c$ may follow an inverse power law: $c = \alpha_1 F^{-\alpha_2}$. The indicated linear
fit, following a logarithmic transformation to each axis, suggests the following values for parameters of this power law: $\alpha_1 \approx 2^{3.014} = 8.078$, $\alpha_2 \approx 1.268$.
It is with some reservation that we give these specific values for $\alpha_1$ and $\alpha_2$. Our estimates of operational capacity, while well-fit by quadratics, undoubtedly have
small amounts of noise. This noise can have a big enough impact on fitted values for $c$ that \emph{fitting the fit} may not be fully justified. However, we do note for the sake of completeness that
this scaling, if it holds for larger values of $F$, would allow us to write operational capacity in terms of both parameters $N$ and $F$ in the limit of large $N$:
\begin{equation}
    \label{eq:speculative_scaling}
    M_{max} \approx \frac{8.078 \, N^2}{F^{1.268}} \quad \forall F\geq 3
\end{equation}

\begin{table}[t!]
\centering
\setlength{\tabcolsep}{12pt} 
\renewcommand{\arraystretch}{1.3} 
\begin{tabular}{| l || c | c | c |}
    \hline
    \multirow{2}{*}{$\mathbf{F}$}  & \multicolumn{3}{c|}{\textbf{Parameters of quadratic fit}} \\
    \cline{2-4}
     & $\mathbf{a}$ & $\mathbf{b}$ & $\mathbf{c}$ \\
    \hhline{|=||=|=|=|}
    $\mathbf{2}$ & $1.677\times10^5$ & $-3.253\times10^2$ & $0.293$ \\
    \hline
    $\mathbf{3}$ & $1.230\times10^6$ & $-3.549\times10^3$ & $2.002$ \\
    \hline
    $\mathbf{4}$ & $-5.663\times10^6$ & $9.961\times10^2$ & $1.404$ \\
    \hline
    $\mathbf{5}$ & $1.140\times10^6$ & $-2.404\times10^3$ & $1.024$ \\
    \hline
    $\mathbf{6}$ & $5.789\times10^6$ & $-4.351\times10^3$ & $0.874$ \\
    \hline
    $\mathbf{7}$ & $-1.503\times10^7$ & $-1.551\times10^3$ & $0.669$ \\
    \hline
\end{tabular}
\caption{$M_{max} = a + bN + cN^2$}
\label{table:op_cap_quadratic_fits}
\end{table}

\clearpage
\section{Table of benchmark algorithms}
\setlength{\cellspacetoplimit}{0.7ex}
\setlength{\cellspacebottomlimit}{0.4ex}
\setlength{\tabcolsep}{0.8ex}
\begin{table}[h!]
\centering
\begin{tabular}{| Sc || Sc | Sc |}
\hline
\textbf{Algorithm} & \textbf{Dynamics for updating $\mathbf{a}_f[t]$} & \textbf{Eq.}\\
\hhline{|=||=|=|}
Alternating Least Squares &
\refstepcounter{equation}$\begin{gathered}\mathbf{a}_f[t+1] = \, {\big( \bm{\xi}^\top \bm{\xi} \big)}^{-1} \bm{\xi}^\top \mathbf{c} \\[8pt]
\bm{\xi} := \, \text{diag}\big(\hat{\mathbf{o}}^{(f)}[t]\big) \, \mathbf{X}_f \end{gathered}$ & (\ref{eq:ALS_dynamics})\tabularnewline
\hline
Iterative Soft Thresholding & \refstepcounter{equation}\label{eq:ISTA_dynamics}$\begin{gathered}
                              \mathbf{a}_f[t+1] = \mathcal{S}[\mathbf{a}_f[t] - \eta \, \factorgradient \,\, ; \lambda \eta] \\[8pt]
                              \big( \mathcal{S}[\, \mathbf{x} \, ; \gamma \, ] \big)_i :=  \text{sgn}(x_i) \, \text{max}(|x_i| - \gamma,\, 0\,)
                              \end{gathered}$& (\theequation)\tabularnewline
\hline
Fast Iterative Soft Thresholding & \refstepcounter{equation}\label{eq:FISTA_dynamics}$ \begin{gathered} \begin{aligned}
     \alpha_t &= \frac{1 + \sqrt{1 + 4\alpha_{t-1}^2}}{2} \\[2pt]
     \beta_t &= \frac{\alpha_{t-1} - 1}{\alpha_{t}} \\[4pt]
     \mathbf{p}_f[t+1] &= \mathbf{a}_f[t] + \beta_t (\mathbf{a}_f[t] - \mathbf{a}_f[t-1]) \\[4pt]
     \mathbf{a}_f[t+1] &= \mathcal{S}[\mathbf{p}_f[t+1] - \eta \, \nabla_{\mspace{-4mu}\mathbf{p}_f} \mathcal{L} \,\, ; \lambda \eta] \end{aligned} \\[10pt]
     \big( \mathcal{S}[\, \mathbf{x} \, ; \gamma \, ] \big)_i :=  \text{sgn}(x_i) \, \text{max}(|x_i| - \gamma,\, 0\,)
     \end{gathered}$ & (\theequation)\tabularnewline
\hline
Projected Gradient Descent & \refstepcounter{equation}\label{eq:PGD_dynamics}$ \begin{gathered}
                               \mathbf{a}_f[t+1] = \, \mathcal{P}_{C_f} \big[ \mathbf{a}_f[t] - \eta \factorgradient \big] \\[8pt]
                               \mathcal{P}_{C_f}[\mathbf{x}] := \argmin_{\mathbf{z} \in C_f} \tfrac{1}{2} {\big| \big| \mathbf{x} - \mathbf{z} \big| \big|}^2_2
                             \end{gathered}$ &  (\theequation)\tabularnewline
\hline
Multiplicative Weights & \refstepcounter{equation}\label{eq:MW_dynamics}$ \begin{gathered} \begin{aligned}
                            \mathbf{w}_f[t+1] =& \, \mathbf{w}_f[t] \odot \big( \mathbf{1} - \frac{\eta}{\rho} \factorgradient \big) \\
                            \mathbf{a}_f[t+1] =& \, \frac{\mathbf{w}_f[t+1]}{\sum_i w_{fi}[t+1]}
                          \end{aligned} \\[8pt]
                            \rho := \max_{i} \big| (\factorgradient)_i \big|
                            \end{gathered}$ &  (\theequation)\tabularnewline
\hline
Map Seeking Circuits & \refstepcounter{equation}\label{eq:MSC_dynamics}$ \begin{gathered}
                         \mathbf{a}_f[t+1] = \,\mathcal{T}\Big(\mathbf{a}_f[t] - \eta \big( \mathbf{1} + \frac{1}{\rho}\factorgradient \big) \,; \,\, \epsilon \,\, \Big) \\[8pt]
                         \mathcal{T} \big( \mathbf{x} \,; \, \epsilon \, \big)_i := \begin{cases}  x_i & \text{if} \,\, x_i \geq \epsilon \\ 0 & \text{otherwise} \end{cases} \\[8pt]
                         \rho := \, \big| \min_{i} (\factorgradient)_i \big|
                         \end{gathered}$ & (\theequation)\tabularnewline
\hline
\end{tabular}
\caption{Dynamics for $\mathbf{a}_f$, benchmark algorithms. (see Appendices \ref{appendix:Tensor_decomp} - \ref{appendix:MSC} for discussion of each algorithm, including hyperparameters $\eta$, $\lambda$, and $\epsilon$, as well as
         initial conditions).}
\label{tab:linear_unit_dynamics}
\end{table}

\newpage
\section{Tensor Decompositions and Alternating Least Squares}
\label{appendix:Tensor_decomp}
Tensors are multidimensional arrays that generalize vectors and matrices. An Fth-order tensor has elements that can be indexed by F separate indexes -- a vector is a tensor of order $1$ and a matrix is a tensor of order $2$. As devices for measuring
multivariate time series have become more prevalent, the fact that this data can be expressed as a tensor has made the study of tensor decomposition a very popular subfield of applied mathematics. Hitchcock \citep{hitchcock1927expression}
is often credited with originally formulating tensor decompositions, but modern tensor decomposition was popularized in the field of psychometrics by the work of Tucker \citep{tucker1966some}, Carroll and Chang \citep{carroll1970analysis}, and
Harshman \citep{harshman1970foundations}. This section will highlight the substantial difference between tensor decomposition and the factorization problem solved by Resonator Networks.

The type of tensor decomposition most closely related to our factorization problem (given in equation (\ref{eq:general_factorization}))
decomposes an Fth-order tensor $\mathcal{C}$ into a sum of tensors each generated by the outer product $\circ$:
\begin{equation}
\label{eq:cp_decomposition}
\mathcal{C} = \sum_{r=1}^R \mathbf{x}_r^{(1)} \circ  \mathbf{x}_r^{(2)} \circ \ldots \circ \mathbf{x}_r^{(F)}
\end{equation}
The outer product contains all pairs of components from its two arguments, so $\big( \mathbf{w} \circ \mathbf{x} \circ \mathbf{y} \circ \mathbf{z} \big)_{ijkl} = w_i x_j y_k z_l$.
The interpretation is that each term in the sum is a ``rank-one''
tensor of order F and that $\mathcal{C}$ can be generated from the sum of $R$ of these rank-one tensors. We say that $\mathcal{C}$ is ``rank-R''. This particular decomposition has at least three different names in
the literature - they are Canonical Polyadic Decomposition, coined by Hitchcock, CANonical DECOMPosition (CANDECOMP), coined by Carroll and Chang, and  PARAllel FACtor analysis (PARAFAC), coined by Harshman.
We will simply call this the CP decomposition, in accordance with the convention used by Kolda \citep{kolda2009tensor} and many
others.

CP decomposition makes no mention of a codebook of vectors, such as we have in (\ref{eq:general_factorization}).
In CP decomposition, the search is apparently over all of the vectors in a real-valued vector space.
One very useful fact about CP decomposition is that under relatively mild conditions, \emph{if the decomposition exists, it is unique} up to a scaling and permutation indeterminacy.
Without going into the details, a result in \citet{kruskal1977three} and extended by \citet{sidiropoulos2000uniqueness} gives a sufficient condition for uniqueness of the CP decomposition based on
what is known as the Kruskal rank $k_{\mathbf{X}_f}$ of the matrix $\mathbf{X}_f := [\mathbf{x}_1^{(f)}, \mathbf{x}_2^{(f)}, \ldots \mathbf{x}_R^{(f)}]$:
\begin{equation}
\label{eq:cp_uniqueness_sufficient}
\sum_{f=1}^F k_{\mathbf{X}_f} \geq 2R + (F - 1)
\end{equation}

This fact of decomposition uniqueness illustrates one way that basic results from matrices fail to generalize to higher-order tensors (by higher-order we simply mean where the order is $\geq 3)$.
Low-rank CP decomposition for matrices (tensors of order $2$) may be computed with the truncated Singular Value Decomposition (SVD). However, if $\mathcal{C}$ is a matrix and its truncated SVD is
$\mathbf{U}\mathbf{\Sigma}\mathbf{V}^\top := \mathbf{X}_1 \mathbf{X}_2^\top$, then any non-singular matrix $\mathbf{M}$ generates an equally-good CP decomposition
$(\mathbf{U}\mathbf{\Sigma}\mathbf{M})(\mathbf{V}\mathbf{M}^{-1})^\top$. The decomposition is \emph{highly} non-unique. All matrices have an SVD, whereas generic higher-order tensors are not guaranteed to
have a CP decomposition. And yet, if a CP decomposition exists, under the mild condition of equation (\ref{eq:cp_uniqueness_sufficient}), it is unique. This is a somewhat miraculous
fact, suggesting that \emph{in this sense, CP decompostion of higher-order tensors is easier than matrices.
The higher order of the composite object imposes many more constraints that make the decomposition unique}.

Another interesting way that higher-order tensors differ from matrices is that computing
matrix rank is easy, whereas in general computing tensor rank is NP-hard, along with many other important tensor problems \citep{hillar2013most}. Our intuition about matrices largely
fails us when dealing with higher-order tensors. In some ways the problems are easier and in some ways they are harder. Please see \citet{sidiropoulos2017tensor} for a more comprehensive comparison.

The vector factorization problem defined by (\ref{eq:general_factorization}) differs from CP decomposition in three key ways:
\begin{enumerate}
    \item The composite object to be factored is a vector, not a higher-order tensor. This is an even more extreme difference than between matrices and higher-order tensors.
    In CP decomposition, the arrangement and numerosity of tensor elements constitute many constraints on what the
    factorization can be, so much so that it resolves the uniqueness issue we outlined above. In this sense, \emph{tensors contain much more information about the valid factorization, making the problem
    significantly easier}. The size and form of these tensors may make finding CP decompositions a computational challenge, but CP decomposition is analytically easier than our vector factorization problem.
    \item Search is conducted over a discrete set of possible factors. This differs from the standard formulation of CP decomposition, which makes no restriction to a discrete set of factors.
    It is however worth noting that a specialization of CP decomposition called CANonical DEcomposition with LINear Constraints (CANDELINC) \citep{carroll1980candelinc} does in fact impose the additional
    requirement that factors are formed from a linear combination of some basis factors. In our setup the solutions are `one-hot' linear combinations.
    \item The factors are constrained to $\{-1, 1\}^N$, a small sliver of $R^N$. This difference should not be underestimated. We have shown in Section \ref{sec:explaining_operational_capacity} that
    the interior of this hypercube is treacherous from an optimization perspective and Resonator Networks avoid it by using a highly nonlinear activation function. This would not make sense in the
    context of standard CP decomposition.
\end{enumerate}

Perhaps the most convincing demonstration that (\ref{eq:general_factorization}) is \emph{not} CP decomposition comes from the fact that we applied Alternating Least Squares to it and found that its
performance was relatively poor. Alternating Least Squares is in fact the `workhorse' algorithm of CP decomposition \citep{kolda2009tensor}, but it cannot compete with Resonator Networks on our
different factorization problem (\ref{eq:general_factorization}). The excellent review of \citet{kolda2009tensor} covers CP decomposition and Alternating Least Squares in significant depth,
including the fact that ALS always converges to a local minimum of the squared error reconstruction loss. See, in particular, section 3.4 of their paper for more detail.

One special case of CP decomposition involves rank-1 components that are \emph{symmetric} and \emph{orthogonal}. For this problem, a special case of ALS called the tensor power method can be used
to iteratively find the best low-rank CP decomposition through what is known as `deflation', which is identical to the explaining away we introduced in part one of this series \citep{frady2020resonator}.
The tensor power method directly generalizes the matrix power method, and in this special case of symmetric, orthogonal tensors is effective at finding the CP decomposition. A good initial reference for
the tensor power method is \citet{de2000best}. A discussion of applying tensor decompositions to statistical learning problems is covered by \citet{anandkumar2014tensor}, which develops a robust version of
the tensor power method and contains several important probabilistic results for applying tensor decompositions to noisy data. The tensor power method differs from Resonator Networks in the same key ways
as ALS -- composite objects are higher-order tensors, not vectors, search is not necessarily over a discrete set, the vectors are not constrained to $\{-1, 1\}^N$, and the dynamics make \emph{linear} least squares
updates in each factor.

\begin{figure}[t]
    \centering
    \includegraphics[width=0.85\textwidth]{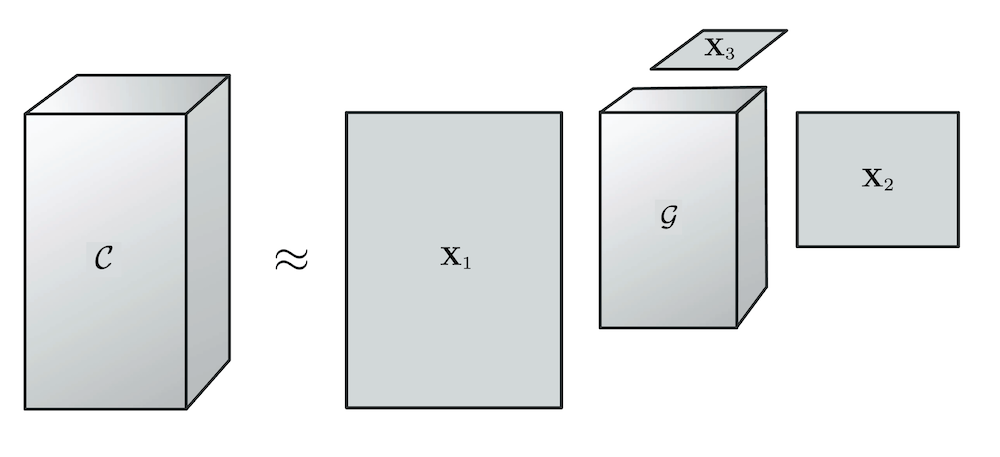}
    \caption{Tucker decomposition with $3$ factors}
    \label{fig:tucker_diagram}
\end{figure}

Another popular tensor decomposition is known as the Tucker Decomposition \citep{tucker1963implications, tucker1966some}. It adds to CP decomposition an order-F ``core tensor'' $\mathcal{G}$ that modifies the interaction between each of the factors:
\begin{equation}
\label{eq:tucker_decomposition}
\mathcal{C} = \sum_{p=1}^P\sum_{q=1}^Q\ldots\sum_{r=1}^R \, g_{pq\ldots r} \,\, \mathbf{x}_p^{(1)} \circ  \mathbf{x}_q^{(2)} \circ \ldots \circ \mathbf{x}_r^{(F)}
\end{equation}
This adds many more parameters compared to CP decomposition, which is a special case of Tucker decomposition when $\mathcal{G}$ is the identity. For the purpose of illustration, we reprint in Figure \ref{fig:tucker_diagram} (with a slight relabeling) a figure from \citet{kolda2009tensor} that depicts an order-$3$ Tucker decomposition.
This decomposition goes by many other names, most popularly the Higher-order SVD, coined in \citet{de2000multilinear}. The Tucker decomposition can also be found via Alternating Least Squares (see \citet{kolda2009tensor}, Section 4.2, for a tutorial), although the problem
is somewhat harder than CP decomposition, both by being computationally more expensive and by being non-unique. Despite this fact, the applications of Tucker decomposition are wide-ranging -- it has been used in psychometrics, signal processing, and computer vision. One well-known application
of Tucker decomposition in computer vision was TensorFaces \citep{vasilescu2002multilinear}. This model was able to factorize identity, illumination, viewpoint, and facial expression in a dataset consisting of face images.

The summary of this section is that vector factorization problem (\ref{eq:general_factorization}) is not tensor decomposition. In some sense it is more challenging. Perhaps not surprisingly, the
standard algorithm for tensor decompositions, Alternating Least Squares, is not particularly competitive on this problem when compared to Resonator Networks. It is interesting to consider whether
tensor decomposition might be cast into a form amenable to solution by Resonator Networks. Given the importance of tensor decomposition as a tool of data analysis, we believe this warrants a closer look.

\newpage
\section{General notes on gradient-based algorithms}
\label{appendix:general_grad_stuff}
When $\mathcal{L}$ is the negative inner product, the gradient with respect to $\mathbf{a}_f$ is:
\begin{align*}
    \label{eq:smooth_grad_neginnerprod}
    \factorgradient =& - \mathbf{X}_f^\top \big( \mathbf{c} \odot \hat{\mathbf{x}}^{(1)} \odot \ldots \odot \hat{\mathbf{x}}^{(f-1)}
      \odot \hat{\mathbf{x}}^{(f+1)} \odot \ldots \odot \hat{\mathbf{x}}^{(F)} \big) \\
      =& - \mathbf{X}_f^\top \big( \mathbf{c} \odot \hat{\mathbf{o}}^{(f)} \big) \numberthis
\end{align*}
The term $\mathbf{c} \odot \hat{\mathbf{o}}^{(f)}$ can be interpreted as an estimate for what
$\hat{\mathbf{x}}^{(f)}$ should be based on the current estimates for the \emph{other} factors.
Multiplying by $\mathbf{X}^\top_f$ compares the similarity of this vector to each of the candidate codevectors we are entertaining,
with the smallest element of $\factorgradient$ (its value is likely to be negative with large absolute value) indicating the codevector which matches best. Following the negative gradient will cause this coefficient to increase more than the coefficients corresponding to the other codevectors. When $\mathcal{L}$ is the squared error, the gradient with respect to $\mathbf{a}_f$ is:
\begin{align*}
    \label{eq:smooth_grad_squarederror}
    \factorgradient =& \, \mathbf{X}_f^\top \Big( \big( \mathbf{c} - \hat{\mathbf{x}}^{(1)} \odot \ldots \odot \hat{\mathbf{x}}^{(F)} \big) \odot
                        \big( - \hat{\mathbf{x}}^{(1)} \odot \ldots \odot \hat{\mathbf{x}}^{(f-1)} \odot \hat{\mathbf{x}}^{(f+1)} \odot \ldots \odot \hat{\mathbf{x}}^{(F)} \big) \Big) \\
                    :=& \,\mathbf{X}_f^\top \Big( \hat{\mathbf{x}}^{(f)} \odot {\big( \hat{\mathbf{o}}^{(f)} \big)}^2  - \mathbf{c} \odot \hat{\mathbf{o}}^{(f)} \Big) \numberthis
\end{align*}
This looks somewhat similar to the gradient for the negative inner product -- they differ by an additive term given by $\mathbf{X}_f^\top \Big(\hat{\mathbf{x}}^{(f)} \odot {\big( \hat{\mathbf{o}}^{(f)} \big)}^2 \Big)$. At the vertices of the hypercube all the elements of $\hat{\mathbf{x}}^{(f)}$ are $1$ or $-1$ and the term ${\big( \hat{\mathbf{o}}^{(f)} \big)}^2$ disappears, making the difference between the two gradients just
$\mathbf{X}_f^\top \hat{\mathbf{x}}^{(f)}$. Among other things, this makes the gradient of the squared error equal to zero at the global minimizer $\mathbf{x}_{\star}^{(1)} \ldots \mathbf{x}_{\star}^{(F)}$, which is not
the case with the negative inner product. To be clear, (\ref{eq:smooth_grad_neginnerprod}) is the gradient when the loss function is the negative inner product, while (\ref{eq:smooth_grad_squarederror}) is the gradient
when the loss function is the squared error.

\subsection{Fixed-stepsize gradient descent on the squared error}
\label{appendix:fixed_stepsize_doesnt_work}
In fixed-step-size gradient descent for unconstrained convex optimization problems, one must often add a restriction on the stepsize, related to the \emph{smoothness} of the loss function, in order to ensure that the iterates converge to a fixed point. We say that
a function $\mathcal{L}$ is $L$-smooth when its gradient is Lipschitz continuous with constant $L$:
\begin{equation}
    ||\nabla \mathcal{L}(\mathbf{x}) -  \nabla \mathcal{L}(\mathbf{y}) ||_2 \leq L||\mathbf{x} - \mathbf{y} ||_2 \,\,\, \forall \, \mathbf{x}, \, \mathbf{y}
\end{equation}
For a function that is twice-differentiable, this is equivalent to the condition
\begin{equation}
    \mathbf{0} \preceq \nabla^2 \mathcal{L}(\mathbf{x}) \preceq L\mathbf{I} \quad \forall \, \mathbf{x}
\end{equation}
Where $\mathbf{0}$ is the matrix of all zeros and and $\mathbf{I}$ is the identity. Absent some procedure for adjusting the stepsize $\eta$ at each iteration to account for the degree of local smoothness, or some additional assumption we place on the loss to make
sure that it is sufficiently smooth, we should be wary that convergence may not be guaranteed. On our factorization problem we find this to be an issue. Unconstrained gradient descent on the squared error works for the simplest problems,
where $M$ is small and the factorization can be easily found by any of the algorithms in this paper. However, as $M$ increases, the exceedingly ``jagged'' landscape of the squared error loss makes the iterates very sensitive to the step size $\eta$,
and the components of $\mathbf{a}_f[t]$ can become very large. When this happens, the term $\hat{\mathbf{o}}^{(f)}[t]$ amplifies this
problem (it multiplies all but one of the $\mathbf{a}_f[t]$'s together) and causes numerical instability issues.  With the squared error loss, the smoothness is very poor: we found that fixed-stepsize gradient descent on the squared error was
so sensitive to $\eta$ that it made the method practically useless for solving the factorization problem. Iterative Soft Thresholding and Fast Iterative Soft Thresholding use a dynamic step size to avoid this issue
(see equation (\ref{eq:lipschitz_ista})). In contrast, the negative inner product loss, with respect to each factor, is in some sense \emph{perfectly smooth} (it is linear), so the step size does not factor into convergence proofs.

\section{Iterative Soft Thresholding (ISTA) and Fast Iterative Soft Thresholding (FISTA)}
\label{appendix:ISTA}
Iterative Soft Thresholding is a type of \emph{proximal gradient descent}. The proximal operator for any convex function $h(\cdot)$ is defined as
\[ \text{prox}_h(\mathbf{x}) := \argmin_{\mathbf{z}} \,\, \tfrac{1}{2} ||\mathbf{z} -  \mathbf{x}||_2^2 + h(\mathbf{z}) \]
When $h(\mathbf{z})$ is $\lambda ||\mathbf{z}||_1$, the proximal operator is the so-called ``soft-thresholding'' function, which we denote by $\mathcal{S}$ :
\[\big( \mathcal{S}[\, \mathbf{x} \, ; \gamma \, ] \big)_i :=  \text{sgn}(x_i) \, \text{max}(|x_i| - \gamma,\, 0\,) \]
Consider taking the squared error loss and adding to it $\lambda ||\mathbf{a}_f||_1$:
\[ \mathcal{L}(\mathbf{c}, \hat{\mathbf{c}}) + \lambda ||\mathbf{a}_f||_1 = \tfrac{1}{2} || \mathbf{c} - \hat{\mathbf{c}} ||^2_2 + \lambda ||\mathbf{a}_f||_1 \]
Applying soft thesholding clearly minimizes this augmented loss function. The strategy is to take gradient steps with respect to the squared error loss but then to pass those updates through the soft thresholding function $\mathcal{S}$.
This flavor of proximal gradient descent, where $\hat{\mathbf{c}}$ is a linear function of $\mathbf{a}_f$ and $h(\cdot)$ is the $\ell_1$ norm, is called the Iterative Soft Thresholding Algorithm \citep{daubechies2004iterative}, and is a
somewhat old and popular approach for finding sparse solutions to large-scale linear inverse problems.

The dynamics of ISTA are given in equation (\ref{eq:ISTA_dynamics}) and there are a few parameters worth discussing. First, the dynamic stepsize $\eta$ can be set via backtracking line search or, as we did, by computing the Lipschitz
constant of the loss function gradient:
\begin{equation}
    \label{eq:lipschitz_ista}
    \eta = \frac{1}{L} \quad \big\lvert \quad ||\nabla_{\mathbf{a}} \mathcal{L}(\mathbf{x}) -  \nabla_{\mathbf{a}} \mathcal{L}(\mathbf{y}) ||_2 \leq L||\mathbf{x} - \mathbf{y} ||_2 \,\,\, \forall \, \mathbf{x}, \, \mathbf{y}
\end{equation}
The scalar $\lambda$ is a hyperparameter that effectively sets the sparsity of the solutions considered -- its value should be tuned in order to get good performance in practice. In the experiments we show in this paper, $\lambda$ was $0.01$. The initial state $\mathbf{a}_f[0]$ is set to $\mathbf{1}$.

Convergence analysis of ISTA is beyond the scope of this paper, but it has been shown in various places (\cite{bredies2008linear}, for instance) that ISTA will converge at a rate $\simeq \mathcal{O}(1/t)$.
Iterative Soft Thresholding works well in practice, although for $4$ or more factors we find that it is not quite as effective as the algorithms that do constrained descent on the negative inner product loss.
By virtue of not directly constraining the coefficients, ISTA allows them to grow outside of $[0, 1]^N$. This may make it easier to find the minimizers $\mathbf{a}_1^\star, \mathbf{a}_2^\star, \ldots, \mathbf{a}_F^\star$,
but it may also lead the method to encounter more suboptimal local minimizers, which we found to be the case in practice.

One common criticism of ISTA is that it can get trapped in shallow parts of the loss surface and thus suffers from slow convergence \citep{bredies2008linear}. A straightforward improvement, based on Nesterov's momentum for accelerating first-order
methods, was proposed by \citet{beck2009fast}, which they call Fast Iterative Soft Thresholding (FISTA). The dynamics of FISTA are written in equation (\ref{eq:FISTA_dynamics}), and converge at the significantly better rate of
$\simeq \mathcal{O}(1/{t^2})$, a result proven in \cite{beck2009fast}. Despite this difference in \emph{worst-case} convergence rate, we find that the average-case convergence rate on our particular factorization problem does
not significantly differ. Initial coefficients $\mathbf{a}_f[0]$ are set to $\mathbf{1}$ and auxiliary variable $\alpha_t$ is initialized to $1$. For all experiments $\lambda$ was set the same as for ISTA, to $0.01$.

\section{Projected Gradient Descent}
\label{appendix:PGD}
Starting from the general optimization form of the factorization problem (\ref{eq:superposition_factorization}), what kind of constraint might it be reasonable to enforce on $\mathbf{a}_f$? The most obvious is that $\mathbf{a}_f$ lie on the simplex $\Delta_{D_f} := \{\mathbf{x} \in \mathbb{R}^{D_f} \mid \sum_i x_i = 1, x_i \geq 0 \,\, \forall i\}$.
Enforcing this constraint means that $\hat{\mathbf{x}}^{(f)}$ stays within the $-1,1$ hypercube
at all times and, as we noted, the optimal values $\mathbf{a}_1^\star, \mathbf{a}_2^\star, \ldots, \mathbf{a}_F^\star$ happen to lie at vertices of the simplex, the standard basis vectors $\mathbf{e}_i$.
Another constraint set worth considering is the $\ell_1$ ball $\mathcal{B}_{||\cdot||_1}[1] := \{\mathbf{x} \in \mathbb{R}^{D_f} \mid ||\mathbf{x}||_1 \leq 1 \}$. This set contains the simplex, but it encompasses much more of $\mathbb{R}^{D_f}$.
One reason to consider the $\ell_1$ ball is that it dramatically increases the number of feasible global optimizers of (\ref{eq:superposition_factorization}), from which we can easily recover the specific solution to (\ref{eq:general_factorization}).
This is due to the fact that:
\[ c = \mathbf{X}_1 \mathbf{a}^{\star}_1 \odot \mathbf{X}_2 \mathbf{a}^{\star}_2 \odot \ldots \odot \mathbf{X}_F \mathbf{a}^{\star}_F  \iff
   c = \mathbf{X}_1 (-\mathbf{a}^{\star}_1) \odot \mathbf{X}_2 (-\mathbf{a}^{\star}_2) \odot \ldots \odot \mathbf{X}_F \mathbf{a}^{\star}_F \]
and moreover any number of distinct pairs of factor coefficients can be made negative -- the sign change cancels out. The result is that while the simplex constraint only allows solution
$\mathbf{a}_1^\star, \mathbf{a}_2^\star, \ldots, \mathbf{a}_F^\star$, the $\ell_1$ ball constraint
also allows solutions $-\mathbf{a}_1^\star, -\mathbf{a}_2^\star, \mathbf{a}_3^\star, \ldots, \mathbf{a}_F^\star$, and $\mathbf{a}_1^\star, \mathbf{a}_2^\star, -\mathbf{a}_3^\star, \ldots, -\mathbf{a}_F^\star$, and
$-\mathbf{a}_1^\star, -\mathbf{a}_2^\star, -\mathbf{a}_3^\star, \ldots, -\mathbf{a}_F^\star$, etc. These spurious global minimizers can easily be detected by checking the sign of the largest-magnitude component of $\mathbf{a}_f$. If it is negative we can
then multiply by $-1$ to get $\mathbf{a}_f^\star$. Choosing the $\ell_1$ ball over the simplex is purely motivated from the perspective that increasing the size of the constraint set may make finding the
global optimizers easier. However, we found that in practice, it did not significantly matter whether $\Delta_{D_f}$ or $\mathcal{B}_{||\cdot||_1}[1]$ was used to constrain $\mathbf{a}_f$.

There exist algorithms for efficiently computing projections onto both the simplex and
the $\ell_1$ ball (see \cite{held1974validation}, \cite{duchi2008efficient}, and \cite{condat2016fast}). We use a variant summarized in \citet{duchi2008efficient} that has computational complexity
$\mathcal{O}(D_f \log D_f)$ -- recall that $\mathbf{a}_f$ has $D_f$ components, so this is the dimensionality of the simplex or the $\ell_1$ ball being projected onto.
When constraining to the simplex, we set the initial coefficients $\mathbf{a}_f[0]$ to $\tfrac{1}{D_f} \mathbf{1}$, the center of the simplex. When constraining to the unit $\ell_1$ ball we set $\mathbf{a}_f[0]$ to $\tfrac{1}{2D_f} \mathbf{1}$, so that
all coefficients are equal but the vector is on the interior of the ball. The only hyperparameter is $\eta$, which in all experiments was
set to $0.01$. We remind the reader that we defined the nullspace of the projection operation with equation (\ref{eq:defn_projection_operator}) in Section \ref{sec:explaining_operational_capacity}, and
the special case for the simplex constraint in (\ref{eq:null_space_ver1}) and (\ref{eq:null_space_ver2}).

Taking projected gradient steps on the negative inner product loss works
well and is guaranteed to converge, whether we use the simplex or the $\ell_1$ ball constraint. Convergence is guaranteed due to this intuitive fact:
any part of $- \eta \, \factorgradient$ not in $\mathcal{N}\big(\mathcal{P}_{C_f}[\mathbf{x}] \big)$, induces a change in $\mathbf{a}_f$, denoted by $\Delta \mathbf{a}_f[t]$ which must make an acute angle
with $- \factorgradient$. This is by \emph{the definition of orthogonal projection}, and it is a sufficient condition for showing that $\Delta \mathbf{a}_f[t]$ decreases the value of the loss function.
Projected Gradient Descent iterates always reduce the value of the negative inner product loss or leave it unchanged; the function is bounded below on the simplex and the $\ell_1$ ball, so this algorithm is guaranteed to converge.

Applying projected gradient descent on the squared error did not work, which is related to the smoothness issue we discussed in Appendix \ref{appendix:fixed_stepsize_doesnt_work}, although the behavior was not as dramatic as with unconstrained gradient descent.
We observed in practice that projected gradient descent on the squared error loss easily falls into limit cycles of the dynamics.
It was for this reason that we restricted our attention with projected gradient descent to the negative inner product loss.

\section{Multiplicative Weights}
\label{appendix:MW}
When we have simplex constraints $C_f = \Delta_{D_f}$, the Multiplicative Weights algorithm is an elegant way to perform the superposition search. It naturally enforces the simplex constraint by maintaining a set of  auxiliary variables, the `weights', which define the choice of $\mathbf{a}_f$ at each iteration. See equation (\ref{eq:MW_dynamics}) for the dynamics of Multiplicative Weights.
We choose a fixed stepsize $\eta \leq 0.5$ and initial values for the weights all one: $\mathbf{w}_f[0] = \mathbf{1}$. In experiments in this
paper we set $\eta=0.3$.
The variable $\rho$ exists to normalize the term $\frac{1}{\rho}\factorgradient$ so that each element lies in the interval $[-1, 1]$.

Multiplicative Weights is an algorithm primarily associated with game theory and online optimization, although it has been independently discovered in a wide variety of fields
\citep{arora2012multiplicative}. Please see Arora's excellent review of Multiplicative Weights for a discussion of the fascinating historical and analytical details of this algorithm.
Multiplicative Weights is often presented as a decision policy for discrete-time games. However, through a straightforward generalization of the discrete actions into directions
in a continuous vector space, one can apply Multiplicative Weights to problems of \emph{online convex optimization}, which is discussed at length in
\citet{arora2012multiplicative} and \citet{hazan2016introduction}. We can think of solving our problem (\ref{eq:superposition_factorization}) as if it were an online convex optimization
problem where we update each factor $\hat{\mathbf{x}}^{(f)}$ according to its own Multiplicative Weights update, one at a time. The function $\mathcal{L}$ is convex with
respect to $\mathbf{a}_f$, but is changing at each iteration due the updates for the \emph{other} factors - it is in this sense that we are treating
(\ref{eq:superposition_factorization}) as an online convex optimization problem.

\subsection{Multiplicative Weights is a descent method}
\label{sec:MW_lyapunov}
A descent method on $\mathcal{L}$ is any algorithm that iterates $\mathbf{a}_f[t+1] = \mathbf{a}_f[t] + \eta[t] \Delta \mathbf{a}_f[t]$ where the update $\Delta \mathbf{a}_f[t]$ makes
an acute angle with $-\factorgradient$: $\factorgradient^\top \Delta \mathbf{a}_f[t] < 0$. In the case of Multiplicative Weights, we can equivalently define a descent method based on $\weightgradient^\top \Delta \mathbf{w}_f[t] < 0$
where $\tilde{\mathcal{L}}(\mathbf{w}_f)$ is the loss as a function of the \emph{weights} and $\weightgradient$ is its gradient with respect to those weights.
The loss as a function of the weights comes via the substitution $\mathbf{a}_f = \frac{\mathbf{w}_f}{\sum_i w_{fi}} := \frac{\mathbf{w}_f}{\Phi_f}$.
We now prove that $\weightgradient^\top \Delta \mathbf{w}_f[t] < 0$:
\begin{align*}
    \weightgradient &= \frac{\partial \mathbf{a}_f}{\partial \mathbf{w}_f} \frac{\partial \mathcal{L}}{\partial \mathbf{a}_f} \\
&= \begin{bmatrix}
        \frac{\Phi_f - w_{f1}}{\Phi_f^2} & \frac{-w_{f2}}{\Phi_f^2} & \cdots & \frac{-w_{fk}}{\Phi_f^2} \\
        \frac{-w_{f1}}{\Phi_f^2} & \frac{\Phi_f - w_{f2}}{\Phi_f^2} & \cdots & \frac{-w_{fk}}{\Phi_f^2} \\
        \vdots & \vdots & \ddots & \vdots \\
        \frac{-w_{f1}}{\Phi_f^2} & \frac{-w_{f2}}{\Phi_f^2} & \cdots & \frac{\Phi_f -w_{fk}}{\Phi_f^2} \\
    \end{bmatrix} \factorgradient \\
&= \Big( \frac{1}{\Phi_f} \mathbf{I} - \frac{1}{\Phi_f^2} \mathbf{1} \mathbf{w}^\top \Big) \factorgradient \\
&= \frac{1}{\Phi_f} \factorgradient - \frac{\mathcal{L}(\mathbf{a}_f)}{\Phi_f} \mathbf{1} \numberthis
\end{align*}
This allows us to write down $\Delta \mathbf{w}_f[t]$ in terms of $\weightgradient$:
\begin{align*}
\Delta \mathbf{w}_f[t] &= - \frac{1}{\rho} \mathbf{w}_f[t] \odot \factorgradient = - \frac{1}{\rho}  \mathbf{w}_f[t] \odot \Big( \Phi_f \weightgradient +
\mathcal{L}(\mathbf{a}_f[t]) \mathbf{1} \Big) \\
&= - \frac{\Phi_f}{\rho} \text{diag}(\mathbf{w}_f[t]) \weightgradient - \frac{\mathcal{L}(\mathbf{a}_f[t])}{\rho} \mathbf{w}_f[t]  \numberthis
\end{align*}
And then we can easily show the desired result:
\begin{align*}
    \weightgradient^\top  \Delta \mathbf{w}_f[t] &= - \frac{\Phi_f}{\rho} \weightgradient^\top \text{diag}(\mathbf{w}_f[t]) \weightgradient  - \frac{\mathcal{L}(\mathbf{a}_f[t])}{\rho}
    \weightgradient^\top \mathbf{w}_f[t] \\
&= - \frac{\Phi_f}{\rho} \weightgradient^\top \text{diag}(\mathbf{w}_f[t]) \weightgradient  - \frac{\mathcal{L}(\mathbf{a}_f[t])}{\rho}
    \Big(\frac{1}{\Phi_f} \factorgradient^\top - \frac{\mathcal{L}(\mathbf{a}_f[t])}{\Phi_f} \mathbf{1}^\top \Big) \mathbf{w}_f[t] \\
&= - \frac{\Phi_f}{\rho} \weightgradient^\top \text{diag}(\mathbf{w}_f[t]) \weightgradient  - \frac{\mathcal{L}(\mathbf{a}_f[t])}{\rho}
    \Big(\mathcal{L}(\mathbf{a}_f[t]) - \mathcal{L}(\mathbf{a}_f[t]) \Big) \\
&= - \frac{\Phi_f}{\rho} \weightgradient^\top \text{diag}(\mathbf{w}_f[t]) \weightgradient \\
&< 0  \numberthis
\end{align*}
The last line follows directly from the fact that $\mathbf{w}_f$ are always positive by construction in Multiplicative Weights. Therefore, the matrix $\text{diag}(\mathbf{w}_f[t])$
is positive definite and the term $\frac{\Phi_f}{\rho}$ is strictly greater than $0$. We've shown that the iterates of Multiplicative Weights always make steps in descent
directions. When the loss $\mathcal{L}$ is the negative inner product, it is guaranteed to decrease at each iteration. Empirically, multiplicative weights applied to the squared error loss
\emph{also always decreases the loss function}. We said in Appendix \ref{appendix:fixed_stepsize_doesnt_work} that descent on the squared error with a fixed step size is not in general
guaranteed to converge. However, the behavior we observe with Multiplicative Weights descent on the squared error might be explained by the fact that the stepsize is normalized
by $\rho$ at each iteration in this algorithm. Both functions are bounded below over the constraint set $\Delta_{D_f}$, so therefore Multiplicative Weights must converge to a fixed point.
In practice, we pick a step size $\eta$ between $0.1$ and $0.5$ and run the algorithm until the normalized magnitude of the change in the coefficients is below some small threshold:
\[\frac{\Big| \mathbf{a}_f[t+1] - \mathbf{a}_f[t] \Big|}{\eta} < \epsilon \]
The simulations we showed in the Results section utilized $\eta = 0.3$ and $\epsilon = 10^{-5}$.

\section{Map Seeking Circuits}
\label{appendix:MSC}
Map Seeking Circuits (MSCs) are neural networks designed to solve invariant pattern recognition problems. Their theory and applications have been gradually developed by
Arathorn and colleagues over the past $18$ years (see, for example, \citet{arathorn2001recognition, arathorn2002map},
\citet{gedeon2007convergence}, and \citet{harker2007analysis}), but remain largely unknown outside of a small community of vision researchers. In their
original conception, they solve a
``correspondence maximization'' or ``transformation discovery'' problem in which the network is given a visually transformed instance of some template object and has to
recover the identity of the object as well as a set of transformations that explain its current appearance. The approach taken in Map Seeking Circuits is to superimpose the
possible transformations in the same spirit as we have outlined for solving the factorization problem. We cannot give the topic a full treatment
here but simply note that the original formulation of Map Seeking Circuits can be directly translated to our factorization problem wherein each type of transformation
(e.g. translation, rotation, scale) is one of the $F$ factors, and the particular values of the transformation are vectors in the codebooks $\codebooksets$.
The loss function is $\neginnerprod$ and the constraint set is $[0, 1]^{D_f}$ (both by convention in Map Seeking Circuits). The dynamics of Map Seeking Circuits are given in equation (\ref{eq:MSC_dynamics}), with
initial values $\mathbf{a}_f[0]=\mathbf{1}$ for each factor. The small threshold $\epsilon$ is a hyperparameter, which we set to $10^{-5}$ in experiments, along with the stepsize $\eta=0.1$.
\citet{gedeon2007convergence} and \citet{harker2007analysis} proved (with some minor technicalities we will not detail here)
that Map Seeking Circuits always converge to either a scalar multiple of a canonical basis vector, or the zero vector.
That is, $\mathbf{a}_f[\infty] = \beta_f \mathbf{e}_i \,\, \text{or} \,\, \mathbf{0}$ (where ${(\mathbf{e}_i)}_j = 1$ if $j = i$ and $0$ otherwise, and $\beta_f$ is a positive scalar).

Due to the normalizing term $\rho$, the updates to $\mathbf{a}_f$ \emph{can never be positive}. Among the
components of $\factorgradient$ which are negative, the one with the largest magnitude corresponds to a component of $\mathbf{a}_f$ which sees an update of $0$.
All other components are decreased by an amount which is proportional to the gradient. We noted in comments on (\ref{eq:smooth_grad_neginnerprod}) that the smallest element of $\factorgradient$
corresponds to the codevector which best matches $\mathbf{c} \odot \hat{\mathbf{o}}^{(f)}$, a ``suggestion'' for $\hat{\mathbf{x}}^{(f)}$ based on the current states of the other factors.
The dynamics of Map Seeking Circuits thus preserve the weight of the codevector which matches best and
decrease the weight of the other codevectors, by an amount which is proportional to their own match. Once the weight on a codevector drops below the threshold, it is set to zero and no
longer participates in the search. The phenomenon wherein the correct coefficient $a_{fi^\star}$ drops out of the search is called ``sustained collusion'' by Arathorn \citep{arathorn2002map} and
is a failure mode of Map Seeking Circuits.

\section{Percolated noise in Outer Product Resonator Networks}
\label{appendix:stable_memory_cap}
A Resonator Network with outer product weights $\OPsynapses$ that is initialized to the correct factorization is not guaranteed to remain there, just as a Hopfield Network with outer product weights
initialized to one of the `memories' is not guaranteed to remain there. This is in contrast to a Resonator Network (and a Hopfield Network) with Ordinary Least Squares weights $\OLSsynapses$,
for which each of the codevectors are always fixed points. In this section, when we refer simply to a Resonator Network or a Hopfield Network we are referring to the variants of these models that use outer product weights.

The bitflip probability for the $f$th factor of a Resonator Network is denoted $r_f$ and defined in (\ref{eq:bitflip}). Section \ref{sec:appendix_perc_noise_first_fac} derives $r_1$, which is equal to the bitflip probability for a
Hopfield network, first introduced by (\ref{eq:hopfield_bitflip}) in the main text. Section \ref{sec:appendix_perc_noise_second_fac} derives $r_2$, and then section \ref{sec:appendix_perc_noise_other_fac} collects all of the ingredients
to express the general $r_f$.

\subsection{First factor}
\label{sec:appendix_perc_noise_first_fac}
\begin{figure}[t]
    \centering
    \begin{subfigure}[b]{0.45\textwidth}
        \centering
        \includegraphics[width=\textwidth]{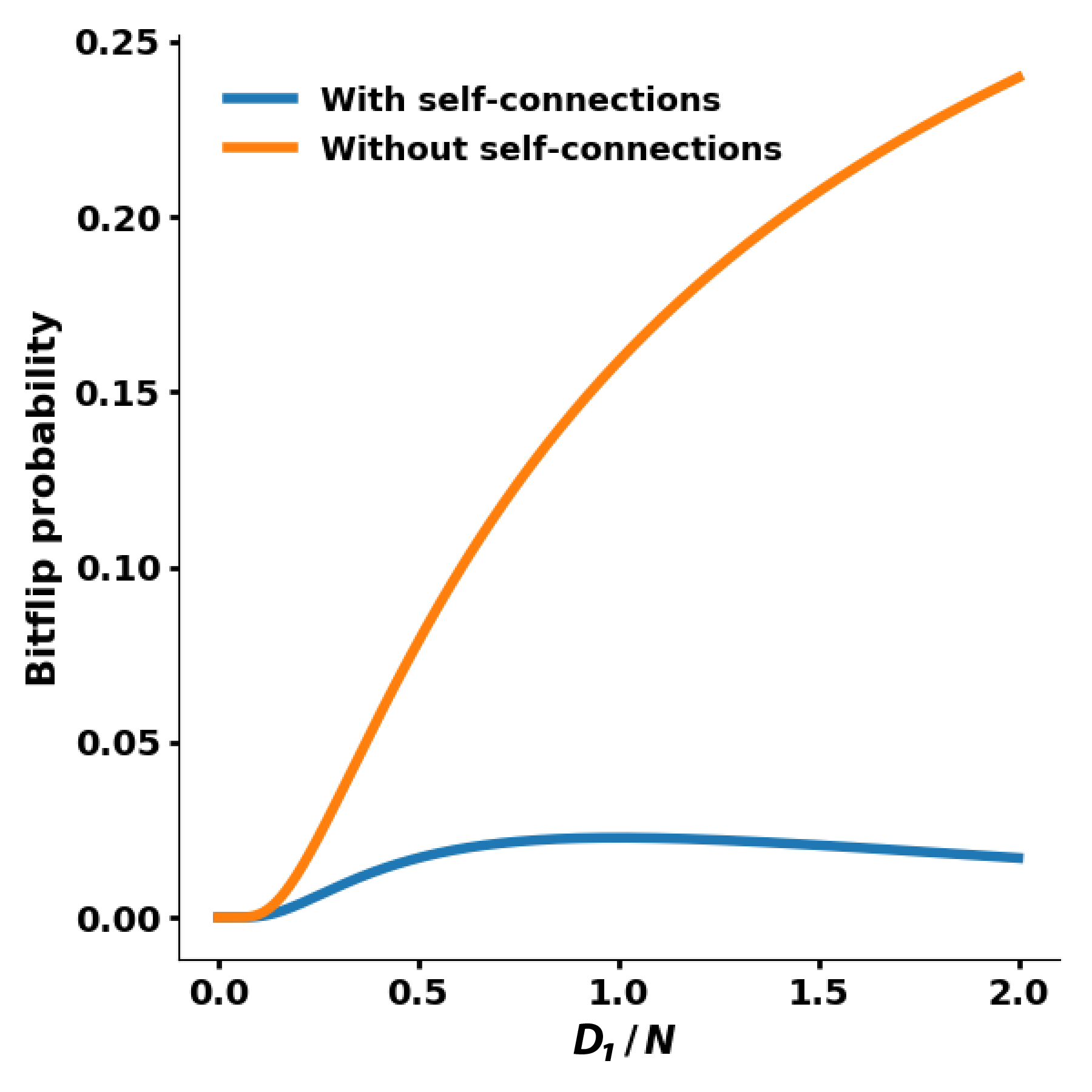}
        \caption{Bitflip prob. for $D_1 \, / \, N \in (0, 2]$}
        \label{fig:appendix_percolated_noise_self_vs_noself_1}
    \end{subfigure}
    \hspace{0.05\textwidth}
    \begin{subfigure}[b]{0.45\textwidth}
        \centering
        \includegraphics[width=\textwidth]{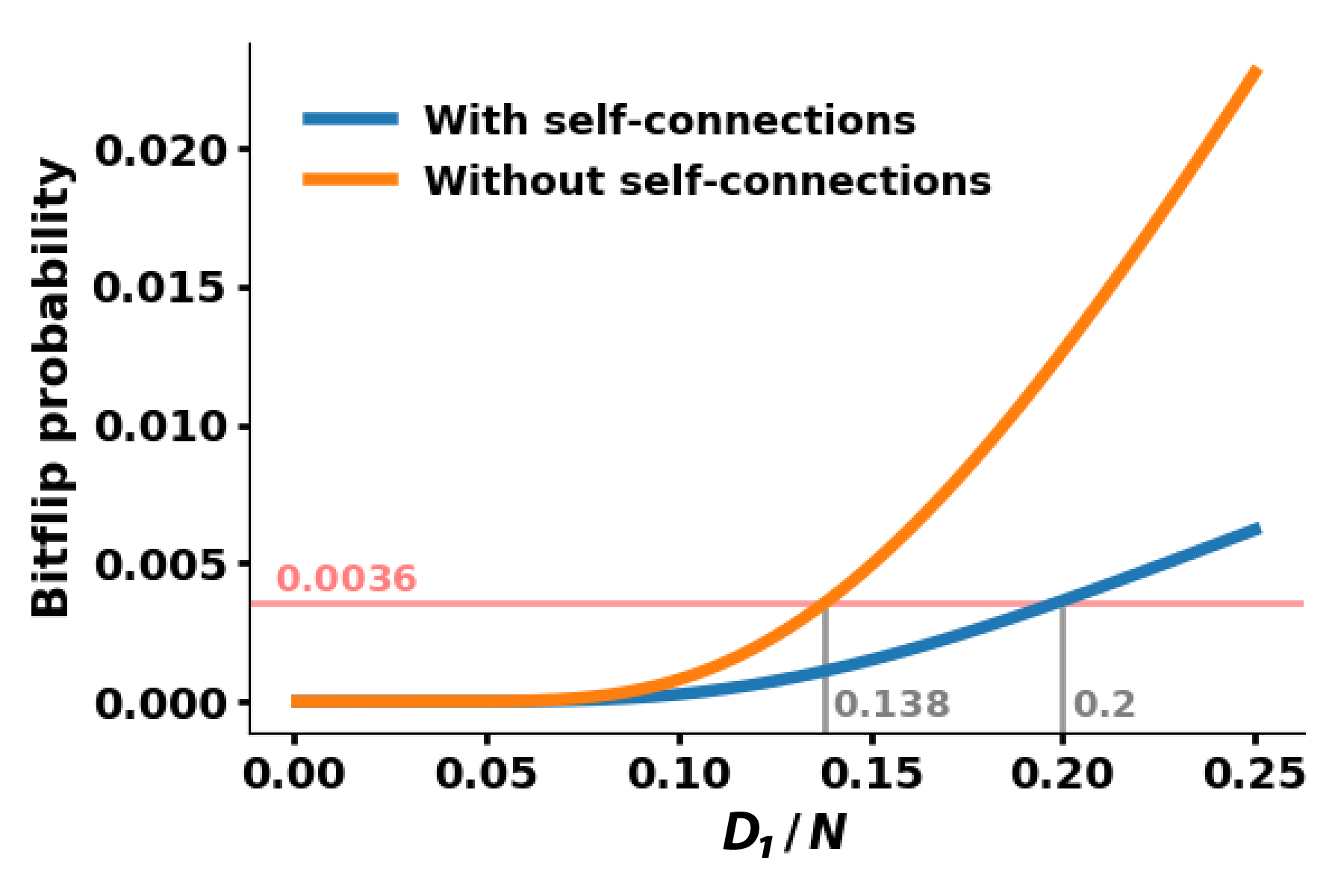}
        \caption{Bitflip prob. for $D_1 \, / \, N \in (0, 0.25]$}
        \label{fig:appendix_percolated_noise_self_vs_noself_2}
    \end{subfigure}
    \caption{Effect of self-connections on bitflip probability}
    \label{fig:appendix_percolated_noise_self_vs_noself}
\end{figure}

The stability of the first factor in a Resonator Network is the same as the stability of the state of a Hopfield network -- at issue is the distribution of $\hat{\mathbf{x}}^{(1)}[1]$:
\begin{equation*}
    \hat{\mathbf{x}}^{(1)}[1] = \text{sgn} \big( \mathbf{X}_1 \mathbf{X}_1^\top \mathbf{x}_\star^{(1)} \big) := \text{sgn} \big( \bm{\Gamma} \big)
\end{equation*}
Assuming each codevector (each column of $\mathbf{X}_1$, including the vector $\mathbf{x}_\star^{(1)}$) is a random bipolar vector, each component of $\bm{\Gamma}$ is a random variable. Its distribution can be deduced from writing it out in terms of constant and random components:
\begin{align*}
\label{eq:derivation_of_first_factor_update}
\Gamma_i &= \sum_m^{D_1} \sum_j^N \big(\mathbf{x}^{(1)}_m\big)_i \, \big(\mathbf{x}^{(1)}_m\big)_j \, \big(\mathbf{x}^{(1)}_\star \big)_j \\
&= N \big(\mathbf{x}^{(1)}_\star \big)_i + \sum_{m \neq \star}^{D_1} \sum_j^N \big(\mathbf{x}^{(1)}_m\big)_i \, \big(\mathbf{x}^{(1)}_m\big)_j \, \big(\mathbf{x}^{(1)}_\star \big)_j \\
&= N \big(\mathbf{x}^{(1)}_\star \big)_i + (D_1 - 1)\big(\mathbf{x}^{(1)}_\star \big)_i + \sum_{m \neq \star}^{D_1} \sum_{j \neq i}^N \big(\mathbf{x}^{(1)}_m\big)_i \, \big(\mathbf{x}^{(1)}_m\big)_j \, \big(\mathbf{x}^{(1)}_\star \big)_j \numberthis
\end{align*}
The third term is a sum of $(N-1)(D_1 - 1)$ i.i.d. Rademacher random variables, which in the limit of large $N D_1$ can be well-approximated by a Gaussian random variable with mean zero and variance $(N-1)(D_1 - 1)$. Therefore, $\Gamma_i$ is approximately
Gaussian with mean $(N + D_1 - 1)\big(\mathbf{x}^{(1)}_\star \big)_i$ and variance $(N-1)(D_1 - 1)$. The probability that $\big( \hat{\mathbf{x}}^{(1)}[1]\big)_i \neq \big(\mathbf{x}^{(1)}_\star \big)_i$ is given by the cumulative density function of the Normal distribution:
\begin{align*}
   \label{eq:appendix_first_factor_update}
   h_1 :=& \,\, Pr \big[\, \big( \hat{\mathbf{x}}^{(1)}[1]\big)_i \neq \big(\mathbf{x}^{(1)}_\star \big)_i \, \big] \\
   =&  \,\, \Phi\Big(\frac{-N-D_1+1}{\sqrt{(N-1)(D_1-1)}} \Big) \numberthis
\end{align*}
We care about the ratio $D_1 \, / \, N$ and how the bitflip probability $h_1$ scales with this number. We've called this $h_1$ to denote the Hopfield bitflip probability but it is also $r_1$, the bitflip probability for the \emph{first} factor of a Resonator Network. We'll see that for the
second, third, fourth, and other factors, $h_f$ will not equal $r_f$, which is what we mean by percolated noise, the focus of Section \ref{sec:results_percolated_noise} in the main text.
If we eliminate all ``self-connection'' terms from $\mathbf{X}_1\mathbf{X}_1^\top$, by setting each element on the diagonal to zero, then the second term in
(\ref{eq:derivation_of_first_factor_update}) is eliminated and the bitflip probability is $\Phi\big(\frac{-N}{\sqrt{(N-1)(D_1-1)}} \big)$.
This is actually significantly different from (\ref{eq:appendix_first_factor_update}), which we can see in Figure \ref{fig:appendix_percolated_noise_self_vs_noself}. With self-connections, the bitflip probability is maximized when $D_1 = N$ (the reader can verify this via simple algebra),
and its maximum value is $\approx 0.023$. Without self-connections, the bitflip probability asymptotes at $0.5$. The actual useful operating regime of both these networks is where $D_1$ is significantly less than
$N$, which we zoom in on in Figure \ref{fig:appendix_percolated_noise_self_vs_noself_2}.
A ``mean-field'' analysis of Hopfield Networks developed by Amit, Gutfreund, and Sompolinsky \citep{amit1985storing, amit1987information} showed that when $D_1 \, / \, N > 0.138$,
a phase-change phenomenon occurs in which a small number of initial bitflips (when the probability is $0.0036$ according to the above approximation) build up over subsequent iterations and the network almost always moves far away from $\mathbf{x}^{(1)}_\star$, making it essentially useless. We can see that the same bitflip
probability is suffered at a significantly higher value for $D_1 \, / \, N$ when we have self-connections -- the vector $\mathbf{x}^{(1)}_\star$ is significantly more stable in this sense. We also found that a Resonator Network has higher operational capacity (see Section
\ref{sec:results_op_cap}) when we leave in the self-connections. As a third point of interest, computing $ \mathbf{X}_f \mathbf{X}_f^\top \mathbf{x}_\star^{(1)}$ is often much faster when we keep each codebook matrix separate (instead of forming the synaptic matrix $\OPsynapses$ directly),
in which case removing the self-connection terms involves extra computation in each iteration of the algorithm. For all of these reasons, we choose to keep self-connection terms in the Resonator Network.

\subsection{Second factor}
\label{sec:appendix_perc_noise_second_fac}
When we update the second factor, we have
\begin{equation*}
    \hat{\mathbf{x}}^{(2)}[1] = \text{sgn} \Big(\mathbf{X}_2 \mathbf{X}_2^\top \big(\hat{\mathbf{o}}^{(2)}[1] \odot \mathbf{c}\big) \Big) := \text{sgn} \big( \bm{\Gamma} \big)
\end{equation*}
Here we're just repurposing the notation $\bm{\Gamma}$ to indicate the vector which gets thresholded to $-1$ and $+1$ by the sign function to generate the new state $\hat{\mathbf{x}}^{(2)}[1]$.
Some of the components of the vector $\hat{\mathbf{o}}^{(2)}[1] \, \odot \, \mathbf{c}$ will be the same as $\mathbf{x}^{(2)}_\star$ and some (hopefully small) number of the components will have been flipped compared to $\mathbf{x}^{(2)}_\star$ by the update to factor $1$. Let us denote the set of
components which flipped as $\mathbb{Q}$. The set of components that did not flip is $\mathbb{Q}^c$. The number of bits that did or did not flip is the \emph{size} of these sets, denoted by $|\mathbb{Q}|$ and $|\mathbb{Q}^c|$, respectively.
We have to keep track of these two sets separately because it will affect the probability that a component of $\hat{\mathbf{x}}^{(2)}[1]$ is flipped relative to $\mathbf{x}^{(2)}_\star$. We can write out the constant and random parts of $\Gamma_i$ along the same lines as what we did in (\ref{eq:derivation_of_first_factor_update}).
\begin{align*}
\label{eq:derivation_of_second_factor_update}
\Gamma_i &= \sum_m^{D_2} \sum_j^N \big(\mathbf{x}^{(2)}_m\big)_i \, \big(\mathbf{x}^{(2)}_m\big)_j \, \big(\hat{\mathbf{o}}^{(2)}[1] \, \odot \, \mathbf{c}\big)_j \\
&= \sum_m^{D_2} \sum_{j \in \mathbb{Q}^c}^N \big(\mathbf{x}^{(2)}_m\big)_i \, \big(\mathbf{x}^{(2)}_m\big)_j \, \big(\mathbf{x}^{(2)}_\star \big)_j - \sum_m^{D_2} \sum_{j \in \mathbb{Q}}^N \big(\mathbf{x}^{(2)}_m\big)_i \,
   \big(\mathbf{x}^{(2)}_m\big)_j \, \big(\mathbf{x}^{(2)}_\star \big)_j \\
&= |\mathbb{Q}^c| \big(\mathbf{x}^{(2)}_\star \big)_i + \sum_{m \neq \star}^{D_2} \sum_{j \in \mathbb{Q}^c}^N \big(\mathbf{x}^{(2)}_m\big)_i \, \big(\mathbf{x}^{(2)}_m\big)_j \, \big(\mathbf{x}^{(2)}_\star \big)_j - |\mathbb{Q}| \big(\mathbf{x}^{(2)}_\star \big)_i -
   \sum_{m \neq \star}^{D_2} \sum_{j \in \mathbb{Q}}^N \big(\mathbf{x}^{(2)}_m\big)_i \, \big(\mathbf{x}^{(2)}_m\big)_j \, \big(\mathbf{x}^{(2)}_\star \big)_j \\
&= (N - 2|\mathbb{Q}|)\big(\mathbf{x}^{(2)}_\star \big)_i + \sum_{m \neq \star}^{D_2} \sum_{j \in \mathbb{Q}^c}^N \big(\mathbf{x}^{(2)}_m\big)_i \, \big(\mathbf{x}^{(2)}_m\big)_j \, \big(\mathbf{x}^{(2)}_\star \big)_j -
   \sum_{m \neq \star}^{D_2} \sum_{j \in \mathbb{Q}}^N \big(\mathbf{x}^{(2)}_m\big)_i \, \big(\mathbf{x}^{(2)}_m\big)_j \, \big(\mathbf{x}^{(2)}_\star \big)_j \numberthis
\end{align*}
If $i$ is in the set of bits which did not flip previously, then there is a constant $(D_2 - 1)\big(\mathbf{x}^{(2)}_\star \big)_i$ which comes out of the second term above. If $i$ is in the set of bits which did flip previously, then there is a constant
$-(D_2 - 1)\big(\mathbf{x}^{(2)}_\star \big)_i$ which comes out of the third term above. The remaining contribution to $\Gamma_i$ is, in either case, a sum of $(N-1)(D_2 - 1)$ i.i.d. Rademacher random variables, analogously to what we had in (\ref{eq:derivation_of_first_factor_update}).
Technically $|\mathbb{Q}|$ is a random variable but when $N$ is of any moderate size it will be close to $r_1 N$, the bitflip probability for the first factor.
Therefore, $\Gamma_i$ is approximately Gaussian with mean either $\big(N(1 - 2r_1) + (D_2 - 1)\big) \big(\mathbf{x}^{(2)}_\star \big)_i$ or $\big(N(1 - 2r_1) - \big(D_2 - 1) \big)\big(\mathbf{x}^{(2)}_\star \big)_i$, depending on whether $i \in \mathbb{Q}^c$ or $i \in \mathbb{Q}$. We call the
\emph{conditional} bitflip probabilities that result from these two cases $r_{2^\prime}$ and $r_{2^{\prime\prime}}$:
\begin{align*}
    \label{eq:appendix_second_factor_update_1}
    r_{2^\prime} :=& \,\, Pr \big[\, \big(\hat{\mathbf{x}}^{(2)}[1]\big)_i \neq \big(\mathbf{x}_\star^{(2)}\big)_i \,\, \big| \,\, \big(\hat{\mathbf{o}}^{(2)}[1] \odot \mathbf{c}\big)_i = \big(\mathbf{x}_\star^{(2)}\big)_i \, \big] \\
    =&  \,\, \Phi\Big(\frac{-N(1-2r_1)-(D_2-1)}{\sqrt{(N-1)(D_2-1)}} \Big) \numberthis
\end{align*}
\begin{align*}
    \label{eq:appendix_second_factor_update_2}
    r_{2^{\prime\prime}} :=& \,\, Pr \big[\, \big(\hat{\mathbf{x}}^{(2)}[1]\big)_i \neq \big(\mathbf{x}_\star^{(2)}\big)_i \,\, \big| \,\, \big(\hat{\mathbf{o}}^{(2)}[1] \odot \mathbf{c}\big)_i \neq \big(\mathbf{x}_\star^{(2)}\big)_i \, \big] \\
    =&  \,\, \Phi\Big(\frac{-N(1-2r_1)+(D_2-1)}{\sqrt{(N-1)(D_2-1)}} \Big) \numberthis
\end{align*}
The total bitflip probability for updating the second factor, $r_2$, is then $r_{2^{\prime}}(1-h_{1}) + r_{2^{\prime\prime}}h_1$.

\subsection{All other factors}
\label{sec:appendix_perc_noise_other_fac}
It hopefully should be clear that the general development above for the bitflip probability of the second factor will apply to all subsequent factors -- we just need to make one modification to notation. We saw that bitflip probability
was different depending on whether the component had flipped in the previous factor (the difference between (\ref{eq:appendix_second_factor_update_1}) and (\ref{eq:appendix_second_factor_update_2})). In the general case, what
really matters is whether the factor sees a \emph{net} bitflip from the other factors. It might be the case that the component had initially flipped but was flipped back by subsequent factors -- all that matters is whether
an \emph{odd} number of previous factors flipped the component. To capture this indirectly, we define the quantity $n_f$ to be the net bitflip probability that is passed on to the next factor (this is equation \ref{eq:net_bitflip}
in the main text):
\begin{equation*}
    n_f := Pr \big[\, \big(\hat{\mathbf{o}}^{(f+1)}[t] \odot \mathbf{c}\big)_i \neq \big(\mathbf{x}_\star^{(f+1)}\big)_i \, \big]
\end{equation*}
For the first factor, $r_1 = n_1$ but in the general case it should be clear that
\begin{equation*}
    r_f = r_{f^\prime}(1 - n_{f-1}) + r_{f^{\prime\prime}} n_{f-1}
\end{equation*}
which is equation (\ref{eq:bitflip_in_terms_of_conditionals}) in the main text. This expression is just marginalizing over the probability that a net biflip was not seen (first term) and the probability that a net bitflip
was seen (second term). The expression for the general $n_f$ is slightly different:
\begin{equation*}
    n_f = r_{f^\prime}(1 - n_{f-1}) + (1 - r_{f^{\prime\prime}}) n_{f-1}
\end{equation*}
which is equation (\ref{eq:net_bitflip_in_terms_of_conditionals}) in the main text. The base of the recursion is $n_0=0$, which makes intuitive sense because factor $1$ sees no percolated noise.

In (\ref{eq:appendix_second_factor_update_1}) and (\ref{eq:appendix_second_factor_update_2}) above we had $r_1$ but what really belongs there in the general case is $n_{f-1}$. This brings us to our general statement for the conditional bitflip probabilities $r_{f^\prime}$ and $r_{f^{\prime\prime}}$, which are equations \ref{eq:conditional_bitflip_1.2} and \ref{eq:conditional_bitflip_2.2} in the main text:
\begin{equation*}
    r_{f^{\prime}} = \Phi\Big(\frac{-N(1-2n_{f-1})-(D_f-1)}{\sqrt{(N-1)(D_f-1)}} \Big)
\end{equation*}
\begin{equation*}
    r_{f^{\prime\prime}} = \Phi\Big(\frac{-N(1-2n_{f-1})+(D_f-1)}{\sqrt{(N-1)(D_f-1)}} \Big)
\end{equation*}

\begin{figure}[t]
    \centering
    \includegraphics[width=0.75\textwidth]{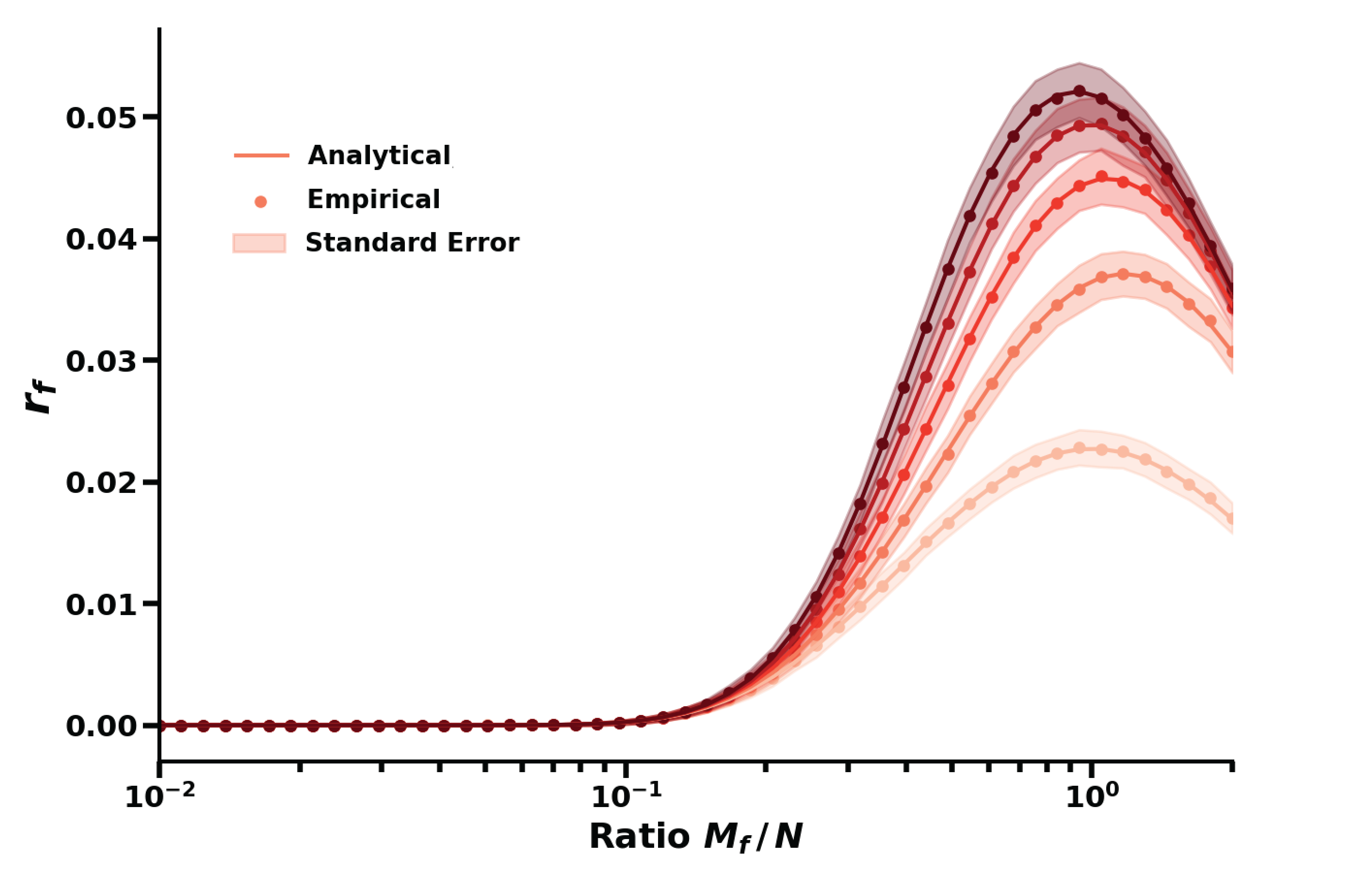}
    \caption{Agreement between simulation and theory for $r_f$. Shades indicate factors $1$-$5$ (light to dark).}
    \label{fig:appendix_percolated_noise_empirical_bitflip_prob}
\end{figure}

What we have derived here in Appendix \ref{appendix:stable_memory_cap} are the equations (\ref{eq:hopfield_bitflip}) - (\ref{eq:conditional_bitflip_2.2}). This result agrees very well with data generated in experiments
where one actually counts the bitflips in a randomly instantiated Resonator Network. In Figure \ref{fig:appendix_percolated_noise_empirical_bitflip_prob} we show the sampling distribution of $r_f$ from these experiments
compared to the analytical expresssion for $r_f$. Dots indicate the mean value for $r_f$ and the shaded region indicates one standard deviation about the mean, the standard error of this sampling distribution. We
generated this plot with $250$ iid random trials for each point. Solid lines are simply the analytical values for $r_f$, which one can see are in very close agreement with the sampling distribution.

\end{document}